\documentclass{article} % For LaTeX2e
\usepackage{iclr2027_conference,times}

\usepackage{amsmath,amsfonts,bm}

\def\eqref#1{equation~\ref{#1}}
\def\1{\bm{1}}

\DeclareMathAlphabet{\mathsfit}{\encodingdefault}{\sfdefault}{m}{sl}
\SetMathAlphabet{\mathsfit}{bold}{\encodingdefault}{\sfdefault}{bx}{n}

\usepackage{wrapfig}
\usepackage{hyperref}
\usepackage{url}
\usepackage{amsmath,amssymb}
\usepackage{booktabs}
\usepackage{array}
\usepackage{tabularx}
\usepackage[T1]{fontenc}
\usepackage[utf8]{inputenc}
\usepackage{multirow} 
\usepackage{graphicx}
\usepackage{adjustbox}
\usepackage{fvextra}
\usepackage{placeins}
\usepackage{url}
\usepackage{amsmath,amssymb}
\usepackage{booktabs,array,longtable,tabularx}
\usepackage{xurl}
\usepackage{graphicx,adjustbox}
\usepackage{fvextra}
\usepackage{placeins}
\usepackage{url}
\newif\ifSCFullProbeTables
\SCFullProbeTablestrue
\providecommand{\scNA}{\textemdash}
\newcolumntype{L}[1]{>{\raggedright\arraybackslash}p{#1}}
\DefineVerbatimEnvironment{scprompt}{Verbatim}{...}
\usepackage{fvextra}

\DefineVerbatimEnvironment{scprompt}{Verbatim}{
  fontsize=\footnotesize,
  breaklines=true,
  frame=single,
  framesep=4pt
}
\newcolumntype{R}[1]{>{\raggedleft\arraybackslash}p{#1}}
\title{Seeing, Saying, but Not Using: From Reportable Spatial Facts to Usable States in Multimodal  Large Language Models}

\author{
Jinchang Zhang, Guoyu Lu
\thanks{Corresponding author. Email: \texttt{guoyulu62@gmail.com}}\\
Intelligent Vision and Sensing (IVS) Lab\\
Indiana University Bloomington\\
Bloomington, IN, USA
}

\iclrfinalcopy % Uncomment for camera-ready version, but NOT for submission.
\begin{document}

\maketitle

\begin{abstract}
A multimodal large language model that correctly reports a spatial fact does
not necessarily use that fact in subsequent reasoning. To study this
distinction, we introduce \textsc{SpaceConflict}, a benchmark of 23{,}196
inputs for the construction and use of spatial state. Under a unified
Supported/Contradictory/Unknown judgment interface, it covers local fact
binding (L1), relational composition (L2), cross-observation consistency
(L3), and state judgment under transformation (L4). Posing a direct-state
query, a full-transformation query, and an explicit-initial-state query on
the same world reveals an availability--utilization gap: models
recover the initial state from visual evidence yet fail when that state must
drive a transformation. For Qwen3.5-9B, 50 of 100 sequences with a correctly
recovered initial state fail the full transformation, and supplying the state
explicitly repairs all 50; the gap narrows with scale but does not close. We
therefore propose Operational State Supervision (OSS), which supervises
task-relevant spatial states and their transformation trajectories and aligns
shared facts across contexts. OSS improves paired accuracy on matched
judgments most on L3 and L4, the levels that depend on organizing and using
state. Evaluating multimodal spatial reasoning thus requires asking not only
whether a model can see and state a spatial fact, but whether that fact
becomes a usable state in subsequent computation. Code and benchmark
artifacts are available at
\url{https://github.com/changcv2021/Seeing-Saying-but-Not-Using-From-Reportable-Spatial-Facts-to-Usable-States-in-Multimodal-Models}.
\end{abstract}

\vspace{-3mm}
\section{Introduction}
\vspace{-2mm}
\label{sec:introduction}
Multimodal large language models (MLLMs) have made rapid progress in visual
understanding and spatial reasoning. However, spatial intelligence requires
more than recognizing objects or reporting pairwise spatial relations. An
agent operating in a changing environment must recover spatial facts from
visual evidence, organize observations into a coherent world state, and
continue to compute over that state as actions, time, or hypothetical
interventions alter the scene. This reveals a distinction that can be obscured
by end-to-end accuracy: a model may correctly report a spatial fact when asked
directly, yet fail when that same fact must serve as the starting point for
subsequent computation.

Existing benchmarks have broadened both the observation settings and the
task complexity of spatial evaluation: VSI-Bench \cite{yang2025thinking} tests memory and reasoning
over layouts from continuous video; Spatial457  \cite{wang2025spatial457} grades tasks from object
recognition to 3D/6D relations; SPAR-Bench \cite{zhang2026flatland} spans perception to higher-order
reasoning in single- and multi-view settings; and Hypo3D \cite{mao2025hypo3d} asks for spatial
states after a described scene change. All of them, however, score the final
spatial judgment. An end-to-end score collapses fact recovery and answer
selection into a single prediction, so an error does not reveal whether the
required state was never recovered or was recovered but not used.

We introduce {SpaceConflict}, a multimodal benchmark designed around
controlled conflicts among visual evidence, spatial claims, and changing world
states. Rather than treating spatial reasoning as a single task,
{SpaceConflict} organizes it into four levels: local fact binding (L1),
relational composition (L2), cross-observation consistency (L3), and
transformation-conditioned state judgment (L4). Together, these levels trace a
progression from spatial facts to an operational world state. The explicit
state trajectory in L4 allows us to query, within the same world, the initial
state, the complete transformation, and the same transformation when the
initial state is explicitly provided, thereby separating whether a fact can be
recovered from whether it can be used. To our knowledge,
{SpaceConflict} is the first multimodal benchmark to combine controlled
spatial conflicts with sequential state transformations while explicitly
distinguishing fact reportability from downstream state usability.
Using this interface, we identify a pronounced
{availability--utilization gap}. Qwen3.5-9B correctly recovers the initial
state under an independent query in 100 of 160 sequences, yet fails the
corresponding full transformation in 50 of those cases. Explicitly supplying
the correct initial state recovers all 50 failures. The gap becomes smaller as
model size increases but does not disappear.
Further controlled experiments narrow the explanation for this phenomenon.
When the same transformations are provided symbolically, models solve nearly
all cases, indicating that the gap therefore cannot be attributed to an inability to execute the transformation itself.
Rewording the query repairs one class of state-selection errors in the
symbolic setting, whereas the visual-task gap appears under fixed wording and
is repaired by supplying the state rather than rewriting the query. In
non-count relation tasks, placing the same state value in the context as an
irrelevant register does not reproduce the benefit of supplying it as the
explicit initial state. Directly injecting candidate state information into
the tested local activation sites likewise fails to reproduce this benefit.
Together, these results indicate that the central issue is not only whether
spatial information is present, but whether it is integrated into subsequent
reasoning in the appropriate computational role.
Motivated by this diagnosis, we introduce
{Operational State Supervision (OSS)}. Rather than supervising only the
final answer, OSS directly supervises task-relevant spatial states and their
transformation trajectories, while aligning spatial facts shared across
different reasoning contexts. Its gains concentrate on L3 and L4, where state
organization and use are most central, and it substantially improves paired
judgment accuracy (PairAcc).
{SpaceConflict} therefore exposes a spatial reasoning problem that is
different from simply ``knowing the answer'': a model may be able to observe
and report a spatial fact without being able to use that fact reliably in
subsequent computation. 

Our contributions are threefold:
1. We introduce {SpaceConflict}, a four-level benchmark that
    traces spatial reasoning from reportable facts to operational states.
    2. We identify a behavioral availability--utilization gap: spatial
    facts that models can report correctly often fail to serve reliably as the
    starting state for subsequent computation.
    3. We propose \textbf{Operational State Supervision (OSS)}, which
    directly supervises task-relevant states and their transformation
    trajectories to strengthen the transition from reportable spatial facts to
    operational states.

\vspace{-3mm}
\section{The SpaceConflict Benchmark}
\vspace{-2mm}
\label{sec:benchmark}
\begin{figure*}[t]
\begin{center}
\includegraphics[width=14cm, height=8cm]{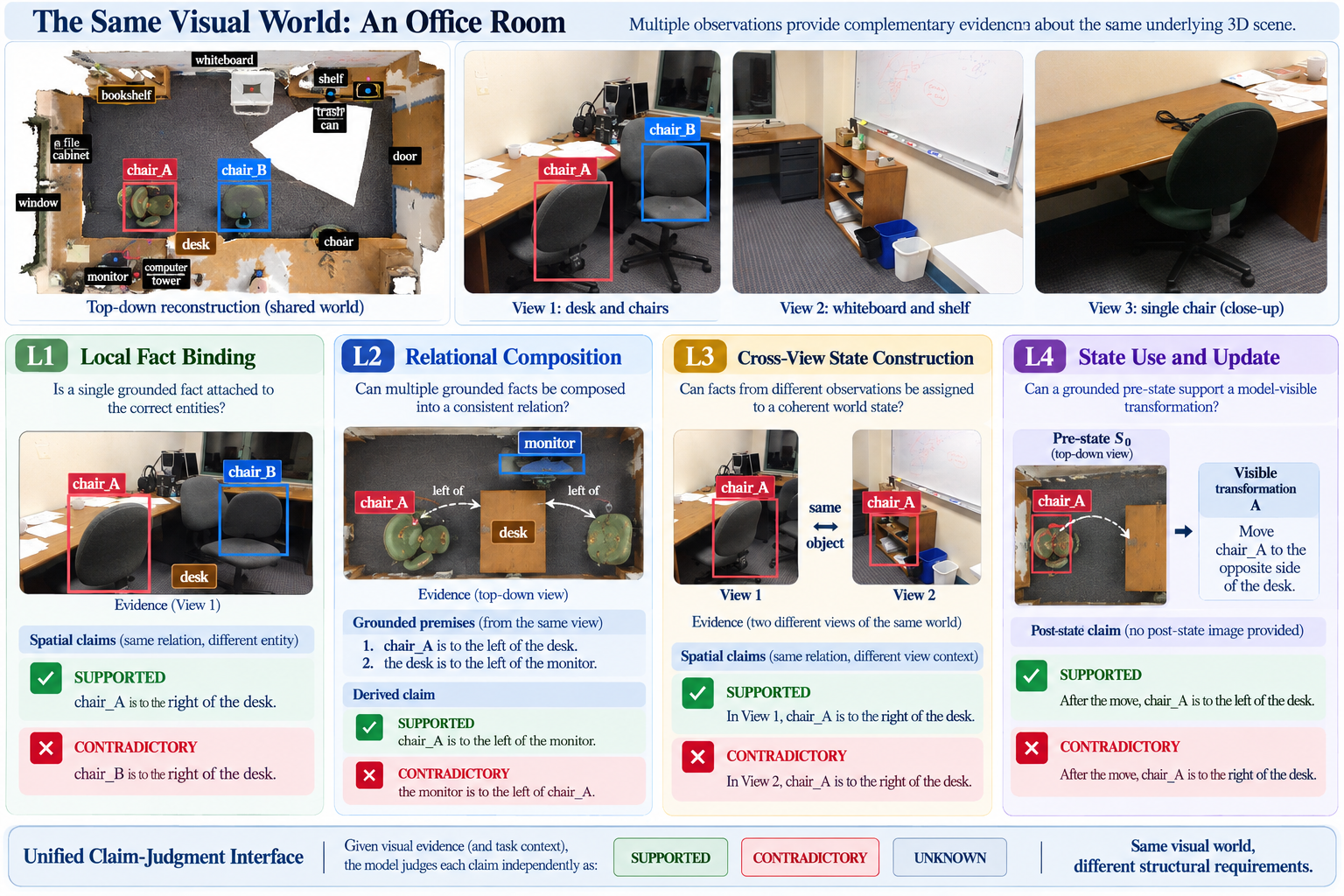}
\end{center}
\vspace{-7mm}
\caption{
Overview of {SpaceConflict} in a shared visual world.
Multiple observations of the same underlying office scene instantiate four
distinct spatial reasoning requirements.
}
\vspace{-7mm}
\label{fig:benchmark}
\end{figure*}
{SpaceConflict} evaluates whether multimodal models can distinguish
spatial claims that agree with an observed world from closely matched claims
that contradict it. Its central design principle is to vary the spatial
semantics of a claim while holding the visual evidence and task context fixed.
As shown in Figure~\ref{fig:benchmark}, \textsc{SpaceConflict} organizes four structural reasoning requirements with progressively increasing dependence on state information under a unified claim-judgment interface, and further asks: {can a directly recoverable spatial fact reliably support a subsequent state transformation?}

\vspace{-3mm}
\subsection{Evidence-Conditioned Spatial Judgments}
\label{sec:definition}
\vspace{-2mm}
We define a {world} $w$ as a set of source-grounded observations that
share the same underlying scene identity. Depending on the organization of the
source dataset, a world may correspond to a single-image scene, a capture, a
parent video, or the base scene of a state transformation. The same world may
produce multiple viewpoints, temporal segments, task contexts, and derived
claims. Thus, one world may give rise to multiple evidence packages
$\mathcal{E}$, while each individual model request observes only one of them.
For a specific model request, we select the evidence visible for the current
task from world $w$ and define the evidence package as
$
    \mathcal{E}=(M,K),
$
where $M$ denotes the model-visible visual evidence, including a single image,
multiple views, or temporally indexed media, and $K$ denotes the task context
relevant to the current judgment, such as a specified viewpoint, temporal
scope, reference frame, or model-visible state transformation.
A model input consists of an evidence package $\mathcal{E}$ and a spatial
claim $c$:
$
    (\mathcal{E},c).
$
The model classifies the claim as
    {Supported},
    {Contradictory},
    {Unknown}
These labels indicate, respectively, that the claim is supported by the
available evidence, explicitly conflicts with the evidence, or cannot be
determined from the available evidence.
For Supported/Contradictory pairs, we define
$
    \mathcal{P}
    =
    (\mathcal{E},c^{+},c^{-},\mathcal{B}),
$
where $c^{+}$ is a claim supported by the evidence, $c^{-}$ is a claim
explicitly refuted under the same evidence conditions, and $\mathcal{B}$
denotes the associated provenance-and-proof bundle, which records the source
facts, controlled edit, relevant evidence, and label justification. The two
claims share exactly the same media and task context, but are submitted to the
model as separate requests.
The negative claim $c^{-}$ is generated by applying one controlled semantic
edit to $c^{+}$; the construction procedure is described in
Section~\ref{sec:construction}.
Label assignment follows the relation semantics and permitted inference rules
defined by each source. A Contradictory claim must be refutable through an
explicit counter-fact, a mutually exclusive relation, a deterministic rule, or
a structural inconsistency; a relation that is absent from the evidence is not
automatically treated as false. Unknown specifically represents cases in which
the current evidence is insufficient to support either the claim or its
negation.
Supported/Contradictory examples are constructed as matched pairs and are
class-balanced. Unknown examples are constructed separately but enter the same
three-way prediction interface. PairAcc is computed only over complete
Supported/Contradictory pairs, whereas Unknown examples participate in the
claim-level three-way evaluation. Complete source mappings, semantic edit
rules, and proof/provenance record formats are provided in
Appendix~\ref{supp:unknown}.

\vspace{-3mm}
\subsection{Four Structural Reasoning Requirements}
\vspace{-2mm}
\label{sec:sc_hierarchy}
\begin{wraptable}{r}{0.33\linewidth}
\vspace{-6mm}
\centering
\small
\setlength{\tabcolsep}{3pt}
\caption{Composition of the evaluation release by reasoning level.}
\label{tab:levels}
\begin{tabular}{@{}lrrr@{}}
\toprule
Level & Pairs & Unknown & Inputs \\
\midrule
L1 & 7,559 & 1,480 & 16,598 \\
L2 & 1,212 & 200 & 2,624 \\
L3 & 1,041 & 520 & 2,602 \\
L4 & 1,136 & 300 & 2,572 \\
\midrule
Total & 10,948 & 2,500 & 24,396 \\
\bottomrule
\end{tabular}
\vspace{-6pt}
\end{wraptable}
Table~\ref{tab:levels} summarizes the four spatial reasoning structures
instantiated in the benchmark.
\textbf{L1 --- Local Fact Binding.}
A claim can be judged from a single source-grounded spatial fact together with
locally permitted relation rules, without composing multiple independent
facts.
\textbf{L2 --- Relational Composition.}
Judging the claim requires combining multiple grounded spatial facts. L2 is
primarily instantiated as irreducible two-premise relation chains under
source-authorized transitivity, such that neither premise alone determines the
answer.
\textbf{L3 --- Cross-Observation Consistency.}
Judging the claim requires placing information from different observation
conditions into the correct viewpoint, temporal, or relational context.
Instances include multiview relations, object appearance order, and
interval-scoped relations.
\textbf{L4 --- Transformation-Conditioned State Judgment.}
The model receives pre-state visual evidence together with a model-visible
hypothetical transformation and determines whether a claim about the
transformed state is consistent with the resulting world. L4 therefore extends
evaluation from current-state judgment to transformation-conditioned
post-state judgment.
Level assignment follows the source-grounded evidence and construction rules.
The actual operators, source composition, and level-specific checks are
detailed in Appendix~\ref{supp:construction}.

\vspace{-3mm}
\subsection{Source-Grounded Construction}
\label{sec:construction}
\vspace{-2mm}
\textsc{SpaceConflict} extracts verifiable spatial facts from existing spatial data sources through deterministic adapters, including structured annotations, metadata, and question--answer records; each adapter normalizes only the information explicitly supported by its source. The specific role and contract of each source are
described in Appendix~\ref{supp:sources}.
Based on these structured facts, we first identify a claim $c^{+}$ supported by the current evidence and then apply a predefined semantic edit to obtain $c^{-}$. Each edit modifies only a single field or a fixed set of fields, such as swapping the subject and object, while leaving all remaining fields unchanged. A claim pair is realized only when the source facts and permitted inference rules explicitly support $c^{+}$ and refute $c^{-}$. The two claims are then generated under a shared language-realization plan, ensuring that their contrast arises from a change in spatial meaning rather than from differences in evidence conditions or surface form. The overall pipeline can therefore be summarized as source evidence $\rightarrow$ structured facts $\rightarrow$ controlled semantic edit $\rightarrow$ matched claims.
Different levels impose different dependencies on these structured facts.
For example, L1 can construct local conflicts through entity rebinding or a
subject--object swap; L2 currently relies primarily on source-authorized
transitivity to construct two-premise relation chains that cannot be resolved
from either premise alone; and L3 uses multiview, appearance-order, or temporal
interval information to construct cross-observation conflicts. The complete
operator set and level coverage are provided in
Appendix~\ref{supp:construction}.
L4 further distinguishes the provenance of the post-state. Among the current
1,136 Supported/Contradictory pairs, 416 are {Native}: their post-state
facts are derived from source-provided context changes and the corresponding
post-state question--answer records. The remaining 720 are
{Controlled}: the benchmark applies model-visible discrete
transformations to source-grounded pre-states and derives local post-states
through deterministic transition procedures.
Pre-state grounding for the Controlled branches combines the available source
annotations for objects, counts, and spatial relations. The specific sources,
geometric rules, and reference-frame conventions are described in
Appendix~\ref{supp:l4}.
For Controlled branches, two independently implemented transition procedures
compute the local post-state and are required to agree. This dual
implementation check validates the deterministic state transition itself.
Both branches use the same L4 model interface described in
Section~\ref{sec:sc_hierarchy}; only the provenance of the post-state differs.
Complete L4 construction, provenance, and validation protocols are provided in
Appendix~\ref{supp:l4}.

\textbf{Provenance and validation.}
\label{sec:certificates}
Validation applies checks appropriate to the corresponding source, level, and
label type, including local refutation, compositional premise verification,
cross-observation dependency checks, and post-state verification. The same
structured records also provide the task-relevant state targets used by the
subsequent diagnostics and OSS training. Complete record formats, validation
scope, and measured coverage are described in
Appendix~\ref{supp:proofs}.
\textbf{Evaluation and Data Protocols.}
The primary task is three-way spatial-claim classification. We report two
metrics: \textbf{ClaimAcc}, the accuracy over all claims including Unknown,
and \textbf{PairAcc}, the fraction of matched Supported/Contradictory pairs
in which both claims are judged correctly; PairAcc is computed only over
paired claims and is the stricter measure. Model-generated explanations are
evaluated only as auxiliary outputs; scoring details are given in
Appendix~\ref{supp:sc_metrics}. The benchmark contains 4{,}284 worlds, with
per-level pair and Unknown counts in Table~\ref{tab:levels} and data splits
in Appendix~\ref{supp:splits}. The diagnostics 
use matched state queries constructed separately from the benchmark claims
(Appendix~\ref{supp:sc_diagnostic_boundary}).

\begin{table}[t]
\centering
\caption{Conditional reporting--transformation gap and rescue (160 sequences/80 worlds per model). Type A counts are rescued sequences/affected worlds; intervals are percentages.}
\label{tab:sc_main_gap}
\small
\setlength{\tabcolsep}{4pt}
\begin{adjustbox}{max width=\linewidth}
\begin{tabular}{lrrrrr}
\toprule
Qwen3.5 & Direct correct & $G$: failures/eligible & 95\% CI & $R$: rescued/gap & Type A seq./worlds \\
\midrule
4B & 104/160 & 64/104 (61.54\%) & [55.77, 68.27] & 35/64 (54.69\%) & 35/30 \\
9B & 100/160 & 50/100 (50.00\%) & [40.00, 60.00] & 50/50 (100.00\%) & 50/37 \\
27B & 96/160 & 11/96 (11.46\%) & [4.17, 19.79] & 11/11 (100.00\%) & 11/8 \\
\bottomrule
\end{tabular}
\end{adjustbox}
\vspace{-8mm}
\end{table}

\vspace{-4mm}

\section{Motivating: From Reportable Facts to Operational States}
\label{sec:motivation}
\vspace{-2mm}
The four levels of {SpaceConflict} place progressively stronger demands on
the role that spatial information plays in reasoning. L1 tests whether a
model can recover local spatial facts; L2 requires composing these facts
relationally; L3 further requires that the resulting world state remain
consistent under changes of observation and reference frame. These three
levels all characterize \textbf{state construction}, differing only in the
degree of structure required. L4 introduces a different requirement: the
inferred state is no longer the final output but must serve as input to
subsequent state transformations. We instantiate the L4 task as a state
trajectory $S_0 \xrightarrow{A_1} S_1 \xrightarrow{A_2} S_2$, where $S_0$
is the spatial state recovered from visual evidence and $A_1$, $A_2$ are
transformations given in text and applied in sequence. Because $S_0$ is
both an output and an input of this trajectory, we can query it directly
or supply it explicitly, and thereby separate three questions that
end-to-end answering conflates: whether $S_0$ can be recovered; whether it
supports the full transformation; and whether the final result changes
when the correct $S_0$ is supplied.
\vspace{-3mm}
\subsection{Availability--Utilization Gap}
\label{sec:sc_availability_gap}
\begin{figure*}[t]
\begin{center}
\includegraphics[width=14cm, height=8cm]{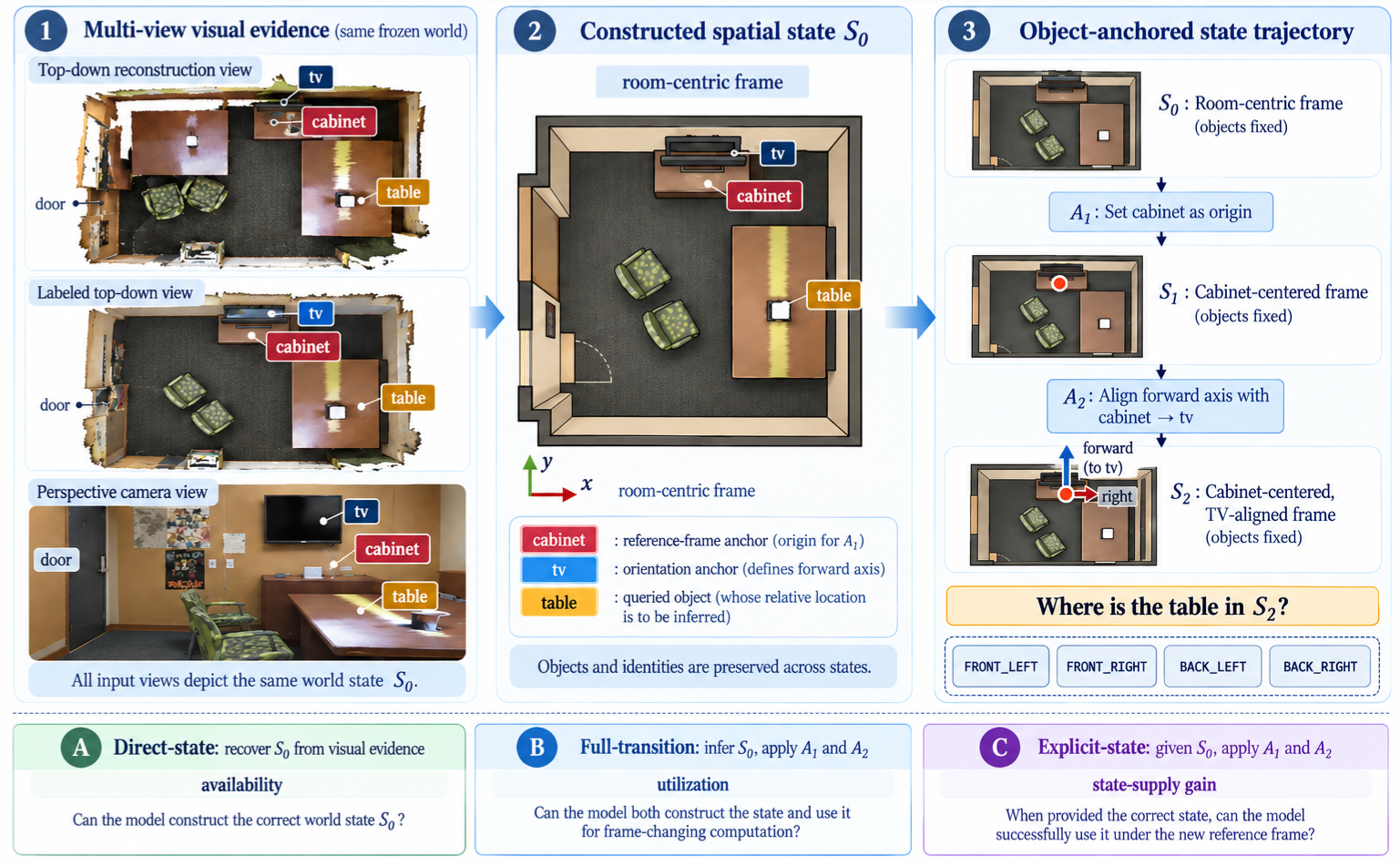}
\end{center}
\vspace{-7mm}
\caption{
From multi-view evidence to an operational world state.
Multiple views of the same world provide visual evidence for
constructing the initial spatial state $S_0$ in a room-centric reference
frame. The state is then transformed without moving any object:
$A_1$ re-centers the reference frame on the cabinet, and $A_2$ aligns
the forward axis with the cabinet-to-TV direction, producing the
cabinet-centered, TV-aligned state $S_2$. The table is then queried in
this transformed reference frame.
}
\vspace{-5mm}
\label{fig:sc_state_trajectory}
\end{figure*}

\vspace{-2mm}
We construct three matched conditions on the same trajectory.
\textbf{Direct-state} asks the model only to recover $S_0$ from visual
evidence, measuring state {availability}. \textbf{Full-transition}
provides the visual input and the transformation descriptions and asks for
the final state $S_2$; the task depends on $S_0$ but does not require the
model to output it, and so measures state {utilization}.
\textbf{Explicit-state} retains the visual input and provides the correct
$S_0$ before the same transformations are applied; its paired difference
from Full-transition measures the gain from supplying the initial state.
Each condition is issued as a separate request, so no condition observes
the output of another. Figure~\ref{fig:sc_state_trajectory} illustrates
the three conditions with a spatial-relation example; condition
definitions and prompts are given in Appendices~\ref{app:sc_behavior}
and~\ref{app:sc_prompts}.
Let $F_i$, $T_i$, and $O_i$ denote whether sequence $i$ is answered correctly
under the direct-state, full-transition, and explicit-state requests,
respectively. We define the conditional availability--utilization gap and the
explicit-state rescue rate as
$
G=
\frac{\sum_i F_i(1-T_i)}
{\sum_i F_i},
R=
\frac{\sum_i F_i(1-T_i)O_i}
{\sum_i F_i(1-T_i)}.
$
Here, $G$ measures how often the full state transformation fails among
sequences for which the required initial state can already be correctly
reported under an independent query. $R$ measures how many of these eligible
failures are recovered once the correct initial state is explicitly supplied.
Table~\ref{tab:sc_main_gap} shows a clear cross-condition behavioral
dissociation. For Qwen3.5-9B, 100 of 160 sequences yield a correct report of
the initial state, yet 50 of these still fail the corresponding full-transition
request, giving $G=50.00\%$. Once the correct initial state is explicitly
provided, all 50 observed failures are recovered ($R=100.00\%$). The other
model scales exhibit the same qualitative pattern, although the magnitude of
the gap differs substantially.
We refer to this cross-condition behavioral dissociation as the
{availability--utilization gap}: a spatial fact can be correctly reported
in one context, yet fail to reliably support a state transformation in another
context that depends on the same underlying state.
The gap thus separates two requirements: having a state available, and
using it in the role the task assigns to it. Sec. \ref{sec:oss} designs
supervision for the second requirement, and Sec. \ref{Diagnosis}
examines what accounts for the gap.
\vspace{-4mm}
\section{Operational State Supervision}
\label{sec:oss}
\vspace{-3mm}
The diagnosis suggests two supervision targets that answer-only training
does not provide. First, the same spatial fact appears in different
roles---as an atomic relation in L1, as part of a world state
in L3, as the initial state of a transformation in L4---and correct
reporting in one role does not transfer to another. We therefore assign a
fact the same canonical target wherever it appears
({cross-context state alignment}). Second, the effect of an
explicit initial state indicates that the weak link is the connection
between the grounded state and the computation that consumes it. The
deterministic L4 trajectory lets us supervise the state at each step
({state-trajectory supervision}). OSS implements both as auxiliary
examples added to fine-tuning, leaving the inference interface
unchanged.

\vspace{-4mm}
\subsection{Canonical State Representation}
\label{sec:oss_representation}
\vspace{-2mm}
We represent each annotated spatial fact as a canonical tuple:
$
    \tau=(e,a,v),
$
where $e$ denotes an entity or entity pair, $a$ denotes the relevant state
variable or spatial relation, and $v$ denotes its value. When required by the
task, $\tau$ is augmented with additional fields such as the reference frame.
A task-relevant state may contain one or more such facts:
$
    \Phi(S)=\{\tau_j\}_{j=1}^{m(S)}.
$
For example, a depth relation can be represented as
{(cabinet, plant; depth\_relation; BEHIND)}
and a count state can be represented as
{(chair; count; 5)}.
Because OSS is trained with an autoregressive language-modeling objective,
each tuple set is converted into a deterministic textual representation
$\operatorname{ser}(\Phi(S))$. We use a fixed field order and vocabulary
throughout the dataset.
For example, a state containing a single fact can be serialized as
{<STATE> entities=...; variable=...; value=... </STATE>}.
For states containing multiple facts, the tuples are ordered deterministically
before serialization.

\vspace{-3mm}
\subsection{Cross-Context State Alignment}
\label{sec:oss_align}
\vspace{-2mm}
The first component of OSS uses the canonical representation above to align
the supervision assigned to the same underlying spatial fact across different
contexts.
Suppose an annotated fact
$
    \tau=(e,a,v)
$
appears as an atomic state queried in L1, as part of a task-relevant state in
L3, and as the initial state variable of an L4 transformation. OSS does not
require the complete states in these tasks to be identical. Instead, it
requires the shared fact itself to retain exactly the same canonical
representation:$
    \tau=\Phi(S^{L1}),
    \tau\in\Phi(S^{L3}),
    \tau=\Phi(S^{L4}_0),$
where the L1 and L4 cases shown here contain a single fact. More generally,
whenever the same annotated fact $\tau$ appears in two state sets, its
serialized form is identical in both.
We implement this constraint through auxiliary state-prediction examples.
Let $Q_{\mathrm{state}}$ specify the task-relevant entity or entity pair and
the corresponding state variable to be reported. An alignment example takes
the form
$
    (I,Q_{\mathrm{state}})
    \longrightarrow
    \operatorname{ser}(\Phi(S)),
$
where $I$ denotes the visual evidence and $\Phi(S)$ denotes the verified
task-relevant state in the current context.
The difference between an alignment example and an original answer example
$(I,Q_{\mathrm{task}})\longrightarrow y
$
lies in the target. In the original example, $y$ is the
{SUPPORTED}/{CONTRADICTORY} label required by the task, whereas
the target of the alignment example is the canonical representation of the
state itself, $\operatorname{ser}(\Phi(S))$. The same construction is applied
to L2 and L3 whenever the benchmark certificate deterministically specifies a
composed relation or world state.

\vspace{-3mm}
\subsection{State-Trajectory Supervision and Training}
\label{sec:oss_training}
\vspace{-2mm}
The second component of OSS supervises the states reached along an L4
trajectory $S_0$ $\xrightarrow{A_1} S_1 \xrightarrow{A_2} S_2$. Together with
the alignment examples of Section~\ref{sec:oss_align}, this yields two
auxiliary pools that are added to the benchmark's own training examples
$\mathcal{D}_{\mathrm{ans}}$, whose targets are the
Supported/Contradictory/Unknown labels:
$
  \mathcal{D}_{\mathrm{align}}
  = \bigl\{\bigl((I, Q_{\mathrm{state}}),\ \operatorname{ser}(\Phi(S))\bigr)\bigr\},
  \mathcal{D}_{\mathrm{traj}}
  = \bigl\{\bigl((I, Q_{\mathrm{state}}, A_{1:t}),\ \operatorname{ser}(\Phi(S_t))\bigr)
    : t \in \{1, 2\}\bigr\},
$
where $Q_{\mathrm{state}}$ names the state variable to be reported and
$A_{1:t}$ is the transformation history applied so far. An alignment
example asks for the canonical state in the current context
(Section~\ref{sec:oss_align}); a trajectory example asks for the state
after $t$ transformations, so its target must be consistent with both the
visual evidence and the transformation history in the input. All targets
are derived from verified source annotations, benchmark certificates, or
deterministic transition programs, never from model predictions, and each
derived example inherits the split of its underlying world, so no
validation or test world contributes a training target.
We fine-tune the same MLLM on
$\mathcal{D}_{\mathrm{OSS}} = \mathcal{D}_{\mathrm{ans}} \cup
\mathcal{D}_{\mathrm{align}} \cup \mathcal{D}_{\mathrm{traj}}$ with the
standard autoregressive objective,
$
  \mathcal{L}_{\mathrm{OSS}}
  = \mathbb{E}_{z \sim \pi,\,(x, y) \sim \mathcal{D}_z}
    \bigl[-\log p_\theta(y \mid x)\bigr],
   z \in \{\mathrm{ans}, \mathrm{align}, \mathrm{traj}\},
$
where $\pi_z$ is the sampling weight of source $z$. We refer to training on
$\mathcal{D}_{\mathrm{ans}} \cup \mathcal{D}_{\mathrm{traj}}$ as
\textbf{OSS-L4} and on the full mixture as {OSS}.
\textbf{Inference.}
At evaluation time the fine-tuned model follows 
\textsc{SpaceConflict} input and decoding protocol. It receives no gold
$S_0$, $S_1$, or $S_2$ and no serialized state, and no state-extraction or
symbolic-execution stage is added, OSS changes only the training signal.
\vspace{-3mm}
\section{Experiment}
\label{sec:sc_results}

\vspace{-3mm}

\subsection{Experimental Setup}
\vspace{-2mm}
\label{experset}

\textbf{Models.}
We first evaluate multiple multimodal large language models on the SpaceConflict benchmark. For the controlled diagnostic experiments, we use Qwen3.5-4B, Qwen3.5-9B, and Qwen3.5-27B, which form a cross-scale comparison within a single model family. Behavioral control experiments and representation probes are run on all three models; causal activation interventions primarily use Qwen3.5-9B as the white-box model. Run details are reported in Appendix~\ref{app:sc_protocol}.

\textbf{Inference protocol.}
All experiments use greedy decoding ({do\_sample=False}). In the main benchmark evaluation, the maximum generation length is 512 tokens, and for models that support a thinking mode, we disable thinking throughout. Detailed settings are given in Appendix~\ref{app:sc_runtime}.

\textbf{Diagnostic panel.}
The three diagnostic conditions and the estimands $G$ and $R$ are defined
in Section~\ref{sec:sc_availability_gap}. The count panel contains 160
transformation sequences from 80 visual worlds and is run for Qwen3.5-4B,
9B, and 27B; the non-count panel contains 47 worlds. Each condition is a
separate request, so all comparisons are between conditions on the same
sequence. Prompts, program controls, and execution settings are given in
Appendices~\ref{app:sc_prompts}, \ref{app:sc_behavior},
and~\ref{app:sc_runtime}.
\textbf{Evaluation Metrics.}
For the L1--L4 benchmark tasks, we report ClaimAcc and PairAcc in Sec. \ref{sec:sc_availability_gap} .  World-level metrics are reported in Appendix~\ref{app:sc_full_results}
State-value queries in the diagnostics are scored by type-constrained exact match against a pre-frozen output schema. For the diagnostic analyses details in Appendix~\ref{app:sc_statistics}.

\vspace{-3mm}
\subsection{Benchmark Results Reveal Heterogeneous Spatial Capabilities}
\label{sec:sc_overall}
\vspace{-2mm}

Table~\ref{tab:main_results} summarizes the complete results of all evaluated
models on \textsc{SpaceConflict}. We make three main observations.
First, different models exhibit clearly different capability profiles across
the four structural reasoning requirements. Qwen3.6-27B achieves the highest
Overall ClaimAcc/PairAcc (\(65.16/52.25\%\)) and the strongest claim-level
performance on L1--L3. However, Qwen3.5-27B obtains higher PairAcc on some
levels, while GPT-5.4 performs best on L4. Thus, stronger aggregate performance
does not imply consistent superiority across different spatial reasoning
requirements.
Second, L1--L4 do not form a monotonically increasing empirical difficulty
ladder. For example, Qwen3.5-9B improves from
\(55.45/39.08\%\) ClaimAcc/PairAcc on L1 to \(89.81/84.57\%\) on L2,
but drops to \(53.65/39.58\%\) on L3 and \(60.73/30.81\%\) on L4.
Similar non-monotonic profiles are observed across other models, supporting
our interpretation of the four levels as distinct structural requirements:
local fact binding, relational composition, cross-observation consistency, and
transformation-conditioned state judgment.
Third, PairAcc reveals consistency failures that are not apparent from
ClaimAcc alone. Qwen3.5-9B and Qwen3.6-27B achieve L4 ClaimAcc scores of
\(60.73\%\) and \(65.05\%\), respectively, while their matched-pair PairAcc
scores are only \(30.81\%\) and \(39.79\%\). Because PairAcc requires both
members of the same Supported/Contradictory pair to be judged correctly, it
provides a stricter measure of consistency under controlled spatial conflicts.
ClaimAcc additionally includes Unknown examples, whereas PairAcc is computed
only over complete Supported/Contradictory pairs; the two metrics therefore
provide complementary rather than directly comparable views of model behavior.
These results motivate a more controlled analysis of L4. Aggregate L4
performance alone cannot distinguish two different sources of failure:
whether the correct initial state is unavailable ({availability}), or
whether a recoverable state is not reliably used in the subsequent
transformation ({utilization}).
\vspace{-2mm}
\subsection{Results of the L4 Diagnosis}
\vspace{-2mm}
\label{Diagnosis}

\textbf{Scale and robustness of the availability--utilization gap.}
Section~\ref{sec:sc_availability_gap}  established a core behavioral dissociation: correctly reporting a state under the Direct-state condition does not guarantee successful downstream state transformation. We next examine the strength of this pattern across model scales and whether it is concentrated in only a small number of underlying worlds.
The conditional availability--utilization gap varies substantially across model scales. Among sequences with correct Direct-state responses, Qwen3.5-4B fails the corresponding Full-transition request in 64/104 cases ($G=61.54\%$), Qwen3.5-9B in 50/100 cases ($G=50.00\%$), and Qwen3.5-27B in 11/96 cases ($G=11.46\%$). When the correct initial state is explicitly supplied, Qwen3.5-4B recovers 35/64 of these eligible failures ($R=54.69\%$), while Qwen3.5-9B and Qwen3.5-27B recover all observed eligible failures, with 50/50 and 11/11 rescued cases, respectively.
These rescued cases span 30, 37, and 8 distinct underlying worlds, respectively, indicating that the effect is not driven by a single repeatedly occurring instance. The following experiments further examine what may account for this phenomenon: whether the failures can be attributed to difficulty in executing the transformation program itself, whether the same pattern extends to non-count spatial states, and whether the benefit of explicit state information depends on how that information is presented and used.
Full condition definitions, behavioral controls, and the complete Type-A--E
case taxonomy are provided in Appendix~\ref{app:sc_behavior};
representative rescued and non-rescued cases are reported in
Appendix~\ref{app:sc_cases}; exact prompts are provided in
Appendix~\ref{app:sc_prompts}; and the bootstrap procedures and conditional
estimands are detailed in Appendix~\ref{app:sc_statistics}.

\begin{table*}[t]
\centering
\caption{Performance comparison across L1--L4 and overall evaluation.}
\label{tab:main_results}
\resizebox{0.8\textwidth}{!}{
\begin{tabular}{lccccc}
\toprule
\multirow{2}{*}{Model}
& L1
& L2
& L3
& L4
& Overall \\
& Claim/Pair Acc
& Claim/Pair Acc
& Claim/Pair Acc
& Claim/Pair Acc
& Claim/Pair Acc \\
\midrule

Qwen2.5-VL-7B
& 23.38/0.85
& 21.86/0.17
& 30.75/0.00
& 26.59/5.81
& 24.36/1.21 \\

Qwen2.5-VL-32B
& 54.43/32.85
& 51.98/43.15
& 52.50/24.88
& 56.18/28.61
& 54.16/32.79 \\

Qwen3.5-4B
& 52.60/34.53
& 83.66/71.20
& 47.08/20.56
& 59.18/28.61
& 55.82/36.65 \\

Qwen3.5-9B
& 55.45/39.08
& 89.81/84.57
& 53.65/39.58
& 60.73/30.81
& 59.26/43.30 \\

Qwen3.5-27B
& 58.62/46.41
& 91.71/87.71
& 57.84/49.47
& 65.05/44.10
& 62.53/51.03 \\

Qwen3.6-27B (non-thinking)
& 61.47/49.36
& 91.75/87.38
& 64.03/45.92
& 65.05/39.79
& 65.16/52.25 \\

Qwen3.8-27B (non-thinking)
& 55.55/39.20
& 63.33/43.99
& 51.33/26.15
& 62.11/41.50
& 56.58/38.73 \\

Qwen3-VL-8B-Instruct
& 50.27/29.77
& 68.56/59.08
& 49.35/20.27
& 41.21/25.70
& 51.04/31.69 \\

Qwen3-VL-2B-Thinking
& 28.74/15.37
& 52.27/40.92
& 4.57/1.92
& 2.72/0.09
& 25.73/15.34 \\

Qwen3-VL-8B-Thinking
& 28.56/16.09
& 63.37/53.71
& 8.38/2.02
& 15.90/7.22
& 28.53/17.99 \\

InternVL3.5-8B
& 57.28/33.77
& 71.86/48.93
& 51.96/21.81
& 53.73/32.83
& 57.79/34.22 \\

InternVL3.5-14B
& 58.90/37.49
& 85.64/73.35
& 53.04/27.47
& 54.74/38.29
& 60.51/40.59 \\

Gemma-4-31B-it
& 56.09/42.49
& 51.36/41.34
& 62.38/37.46
& 62.48/44.54
& 56.97/42.10 \\

MiMo-VL-7B-RL
& 36.23/12.25
& 70.50/60.40
& 24.29/3.65
& 39.04/20.42
& 38.68/17.61 \\

gpt-5.6-luna
& 42.94/22.36
& 73.60/58.00
& 45.12/13.26
& 57.27/35.12
& 47.77/26.76 \\

gpt-5.4 
& 52.55/37.92
& 89.11/85.40
& 58.42/37.08
& 67.57/45.95
& 58.44/43.93 \\

claude-opus-4-7
& 51.89/36.83
& 87.09/82.01
& 55.69/29.01
& 59.18/45.42
& 56.60/41.98 \\

\bottomrule
\end{tabular}
}
\vspace{-6mm}
\end{table*}

\textbf{Explicit state information changes the outcome.}

\label{sec:sc_program_execution}

\begin{wraptable}{r}{0.52\linewidth}
\vspace{-6mm}
\centering
\caption{
Behavioral and program controls on the common count panel.
Values are accuracy (\%) over 160 sequences from 80 worlds.
}
\label{tab:sc_main_bridge}
\small
\setlength{\tabcolsep}{4pt}
\begin{tabular}{@{}lrrr@{}}
\toprule
Condition & 4B & 9B & 27B \\
\midrule
Direct state ($s_0$ from vision)
& 65.00 & 62.50 & 60.00 \\

Full visual transition
& 27.50 & 35.00 & 58.75 \\

Explicit $s_0$ + visual context
& 64.38 & 100.00 & 100.00 \\

Explicit $s_1$ + visual context
& 76.88 & 65.00 & 86.25 \\

Symbolic full program
& 87.50 & 100.00 & 100.00 \\

Symbolic second-action suffix
& 100.00 & 100.00 & 100.00 \\
\bottomrule
\end{tabular}
\vspace{-4mm}
\end{wraptable}

As shown in Table~\ref{tab:sc_main_bridge}, explicitly providing the correct
$s_0$ in the full visual task increases Qwen3.5-9B accuracy from 35.00\% to
100.00\%, while the paired net gains for 4B and 27B are 36.88 and 41.25
percentage points, respectively; details are provided in
Appendix~\ref{app:sc_behavior}. These programs are not intrinsically
inexecutable for the three models: when the initial state $s_0$ and the complete
transformation program are both provided symbolically, 9B and 27B achieve
100\% accuracy; when the intermediate state $s_1$ is directly provided and only
the second action remains to be executed, all three models achieve 100\%.
Therefore, failures in the full visual task cannot be explained simply by an
inability to execute the corresponding transformation rules. Once the relevant
state is explicitly provided, the same programs can be completed reliably.
At the same time, which state is provided also matters. When the full visual
context is retained but $s_1$ rather than $s_0$ is supplied, Qwen3.5-9B
accuracy increases only to 65.00\%, and seven sequences that were correct under
the baseline condition become incorrect. Thus, merely placing a correct state
value in the input is not sufficient; that state must also play the role
required by the task. The following non-count experiment further tests this
role dependence using a matched sham control. Details in
Appendix~\ref{app:sc_behavior}.

\begin{wraptable}{r}{0.49\linewidth}
\centering
\vspace{-8mm}
\caption{Non-count coordinate-transformation results. Accuracy cells are correct worlds/47.}
\label{tab:sc_main_noncount}
\small
\setlength{\tabcolsep}{3.5pt}
\resizebox{\linewidth}{!}{
\begin{tabular}{lrrrrrr}
\toprule
Qwen3.5 & Direct & Full & Explicit & Sham & $\Delta$ (pp) & 95\% CI \\
\midrule
4B  & 35/47 & 12/47 & 9/47  & 10/47 & -2.13 & [-19.15, 14.89] \\
9B  & 42/47 & 11/47 & 37/47 & 9/47  & 59.57 & [40.43, 76.60] \\
27B & 41/47 & 4/47  & 47/47 & 14/47 & 70.21 & [57.45, 82.98] \\
\bottomrule
\end{tabular}
}
\vspace{-5mm}
\end{wraptable}

\textbf{The same contrast appears in non-count relations.}
We repeat the diagnosis on 47 non-count worlds with horizontal, vertical,
and depth-relation transformations: given a coordinate transformation, the
model must state how the relation between two objects changes. A matched
Sham condition contains the same correct relation token (e.g., {BEHIND}) but marks it as irrelevant to the task, so the
Explicit--Sham contrast isolates the role of the state from the presence
of its value. Table. \ref{tab:sc_main_noncount} shows an Explicit-over-Sham gain of
59.57 points for 9B and 70.21 points for 27B, with intervals excluding
zero. For 4B the difference is $-2.13$ points with an interval spanning
zero, and in the depth subset 9B does not benefit either (0/7 vs.\ 2/7).
The role dependence observed on counts therefore carries over to
relational states, with a magnitude that depends on model and relation
type. 
Prompts, relation-specific results, and conditional analyses are
provided in Appendix~\ref{app:sc_noncount} and
Appendix~\ref{app:sc_prompts_noncount}.

\textbf{Target wording is a separate source of error.}
Intermediate-state queries are sensitive to how the target state is named.
Across 96 symbolic programs from 16 families, asking for $S_1$ by its
index alone yields 37/96 for 4B and 0/96 for 9B and 27B, whereas three
prespecified alternatives (a temporal description, explicit state
definitions, and a repeated target instruction) yield 96/96 for all three
models. The index-only failure is therefore an interface effect rather than
a stable preference for the wrong state. More explicit wording is not
uniformly better: for 9B, correct $S_2$ answers fall from 92/96 under
index-only wording to 74/96 under repeated target instruction.Target-by-wording results are in Table~\ref{tab:sc_w1_full} of Appendix~\ref{app:sc_target}.

\vspace{-3mm}
\subsection{Effectiveness of OSS}
\vspace{-3mm}
\label{sec:oss_effectiveness}
\begin{wraptable}{r}{0.52\linewidth}
\vspace{-7mm}
\centering
\small
\setlength{\tabcolsep}{3pt}
\caption{SFT results on SpaceConflict (test).}
\label{tab:OSS_results}

\resizebox{\linewidth}{!}{
\begin{tabular}{lccccc}
\toprule
\textbf{Model / Method} & \textbf{L1} & \textbf{L2} & \textbf{L3} & \textbf{L4} & \textbf{Overall} \\
\midrule

Base (Qwen3.5-9B)
& 53.23 / 38.91
& 92.64 / 87.88
& 48.57 / 39.91
& 62.60 / 31.56
& 56.53 / 44.49 \\

Answer-SFT Natural
& 42.98 / 58.33
& 70.02 / 59.96
& 11.43 / 8.77
& 74.32 / 50.67
& 42.60 / 51.19 \\

Answer-SFT Balanced
& 50.74 / 78.20
& 95.24 / 92.43
& 38.68 / 69.96
& 62.31 / 38.67
& 53.37 / 73.97 \\

CoT-SFT Partial
& 20.50 / 13.85
& 37.23 / 12.99
& 24.54 / 31.80
& 23.94 / 7.56
& 22.90 / 15.25 \\

OSS-L4
& 50.74 / 77.88
& 92.86 / 87.88
& 44.22 / 92.11
& 73.16 / 56.23
& 55.14 / 78.26 \\

\textbf{OSS (Ours)}
& \textbf{53.53 / 86.94}
& \textbf{96.10 / 94.81}
& \textbf{45.29 / 96.05}
& \textbf{82.36 / 68.44}
& \textbf{58.26 / 86.79} \\

\bottomrule
\end{tabular}
}

\vspace{-4mm}
\end{wraptable}

We test whether the state supervision introduced in Section~\ref{sec:oss}
improves spatial reasoning. Table~\ref{tab:OSS_results} compares the
untuned \textbf{Base} model with several alternatives:
\textbf{Answer-SFT Natural}, which supervises only the final label under the
natural training distribution; \textbf{Answer-SFT Balanced}, which balances
sampling across L1--L4; \textbf{CoT-SFT}, which additionally supervises
free-form intermediate reasoning; and the \textbf{OSS-L4} and \textbf{OSS}
variants introduced in Sec. \ref{sec:oss}. Answer-only supervision improves
PairAcc but does not improve ClaimAcc: Answer-SFT Natural increases Overall
PairAcc from 44.49\% to 51.19\%, while ClaimAcc decreases from 56.53\% to
42.60\%; Answer-SFT Balanced reaches 73.97\% PairAcc, but its ClaimAcc remains
at 53.37\%, below the Base model. CoT-SFT performs substantially worse than
Base on both metrics, reaching only 22.90\% / 15.25\%.
OSS, in contrast, improves both metrics. OSS-L4 reaches
55.14\% / 78.26\% Overall ClaimAcc/PairAcc, while full OSS further improves to
58.26\% / 86.79\%, outperforming Answer-SFT Balanced by 4.89 and 12.82
percentage points, respectively. The gains are concentrated in the two levels
that depend most strongly on state organization and use. On L4,
Answer-SFT Balanced achieves 62.31\% / 38.67\% ClaimAcc/PairAcc, whereas OSS
reaches 82.36\% / 68.44\%; relative to OSS-L4, cross-context alignment
contributes an additional 9.20 and 12.21 percentage points, respectively.
On L3, PairAcc increases from 69.96\% to 96.05\%.
The divergence between the two metrics reflects a common behavioral change
after fine-tuning. Across the effective SFT variants, improvements in PairAcc
are substantially larger than those in ClaimAcc; for example, OSS reaches
96.05\% PairAcc but only 45.29\% ClaimAcc on L3.
This gap is associated with
reduced Unknown recall, because ClaimAcc includes Unknown examples whereas
PairAcc is computed only over complete Supported/Contradictory pairs.
Task-specific fine-tuning therefore makes models more likely to issue a
definite judgment even when the available evidence is insufficient, a pattern
shared across the SFT variants. In contrast, PairAcc depends only on
Supported/Contradictory pairs, so the PairAcc gains of OSS over the SFT
baselines directly reflect improved judgments on matched spatial conflicts
rather than changes in Unknown behavior.

\vspace{-4mm}
\section{Related Work}
\vspace{-2mm}
\textbf{Multimodal Spatial Reasoning.}
Recent work has expanded multimodal spatial reasoning from grounded 2D/3D
relations and metric reasoning, as studied by SpatialVLM \citep{chen2024spatialvlm}, SpatialRGPT \citep{cheng2024spatialrgpt}, and
Q-Spatial \cite{liao2024reasoning}, to multi-view and sequential settings in VSI-Bench \cite{yang2025thinking}, Spatial457 \cite{wang2025spatial457},
SPAR-Bench \cite{zhang2026flatland}, and MM-Spatial \cite{daxberger2025mm}. More recent benchmarks further consider
spatial-temporal changes, interaction, and hypothetical or multi-step
transformations, including STI-Bench \cite{li2025sti}, iVISPAR \cite{mayer2025ivispar}, Hypo3D \cite{mao2025hypo3d}, OmniSpatial \cite{jia2026omnispatial}, STARE \cite{li2026unfolding},
and Cambrian-S \cite{yang2026cambrian}. While these works broaden the scope of spatial reasoning,
\textsc{SpaceConflict} focuses on a complementary question: whether a spatial
fact that can be directly recovered can also serve as a reliable state for
subsequent computation.
\textbf{Multimodal Reasoning Benchmarks.}
Beyond spatial reasoning, recent benchmarks examine structured multimodal
reasoning through intermediate representations, controlled variations,
multi-image inputs, physical reasoning, and compositional or multi-step
inference~\citep{zou2025dynamathdynamicvisualbenchmark,
tang2026chartmuseumtestingvisualreasoning,yang2026mmsibenchbenchmarkmultiimagespatial,
xiang2026seephys,chow2025physbenchbenchmarkingenhancingvisionlanguage,
li2025unveilingcompositionalabilitygap,jiang2025mmecotbenchmarkingchainofthoughtlarge}.
These efforts motivate more diagnostic evaluation beyond aggregate accuracy;
\textsc{SpaceConflict} specifically separates spatial-state availability from
its downstream utilization.

\vspace{-4mm}
\section{Conclusion}
\label{sec:conclusion}
\vspace{-3mm}
We presented \textsc{SpaceConflict}, a benchmark that tests, under one
claim-judgment interface, whether multimodal models can carry spatial facts
from local binding through cross-observation consistency to judgments about
transformed states. Controlled diagnostics reveal an
availability--utilization gap: facts a model reports correctly often do not
enter its subsequent state computation, and supplying the initial state
repairs most such failures. OSS extends supervision from final answers to
task-relevant states and their transformation trajectories, and its gains
concentrate in L3 and L4. A spatial fact can thus be seen and said, yet
still fail to become a state the model can use.

\section*{AI Use Statement}

Large language models were used as writing assistants for translation and
language polishing during manuscript preparation. All AI-assisted text was
reviewed and revised by the authors, who take full responsibility for the
final manuscript. AI tools were not used to determine the scientific
conclusions or to replace author verification of experimental results.
The multimodal language models reported in this paper were used as
{subjects of evaluation} in the benchmark experiments, rather than as
writing assistants. Benchmark labels and evaluation targets were derived from
source-grounded annotations, deterministic construction rules, and the
validation procedures described in the paper, rather than from the evaluated
models themselves.

\section*{Ethics Statement}

\textsc{SpaceConflict} is constructed from existing research datasets and
their associated annotations or question--answer records. We do not collect
new personal or sensitive information from human participants. The benchmark
is intended for research on multimodal spatial reasoning and model
reliability.
Our release will respect the licenses and redistribution requirements of the
underlying data sources. When upstream media cannot be redistributed, we will
release only the derived annotations, identifiers, metadata, and construction
artifacts permitted by the corresponding source licenses, together with
instructions for reconstructing the benchmark from the original sources.

The benchmark measures controlled spatial reasoning capabilities and should
not be interpreted as evidence that a model is reliable for safety-critical
physical-world deployment. Performance may also depend on model versions,
decoding settings, and access to proprietary systems.

\section*{Reproducibility Statement}

We provide the materials required to reproduce the benchmark construction,
evaluation, diagnostic experiments, and OSS training through an anonymous
project page at \url{https://anonymous.4open.science/r/spaceconflict-iclr27/}.
The release includes the \textsc{SpaceConflict} benchmark data and split
definitions, benchmark construction and validation metadata, evaluation and
scoring code, model prompts, decoding settings, diagnostic query sets, and
training code and configurations for the reported OSS experiments. We also
provide the model identifiers, random seeds, and preprocessing settings used
in our experiments.

For data derived from external sources, we provide source identifiers and
reconstruction instructions subject to the corresponding upstream licenses.
For proprietary models accessed through external APIs, exact reproduction may
additionally depend on continued availability of the evaluated model version
and provider-side inference behavior.

\bibliography{iclr2027_conference}
\bibliographystyle{iclr2027_conference}

\clearpage
\appendix

\section{Data Sources and Scoped Spatial Facts}
\label{supp:sources}

\subsection{Source Roles}

{SpaceConflict} normalizes heterogeneous source records into scoped
spatial facts. The normalization preserves the source's meaning of an object,
view, temporal interval, relation, and world. Table~\ref{tab:sources}
distinguishes sources of model-visible evidence from additional annotation
resources used to ground Controlled pre-states. These roles are not mutually
exclusive, and an annotation resource is not counted as a new visual world
merely because it supplies additional facts.

\begin{table}[htbp]
\centering
\small
\setlength{\tabcolsep}{4pt}
\caption{Source roles in construction. World units follow the dataset-specific
contract; supplementary grounding resources contribute annotations for
existing scenes rather than a separate Controlled media source.}
\label{tab:sources}
\begin{tabularx}{\linewidth}{@{}>{\raggedright\arraybackslash}p{0.28\linewidth}>{\raggedright\arraybackslash}p{0.18\linewidth}X@{}}
\toprule
Source & Grouping unit & Construction role \\
\midrule
SPAR~\citep{zhang2025spar} & Image or source scene & Object relations and source-annotated multiview relations. \\
\addlinespace
CA-VQA / CA-1M~\citep{daxberger2025mmspatial,lazarow2024cubify} & Capture & Reference/support frames, instance annotations, and scoped counts or relations. \\
\addlinespace
VSI-Bench~\citep{yang2024thinking} & Parent video & Video-scoped spatial facts, counts, and appearance or observation order. \\
\addlinespace
STI-Bench~\citep{li2025stibench} & Video and interval & Temporally scoped qualitative facts; continuous-valued regression targets are outside this release. \\
\addlinespace
OmniSpatial~\citep{jia2025omnispatial} & Item or clip & Source-grounded spatial question--answer records. Unselected answer options are not treated as refutations. \\
\addlinespace
Hypo3D~\citep{mao2025hypo3d} & Base scene and change & Native transformation descriptions and source post-state question--answer records. \\
\addlinespace
EmbodiedScan~\citep{wang2024embodiedscan} & Annotated scene & Object annotations and 9-DoF boxes used for Controlled pre-state grounding. \\
\addlinespace
3DSSG~\citep{wald2020scenegraphs} & Annotated scene & Explicit object relations used in source-grounded pre-states. \\
\addlinespace
ReferIt3D~\citep{achlioptas2020referit3d} & Scene and object reference & Object identification and reference grounding for existing scenes. \\
\bottomrule
\end{tabularx}
\end{table}

A source contract specifies the permitted raw fields, question--answer
conversions, identity mappings, relation rules, and counting scope. Objects
are merged across observations only through an explicit source mapping, such
as an instance ID, track, object index, or marker. Identical category names
alone do not establish instance identity. The source grouping in
Table~\ref{tab:sources} is distinct from the cross-source identity mapping
used for evaluation and split auditing in Appendix~\ref{supp:splits}.

\subsection{Partial World Graphs and Fact Scope}
\label{supp:facts}

For source world $w$, the construction representation is a partial graph
\begin{equation}
G_w=(V_w,E_w^{\mathrm{native}},E_w^{\mathrm{QA}},
E_w^{\mathrm{derived}},\Omega_d),
\label{eq:sc_partial_graph}
\end{equation}
where $V_w$ contains the relevant entities and observation/context nodes;
$E_w^{\mathrm{native}}$ contains direct annotations;
$E_w^{\mathrm{QA}}$ contains deterministic conversions of official answers;
$E_w^{\mathrm{derived}}$ contains rule-derived facts; and $\Omega_d$ specifies
the source's relation semantics and authorized rules.

A scoped fact is represented by
\begin{equation}
f=(w,m,v,t,s,p,o,r,b,\sigma),
\label{eq:sc_scoped_fact}
\end{equation}
where $m$ is the media reference, $v$ the view or frame, $t$ the temporal or
state scope, $s,p,o$ the subject, predicate, and object or discrete value,
$r$ the reference frame, $b$ the actual or hypothetical branch, and $\sigma$
the counting or observation scope. Available polarity and provenance fields
are associated with the fact. The storage of these fields is source- and
version-dependent; Appendix~\ref{supp:c1} explains their organization across
linked artifacts.

This scope is part of a fact's identity. A relation expressed in camera
coordinates is not interchangeable with the same words in source-world
coordinates, and a category count in one frame is not a count over the union
of support views. Missing information remains unspecified rather than being
filled by a language model. The graph therefore records task-relevant
knowledge, not an exhaustive reconstruction of the scene.

\textbf{Official question--answer records.}
A deterministic adapter extracts the queried entities, relation, and scope,
and uses the official answer to instantiate the associated fact. Accepted
conversions are checked against the original answer. Unchosen options supply
no negative fact unless the source ontology or an explicit annotation
justifies the refutation. Source-record references, hashes, and adapter
versions make the conversion traceable.

\subsection{Diagnostic Tags}
\label{supp:tags}

Level identifies the structural requirement of a construction. The tag
vocabulary describes orthogonal spatial phenomena: \texttt{GEO-TOPO}
(geometric and topological relations), \texttt{XFORM-PROJ} (views and reference
frames), \texttt{IDENTITY} (entity correspondence), \texttt{DYNAMIC}
(temporal relations and state changes), and \texttt{EMBODIED-OBS}
(action-conditioned observation or feasibility). Tags are read from the
relevant construction records. This vocabulary describes the annotation
schema; the realized coverage of the present release is given by the
operators and subsets below, rather than by the vocabulary alone.

\section{Released Construction across the Four Levels}
\label{supp:construction}

\subsection{Local Edits and the Realized Operator Inventory}

A minimal pair is generated by one predefined local operation with an
explicit set of changed fields. All other claim fields are preserved. An
argument swap is therefore one edit even though it changes both subject and
object. Language realizations share the paired construction plan, but a
semantic edit need not be a one-token substitution.

Table~\ref{tab:construction} gives the actual L1--L3 operator counts. The
9,812 pairs are dominated by argument swaps and count-category rebindings.
The table specifies the operator coverage used to interpret the level-wise
experimental results.

% Source: mismatch audit M03. Counts are pairs in production_available_v10.
\begin{table}[htbp]
\centering
\small
\setlength{\tabcolsep}{4pt}
\caption{Realized L1--L3 semantic edits in the original release. Each row
counts retained pairs, not candidate generations or individual claims.}
\label{tab:construction}
\begin{tabularx}{\linewidth}{@{}lXr@{}}
\toprule
Operation & Edited dependency & Pairs \\
\midrule
\texttt{SUBJECT\_OBJECT\_SWAP} & Subject/object assignment & 5,284 \\
\texttt{COUNT\_CLASS\_REBIND} & Category associated with a count & 2,577 \\
\texttt{RELATION\_GRAPH\_UNSAT} & Composed relation consistency & 1,212 \\
\texttt{VISIBLE\_CLASS\_REBIND} & Category associated with visibility & 498 \\
\texttt{APPEARANCE\_ORDER\_ERROR} & Temporal appearance order & 195 \\
\texttt{SUBJECT\_OBJECT\_REBIND} & Entities in a scoped relation & 46 \\
\midrule
Total & & 9,812 \\
\bottomrule
\end{tabularx}
\end{table}

\subsection{L1: Atomic Fact Judgments}
\label{supp:l1}

Atomicity is defined at the claim level. A supported or contradictory L1
claim is decided using one independent source-grounded fact and locally
authorized rules, without conjunction of independent factual premises.
Local rule application can include both inversion and exclusivity; it is not
restricted to exactly one rule.

The positive and negative claims may use different deciding facts. For
example, the original relation of one chair can support $c^{+}$, while the
relation of a second chair refutes an entity-rebound $c^{-}$. The pair-level
evidence bundle may consequently contain both facts. This does not change
the atomic status of either judgment. Table~\ref{tab:sc_schematic_cases}
illustrates this distinction.

\subsection{L2: Irreducible Relation Composition}
\label{supp:l2}

The released L2 subset contains 1,212 pairs instantiated from two-premise
relation chains under source-authorized transitivity. Equivalent paraphrases
or inverses of a premise do not supply an additional independent premise.
The resulting target is obtained by composition rather than by rewriting a
single directly annotated answer.

For the evidence bundle $H=\{f_1,f_2\}$, premise-removal checks assess whether
the recorded conflict justification still follows without either necessary
premise. These checks concern the annotated proof and establish the contribution of
each premise to the composed relation.

\subsection{L3: Multiview and Temporal State Construction}
\label{supp:l3}

Table~\ref{tab:sc_l3_composition} resolves the 1,041 L3 pairs into their
actual construction groups. The largest group uses SPAR multiview relations
with subject/object swaps; the remaining groups concern appearance order and
interval-scoped relations. The multiview group thus changes entity assignment within the relation,
whereas the temporal groups alter order or interval-scoped entity binding.

\begin{table}[htbp]
\centering
\small
\setlength{\tabcolsep}{5pt}
\caption{Released L3 construction groups. The SPAR support-view audit and
its 1,000-candidate denominator are reported separately in
Table~\ref{tab:sc_construction_audits}.}
\label{tab:sc_l3_composition}
\begin{tabularx}{\linewidth}{@{}Xlr@{}}
\toprule
Group & Recorded edit & Pairs \\
\midrule
SPAR multiview relations & Subject/object swap & 800 \\
Appearance-order relations & Appearance-order error & 195 \\
Interval-scoped object relations & Subject/object rebinding & 46 \\
\midrule
Total & & 1,041 \\
\bottomrule
\end{tabularx}
\end{table}

For SPAR multiview construction, necessity checks operate on source-annotated
support views and object references. The historical audit contains 1,273
support-view ablation cases over 1,000 candidate pairs; 800 multiview pairs
are retained in the final release. This records annotation-level dependency
checks. The same audit is marked
\texttt{NOT\_PERFORMED\_SOURCE\_REFERENCE\_TIER} for local pixel validation.
Pixel-level observability review and model-response ablations are separate
forms of evidence and retain their own experiment-specific scope.

\subsection{L4: Transformation-Conditioned State Judgments}
\label{supp:l4}

An L4 input contains pre-state media, a model-visible hypothetical
transformation, and a claim about the resulting state. The transformation is
part of the input rather than a hidden branch label. The original media
usually depict the pre-state. L4 targets are qualitative or discrete,
including counts, existence, identity, and spatial relations; the task does
not require continuous distance, velocity, or angle regression.

\textbf{Native and Controlled origins.}
Native-Direct normalizes one source change-conditioned question--answer
record into a local post-state fact. Native-Aggregated groups official
post-state records by the same source scene and change and deduplicates
equivalent facts. Aggregation expands the available branch evidence; it does
not imply that every final claim requires multiple premises.

Controlled branches apply a benchmark-defined operation to a source-grounded
static pre-state. Table~\ref{tab:l4comp} reports both origins and their
construction-time dependency categories. Native contributes 416 pairs and
Controlled contributes 720. The 300 L4 Unknown claims are additional
individual inputs, not rows in the paired composition.

\begin{table}[htbp]
\centering
\small
\setlength{\tabcolsep}{6pt}
\caption{L4 paired composition. Core and Calibration are construction-time
dependency categories. Counts are pairs; the additional 300 Unknown inputs
are reported in Table~\ref{tab:levels}.}
\label{tab:l4comp}
\begin{tabular}{@{}llrrr@{}}
\toprule
Origin & Subtype & Core & Calibration & Total \\
\midrule
Native & Direct & 183 & 18 & 201 \\
Native & Aggregated & 214 & 1 & 215 \\
Controlled & & 630 & 90 & 720 \\
\midrule
Total & & 1,027 & 109 & 1,136 \\
\bottomrule
\end{tabular}
\end{table}

\textbf{Controlled pre-state grounding and coordinate conventions.}
Controlled is a branch-construction type, not an additional upstream media
dataset. Its object references, counts, and relations are assembled from
source annotations, including EmbodiedScan, 3DSSG, and ReferIt3D. Relations
can come from explicit 3DSSG annotations or from strict axis-separation tests
on EmbodiedScan 9-DoF boxes. The implemented geometric rule uses a margin of
$0.02$ in source-coordinate units and records
\texttt{source\_world\_axes} as its reference frame. Relations failing the
strict separation test are not established by that rule. This geometric
construction of a discrete label is distinct from evaluating continuous
metric predictions.

A world-axis relation is interpreted in the recorded source frame, not as
screen-left or screen-right by default. Object references, the applicable
scope, and any model-visible frame specification must therefore be read
together when inspecting an input. This construction detail does not itself
certify that a source-world axis is visually recoverable in every rendering;
media observability is recorded separately from geometric post-state replay.

\textbf{Transformation vocabulary.}
The documented Controlled DSL includes
\texttt{REMOVE\_AND\_RECOUNT}, \texttt{ADD\_AND\_RECOUNT},
\texttt{REPLACE\_AND\_RECOUNT}, \texttt{REPLACEMENT\_IDENTITY},
\texttt{MOVE\_TO\_OPPOSITE\_SIDE}, and \texttt{SWAP\_POSITIONS}.
These names specify local discrete operations. The transformation engine and
a separately implemented checker compute and compare the local post-state
facts. The replay concerns the specified qualitative update rather than a
physical dynamics simulation.

\textbf{Core and Calibration.}
Core denotes constructions intended to depend jointly on the pre-state and
the transformation; Calibration denotes constructions whose transformation
text strongly constrains the post-state answer. Some dependency fields
(scene-only, change-only, and full-input labels) are assigned by construction
rules. They are distinct from independently executed input-ablation tests.
Native source-post-state checks and Controlled dual-implementation replay are
also distinct from these dependency fields, as summarized in
Table~\ref{tab:sc_validation_scope}.

\subsection{Schematic Examples of Claim Construction}

Table~\ref{tab:sc_schematic_cases} illustrates the local contrasts used to
explain the construction. These are schematic examples in a compact notation,
not additional release cases or model outputs. The L1 example uses different
supporting and refuting facts; the L2 example needs both premises; the L3
example is an order judgment over two observations; and the L4 example adds
a visible transformation to the pre-state.

\begin{table}[htbp]
\centering
\small
\setlength{\tabcolsep}{4pt}
\caption{Schematic construction examples. Relation rules and scopes are
explicit in the evidence column. The examples illustrate the definitions;
release counts refer to the actual retained data, not these teaching examples.}
\label{tab:sc_schematic_cases}
\begin{tabularx}{\linewidth}{@{}lXXX@{}}
\toprule
Level & Evidence and context & Supported claim & Contradictory claim \\
\midrule
L1 & Window chair is right of the table; door chair is left of it. Left/right are exclusive in the named frame. & The window chair is right of the table. & The door chair is right of the table. \\
\addlinespace
L2 & $A$ is left of $B$; $B$ is left of $C$. Strict transitivity is authorized. & $A$ is left of $C$. & $C$ is left of $A$. \\
\addlinespace
L3 & First appearance: chair at frame 2, table at frame 7, in the same ordered sequence. & The chair appears before the table. & The table appears before the chair. \\
\addlinespace
L4 & Initially three chairs; the stated transformation removes one identified chair. & After the change, two chairs remain. & After the change, three chairs remain. \\
\bottomrule
\end{tabularx}
\end{table}

\section{Linked Proof Bundles and Validation}
\label{supp:proofs}

\subsection{Organization of Construction Evidence}
\label{supp:c1}

A pair is associated with a linked provenance and proof bundle, rather than
one certificate file containing every field. Table~\ref{tab:certificates}
summarizes the L1--L3 artifact organization. Shared identifiers connect the
claim, its evidence, the edit operation, and the validation outputs. L4
construction paths have corresponding origin-specific records; their schemas
need not be identical to the L1--L3 schema.

\begin{table}[htbp]
\centering
\small
\setlength{\tabcolsep}{4pt}
\caption{Organization of the linked L1--L3 provenance and proof bundle.
Fields are resolved across associated artifacts; this table is not a list of
mandatory keys in every certificate JSON.}
\label{tab:certificates}
\begin{tabularx}{\linewidth}{@{}p{0.34\linewidth}X@{}}
\toprule
Artifact group & Information recorded \\
\midrule
\texttt{claim\_graphs/} & Canonical supported and contradictory propositions. \\
\texttt{evidence\_subgraphs/} & Supporting/refuting facts and media references. \\
\texttt{construction\_traces/} & Source references and hashes, operator, edit, and preserved fields. \\
\texttt{certificates/} & Proof nodes/edges, required facts, and applicable minimality information. \\
Verifier outputs and release/gold & Check outcomes, retained membership, and evaluation targets. \\
\bottomrule
\end{tabularx}
\end{table}

A proof records the construction's justification under source facts and
permitted rules. It is not a model-generated explanation or a trace of neural
computation. Correction or edited-slot metadata can support subsequent
analysis, but conflict localization and structured correction are not
separately scored primary tasks in the experiments accompanying this chapter.

\subsection{Label Semantics and Unknown Construction}
\label{supp:unknown}

Let $\mathcal{F}_{\mathcal E}$ denote the admissible source-grounded facts
for the named evidence context, including applicable transformed-state facts,
and let $\vdash_{\Omega}$ denote derivability using the source-authorized
rules. The intended label semantics are
\begin{equation}
y(c,\mathcal E)=
\begin{cases}
{Supported}, & \mathcal{F}_{\mathcal E}\vdash_{\Omega}c,\\
{Contradictory}, & \mathcal{F}_{\mathcal E}\vdash_{\Omega}\neg c,\\
{Unknown}, & \mathcal{F}_{\mathcal E}\nvdash_{\Omega}c
\ \text{and}\ \mathcal{F}_{\mathcal E}\nvdash_{\Omega}\neg c.
\end{cases}
\label{eq:sc_label_semantics}
\end{equation}
For a consistent fact set, a claim that violates an authorized exclusivity,
uniqueness, ordering, or local state constraint supplies a refutation. These
logical conditions describe the label semantics; the applicable checks are
implemented by the relevant construction path rather than by assuming a
single universal SAT solver.

Unknown construction records the missing decisive evidence and the compatible
positive/negative completions used to justify underdetermination. A missing
edge is not enough to establish Unknown if authorized derivation already
decides the claim. Likewise, absence of a visible object is not automatically
a negative existence judgment when the counting or observation scope extends
beyond the current view.

Production failures are rejected rather than relabeled Unknown. Examples
include invalid source records, unresolved target mappings, violated action
preconditions, and failed text/graph consistency checks. L2 has 200 Unknown
claims in the current release; the distribution is reported in
Table~\ref{tab:levels}.

\subsection{Source- and Level-Specific Validation}
\label{supp:c5}

Table~\ref{tab:sc_validation_scope} separates the objects checked by the
major validation paths. A common provenance framework does not make local
entailment, support-view dependence, Native post-state agreement, and
Controlled transition replay the same test. In particular, a rule-assigned
dependency field and an independently executed ablation have different
evidential roles.

\begin{table}[htbp]
\centering
\small
\setlength{\tabcolsep}{4pt}
\caption{Validation objects and their scope. Checks are applied according to
source, level, and label type. Construction records distinguish symbolic or
annotation-level checks from visual observability and model-response tests.}
\label{tab:sc_validation_scope}
\begin{tabularx}{\linewidth}{@{}p{0.18\linewidth}XX@{}}
\toprule
Path & Main validation object & Scope of the recorded evidence \\
\midrule
L1 & Local support/refutation under authorized rules & Claim-level atomicity; the pair can use distinct deciding facts. \\
\addlinespace
L2 & Two-premise relation proof and premise-removal checks & Dependency of the annotated conflict justification. \\
\addlinespace
L3 SPAR & Source-annotated support views and object references & Annotation-level support-view audit; local pixel validation is separately marked. \\
\addlinespace
L4 Native & Source change and official post-state answer & Agreement with the source post-state record. \\
\addlinespace
L4 Controlled & Source pre-state, DSL action, and computed post-facts & Agreement between two separately implemented transition procedures. \\
\addlinespace
L4 dependency category & Core/Calibration and associated dependency fields & Construction-time categorization, with independent ablations tracked separately. \\
\addlinespace
Unknown & Missing evidence and witness completions & Underdetermination rather than support or refutation; no binary pair-balance requirement. \\
\bottomrule
\end{tabularx}
\end{table}

Text realization is checked against the corresponding claim structure using
the applicable round-trip and verifier outputs. The validation record identifies the source/path and check result. The
independent L1--L3 verification
path checks the realized text against canonical evidence without using the
generator's proposed label and edit trace as its answer.

\subsection{Candidate Audits and Retained Counts}
\label{supp:checks}

Table~\ref{tab:sc_construction_audits} preserves two available
construction-time audit denominators. The SPAR L2 candidate verification
accepts 1,311 of 2,000 candidates. Its stored field named
\texttt{independent\_\allowbreak{}verifier\_\allowbreak{}agreement\_\allowbreak{}rate} is calculated as accepted
candidates divided by all candidates, yielding 65.55\%; it is therefore a
candidate acceptance rate, not an estimated label-accuracy rate for the
retained release. The SPAR L3 support-view audit covers 1,000 candidate pairs,
whereas the released SPAR multiview subset contains 800 pairs after quota
selection. These are batch-specific records, not a single corpus-wide funnel.

\begin{table}[htbp]
\centering
\small
\setlength{\tabcolsep}{4pt}
\caption{Available construction-time audit records. Candidate acceptance,
annotation-level checks, and final quota-selected membership have distinct
denominators.}
\label{tab:sc_construction_audits}
\begin{tabularx}{\linewidth}{@{}p{0.22\linewidth}p{0.24\linewidth}X@{}}
\toprule
Audit & Recorded counts & Interpretation \\
\midrule
SPAR L2 candidate verification & 2,000 candidates; 1,311 accepted; 689 rejected & 65.55\% candidate acceptance. The full retained L2 count is 1,212 pairs (Table~\ref{tab:levels}). \\
\addlinespace
SPAR L3 support-view audit & 1,000 pairs checked; 1,000 pair passes; 1,273 ablation cases & Source-annotation support-view checks. The final multiview quota retains 800 pairs (Table~\ref{tab:sc_l3_composition}). \\
\bottomrule
\end{tabularx}
\end{table}

The final L1--L3 operator counts and L4 origin counts are given in
Tables~\ref{tab:construction} and~\ref{tab:l4comp}. Quality-based rejection
is distinguished from quota exclusion: an invalid reference or unsupported
rule concerns sample validity, whereas a world or operator cap concerns
which valid candidates enter the evaluation set.

\subsection{Shortcut and Input Audits}

Table~\ref{tab:sc_shortcut_checks} reports the statuses in the L1--L3
\texttt{production\_available\_v10} shortcut audit. The executed text-only
check has accuracy and AUC of 0.5, and the equivalent-inversion check records
no failures in 13,474 cases. Identical paired media are checked statically.
Other perturbation conditions have prepared manifests but no model scores in
that audit. These statuses distinguish measured outputs from available
experiment specifications and do not extend automatically to L4 operators.

\begin{table}[htbp]
\centering
\small
\setlength{\tabcolsep}{4pt}
\caption{Recorded shortcut-control scope for the L1--L3 v10 construction
audit. Static input checks and unscored manifests are not model performance
measurements.}
\label{tab:sc_shortcut_checks}
\begin{tabularx}{\linewidth}{@{}p{0.28\linewidth}p{0.22\linewidth}X@{}}
\toprule
Control & Status & Recorded result or scope \\
\midrule
Text-only label prediction & Executed & Accuracy 0.5; AUC 0.5 for the recorded diagnostic. \\
\addlinespace
Equivalent inversion & Executed & 13,474 cases; 0 failures. \\
\addlinespace
Paired media-only input & Static input audit & Exact paired-media equality passed. \\
\addlinespace
First/last frame, frame shuffle, media shuffle, wrong view & Manifests prepared & No model scores in this construction audit. \\
\bottomrule
\end{tabularx}
\end{table}

\section{Language, Versioned Splits, and Reproducibility}
\label{supp:release}

\subsection{Paired Language Realization}
\label{supp:lang}

The supported claim, contradictory claim, and semantic edit are fixed before
language realization. A paired realization plan specifies sentence structure,
information order, entity reference, and relation wording. The operation's
changed-field set is preserved through this realization. For an L4 pair,
the transformation description and target grounding are shared; the paired
contrast concerns the resulting-state claim.

Shared realization reduces differences unrelated to the intended semantic
edit. Its measurable checks remain those recorded for the corresponding
construction version, including the audit in
Table~\ref{tab:sc_shortcut_checks}. A language-realization constraint is not
reported as an independently measured paraphrase or model-robustness result.

\subsection{Historical Evaluation Release}
\label{supp:stats}

The original composite release combines L1--L3
\texttt{production\_available\_v10} with L4 \texttt{v3\_3}.
Table~\ref{tab:stats} gives its historical split counts. The label totals
match Table~\ref{tab:levels}: 21,896 Supported/Contradictory claims and 2,500
Unknown claims. The original source registration reports 4,484 source-world
identifiers, whereas the experimental \texttt{world\_identity\_v2} mapping
merges these into 4,284 underlying world groups. The two numbers describe
different identity units for the same evaluation input inventory.

\begin{table}[htbp]
\centering
\small
\setlength{\tabcolsep}{5pt}
\caption{Historical composite release statistics. The world row uses
registered source-world IDs; it is not the cross-source world grouping used
in the experimental summaries.}
\label{tab:stats}
\begin{tabular}{@{}lrrrr@{}}
\toprule
Unit & Train & Development & Test & Total \\
\midrule
Registered source-world IDs & 2,682 & 627 & 1,175 & 4,484 \\
Binary-pair media groups & 4,589 & 1,084 & 1,446 & 7,119 \\
Transformation branches & 634 & 173 & 225 & 1,032 \\
Supported/Contradictory pairs & 7,470 & 1,707 & 1,771 & 10,948 \\
Supported/Contradictory inputs & 14,940 & 3,414 & 3,542 & 21,896 \\
Unknown inputs & 282 & 152 & 2,066 & 2,500 \\
All inputs & 15,122 & 3,466 & 5,608 & 24,396 \\
\bottomrule
\end{tabular}
\end{table}

The counts in Table~\ref{tab:stats} also show why all-split and test-only
scores are different evaluation views: the historical test set has a much
larger Unknown component. Likewise, L1 supplies most inputs, while L2
contains no Unknown claims. Overall input-weighted accuracy is therefore
reported together with level-wise results, rather than treated as an
unweighted mean of four interchangeable tasks. Worlds can contribute to
multiple levels; level-wise world counts are not additive.

\subsection{Cross-Source Identity Audit and Training Protocol}
\label{supp:splits}

The original splits keep each registered \texttt{global\_world\_id} in one
partition. A subsequent cross-source identity audit finds 123 merged world
clusters spanning historical partitions, involving 1,299 inputs: 75 clusters
span train/test, 32 train/development, and 16 development/test. For example,
\texttt{hypo3d:scene0001\_00} and \texttt{scannet:scene0001\_00} are distinct
source IDs for the same underlying scene. Identifier-level disjointness is
therefore reported separately from physical-scene disjointness.

Training experiments use the separately frozen
\texttt{approved\_v2} protocol summarized in
Table~\ref{tab:sc_training_split}. It resolves known source aliases,
ScanNet physical-space identities, and exact-media connected components,
while retaining the original test payload. The audit's world counts use this
protocol's identity grouping, not the original registration or
\texttt{world\_identity\_v2} grouping. Identity recovery remains limited by
the scene identifiers available from each source.

\begin{table}[htbp]
\centering
\small
\setlength{\tabcolsep}{6pt}
\caption{Frozen training protocol after the cross-source identity audit.
Test inputs are unchanged from the historical release; training and
development membership follow the later audited protocol.}
\label{tab:sc_training_split}
\begin{tabular}{@{}lrr@{}}
\toprule
Split & Inputs & World groups in this audit \\
\midrule
Train & 13,476 & 2,221 \\
Development & 3,267 & 571 \\
Test & 5,608 & 1,124 \\
\bottomrule
\end{tabular}
\end{table}

The full-release benchmark comparison and the fine-tuned model evaluation
thus use distinct declared protocols. Method evaluation uses the held-out
test inputs rather than reporting a mixture of training, development, and
test accuracy. Auxiliary training samples inherit membership from their
underlying world in the training protocol, including any state targets
derived from the same media or hypothetical branch. Historical model scores
remain associated with their original run manifests and clustering choices.

\subsection{Reproducibility Artifacts and Distribution Scope}
\label{supp:artifacts}

Source references, source-record hashes, adapter and operator versions,
construction traces, proof artifacts, split membership, and evaluation
manifests support tracing retained examples to the frozen construction.
Internal evaluation bundles contain the media needed to run the benchmark;
their scope is distinct from public redistribution. Redistribution of
upstream-restricted media is excluded in favor of source references and local
materialization instructions.

For the reported experiments, test labels and proof materials are stored in
internal evaluation artifacts. The release design treats public annotation
access, split-specific certificate access, and media redistribution as
separate components. Reproduction is grounded in the versioned source records,
construction code, and evaluation manifests described above.

\section{Evaluation Interfaces and Diagnostic Scope}
\label{supp:sc_evaluation}

\subsection{Claim-Level and Pair-Level Scoring}
\label{supp:sc_metrics}

The experiments use one three-label classification interface across L1--L4.
The output requests a label, confidence, and brief explanation:
\begin{quote}
\small\ttfamily
\{"label":"SUPPORTED|CONTRADICTORY|UNKNOWN",\\
\hspace*{1em}"confidence":0.0,"reason":"brief explanation"\}.
\end{quote}
Here the pipe-separated label string describes the allowed alternatives;
an actual response contains exactly one label. ClaimAcc and PairAcc. Complete pairs alone enter
PairAcc, and a pair is correct only when both separately generated labels
match their respective gold labels.

The original E0 protocol retains at most 512 generated tokens. Its label
scorer uses the retained output: a recoverable valid label is not discarded
solely because a trailing explanation or confidence field is incomplete.
The benchmark label parser is distinct from the typed value parser used in
state diagnostics. Explanation scores are auxiliary automatic assessments
and do not enter ClaimAcc or PairAcc.

\textbf{Additional metrics.}
BinaryClaimAcc restricts the individual-claim denominator to the
Supported/Contradictory subset. Contradictory and Unknown recall condition on
the corresponding gold labels. World-macro accuracy averages per-world
accuracies, rather than counting entirely correct worlds. When confidence
intervals accompany the original E0 results, they use the stored
source-stratified, world-cluster bootstrap with 5,000 resamples. Training-run
comparisons retain their declared protocol and grouping.

ClaimAcc and PairAcc measure different events and can use different example
populations because ClaimAcc includes Unknown. Their numerical difference is
not by itself an estimate of within-pair error dependence. In particular,
ClaimAcc squared is not the relevant independence reference when the
Supported and Contradictory marginal accuracies differ or Unknown examples
enter ClaimAcc. Such dependence analyses require the two marginals computed
on the same complete-pair population.

\subsection{Benchmark Claims versus Derived State Diagnostics}
\label{supp:sc_diagnostic_boundary}

The release evaluates supported, contradictory, and underdetermined claims.
The behavioral analysis additionally constructs matched state queries from
selected source worlds. Direct-state queries ask for a task-relevant initial
value; full-transition queries provide the visual evidence and transformation
descriptions and ask for the final value; explicit-state queries additionally
supply the correct initial value. These queries are separate model requests
and use the diagnostic value-scoring interface, not the E0 claim classifier.

The primary count panel contains 160 sequences from 80 source worlds. Each
world has an inverse sequence and a two-step sequence; the inverse returns to
its initial count, so the supplied initial value also equals the final value
on that trajectory. The counting scope remains the reference image, rather
than the union of support views. A separate 47-world relation panel covers
34 horizontal, 6 vertical, and 7 depth cases using an axis reflection followed
by identity. These diagnostic panels are not additional inputs in the
24,396-claim release and do not redefine the original level membership of
their source worlds.

The distinction matters for the progression from benchmark to diagnosis and
training. The benchmark provides scoped, source-linked judgments; the
controlled panels expose initial- and resulting-state queries; and auxiliary
supervision is constructed only where the relevant state values and updates
are supported by the corresponding annotations. A shared schema does not
imply that every world supplies every level, a complete scene graph, or a
fully annotated multistep trajectory.

\section{Experimental protocols and statistical conventions}
\label{app:sc_protocol}

It describes execution
and measurement, not benchmark construction. The overall evaluation (E0),
behavioral diagnostics (B1/W1/non-count), representation probes (R1), and
activation interventions (I1/I2) have distinct output and execution contracts.
We preserve those distinctions rather than retrospectively assigning one
protocol to every experiment.

\subsection{Execution settings}
\label{app:sc_runtime}

Table~\ref{tab:sc_runtime} records the diagnostic execution settings. In
particular, the distinction between cached behavioral generation and uncached
intervention replay matters for interpreting the equivalence exclusions in
Appendices~\ref{app:sc_i2} and~\ref{app:sc_i1}. R1 performs forward extraction,
not answer generation. W0 only joins existing predictions and has no additional
decoding configuration.

% Source: supplied evidence report, lines 195--208.
\begin{table}[htbp]
\centering
\caption{Actual diagnostic execution settings. Qwen3.5-4B/9B/27B are abbreviated as the three-model panel. The settings refer to executed runners and recorded environments, not to inferred defaults.}
\label{tab:sc_runtime}
\footnotesize
\setlength{\tabcolsep}{4pt}
\begin{adjustbox}{max width=\linewidth}
\begin{tabular}{L{0.19\linewidth}L{0.25\linewidth}L{0.23\linewidth}L{0.25\linewidth}}
\toprule
Setting & B1 / W1 / non-count & R1 & I1 / I2 \\
\midrule
Models & Three-model panel & Three-model panel & 9B \\
Thinking template & Off & Off & Off \\
Answer decoding & Greedy & No answer generation & Greedy / argmax \\
Output token cap & 512 & Not applicable & 512 \\
Forward precision & BF16 & BF16 & BF16 \\
Other numerical precision & As recorded & Saved residual FP32; probe FP64 & Logits/log-softmax FP32 \\
Generation/forward cache & Enabled & Disabled & Disabled; full-prefix replay \\
Attention backend & SDPA & SDPA & SDPA \\
Batch size & 1 request & 1 context & 1 recipient/trial \\
Image area budget & 100,352--401,408 pixels & Same for visual contexts & Same for I1; I2 has no images \\
Processor & Local checkpoint processor & Same shared pipeline & Same shared pipeline \\
\bottomrule
\end{tabular}
\end{adjustbox}
\end{table}

The recorded environment uses PyTorch 2.8.0+cu126 and Transformers
5.16.0.dev0, without model quantization. The shared processor is
\texttt{Qwen3VLProcessor}. It is loaded from the corresponding local model
directory with \texttt{local\_files\_only=True} and \texttt{use\_fast=True};
no separate processor commit is pinned. Although behavioral records contain a
\texttt{processor\_revision} field, the runner copies the model revision into
that field. It therefore does not constitute an independently verified
processor version.

The image budget in Table~\ref{tab:sc_runtime} is an area bound per image,
not a fixed width/height or a total-request budget. The visual helper resizes
within these bounds before invoking the processor with
\texttt{do\_resize=False}, avoiding a second resize. The recorded patch and
merge sizes are 16 and 2. W1 and the formal I2 interchange panel are symbolic
and use no images. R1 includes both visual B1 contexts and symbolic B2 contexts.

During I1/I2, both patched generation and its unpatched replay baseline set
\texttt{past\_key\_values=None} and recompute the complete causal prefix at
each generated token. Cached references and no-op/self tests are technical
checks, not the formal intervention path. A visual-preprocessing cache is
separate from a model-generation cache. Greedy decoding means
\texttt{do\_sample=False}; no sampling temperature or top-$p$ is used.

\subsection{Metrics, units, and conditioning}
\label{app:sc_statistics}

ClaimAcc weights retained input-level judgments equally. BinaryClaimAcc
restricts the denominator to supported/contradictory judgments. PairAcc counts
a complete pair as correct only when both judgments are correct; unknown
examples are excluded from this paired metric. Contradiction and unknown
recall condition on their respective gold labels. World-macro accuracy is the
mean of per-world accuracies, not the proportion of worlds on which every
answer is correct.

Count diagnostics use a typed nonnegative integer or JSON \texttt{null};
relation diagnostics use the specified enumeration or \texttt{null}. Scores
retain the original normalization and parser decisions. Missing evidence and
null are not interchangeable with numerical zero. Likewise, technical
non-return, an ineligible donor, and an incorrect returned answer are different
statuses. The 512-token budget applies to retained generated output; reaching
that limit does not, by itself, invalidate an otherwise scoreable label or
value. E0's retained-prefix label scorer is distinct from the diagnostic value
scorer, and auxiliary explanation judgments do not enter the primary
accuracies reported here.

For a paired behavioral contrast, rescue counts baseline-wrong cases that
become correct; harm counts baseline-correct cases that become wrong. The
net accuracy difference is
\begin{equation}
 \Delta_{\mathrm{acc}}=
 \frac{n_{\mathrm{rescue}}-n_{\mathrm{harm}}}{N_{\mathrm{paired}}}.
 \label{eq:sc_paired_accuracy}
\end{equation}
Rescue and harm rates have different conditional denominators. A Type A
case already includes successful oracle rescue in its definition; the
non-tautological rescue estimand is $R$ .

Causal probability measurements use the log probability of the complete
canonical JSON candidate. For answer string $y=(y_1,\ldots,y_K)$ and context
$x$, the corresponding quantity is
\begin{equation}
 \ell(y\mid x)=\sum_{k=1}^{K}\log p(y_k\mid x,y_{<k}).
 \label{eq:sc_candidate_lp}
\end{equation}
It is not a single number-token logit, an accuracy percentage, or a normalized
probability over the small candidate list. Changes in this quantity are
reported in natural-log units (nats). No unrecorded candidate scores are
reconstructed.

For I1, let $\ell^{\mathrm{base}}_i$, $\ell^{\mathrm{info}}_i$, and
$\ell^{\mathrm{sham}}_i$ be the correct-answer log probabilities for the
unpatched recipient and the two donor interventions. The matched specificity
estimand is
\begin{equation}
 \delta_w=\frac{1}{|\mathcal I_w|}\sum_{i\in\mathcal I_w}
 \left[(\ell^{\mathrm{info}}_i-\ell^{\mathrm{base}}_i)
      -(\ell^{\mathrm{sham}}_i-\ell^{\mathrm{base}}_i)\right],
 \qquad \bar\delta=\frac{1}{W}\sum_w\delta_w,
 \label{eq:sc_specificity}
\end{equation}
where $\mathcal I_w$ contains eligible paired sequences from world $w$.
Thus I1 weights worlds equally after averaging within-world sequences.
Generation-based endpoints retain their separately stated eligible
sequence denominators.

\textbf{Uncertainty and exposure.}
E0 intervals use the archived source-stratified, world-clustered bootstrap
with 5,000 resamples and seed 20260909. The behavioral and intervention tables
retain their own archived world- or program-family-clustered intervals.
Post-hoc non-count family and Type-A-like summaries use 5,000 world-bootstrap
resamples and seed 20260912; these summaries have no multiplicity adjustment
and are not promoted to predefined primary tests. A reported zero-width
interval from all-correct or all-incorrect empirical samples is not a
population guarantee. Model-dependent conditional cohorts, small families,
and repeated measurements further limit extrapolation.

R1 uses separate train/selection/evaluation partitions. I1 uses LOCALIZE for
window selection and SELECT for frozen-candidate assessment, with a two-window
Bonferroni-adjusted bound in its continuation rule. Neither these labels nor
the existing E0 test-only subset establish that diagnostic worlds were never
encountered during the research process. We do not claim a new globally
unexposed behavioral confirmation set. The manuscript distinguishes missing
confidence intervals from missing predictions and does not invent either.

\FloatBarrier
\section{Complete overall evaluation results}
\label{app:sc_full_results}

Table~\ref{tab:sc_full_overall} expands Table~\ref{tab:main_results} with
label-conditioned and world-macro metrics. For example, high unknown recall
can coexist with very low contradiction recall, as in the Qwen2.5-VL-7B run;
these values describe its end-to-end behavior under the recorded protocol,
not a clean isolation of spatial competence. World-macro and input-micro
rankings can also differ because they weight worlds differently.

% Source: supplied evidence report, lines 63--74.
\begin{table}[htbp]
\centering
\caption{All-split scores for all ten complete models (percent). ClaimAcc uses 24,396 judgments; BinaryClaimAcc uses 21,896 S/C judgments; PairAcc uses 10,948 pairs. C and U are label-conditioned recalls. Intervals available for Qwen3.5 are listed separately in Table~\ref{tab:sc_e0_intervals}.}
\label{tab:sc_full_overall}
\small
\setlength{\tabcolsep}{4pt}
\begin{adjustbox}{max width=\linewidth}
\begin{tabular}{lrrrrrr}
\toprule
Model & Claim & Binary & Pair & C recall & U recall & World macro \\
\midrule
Qwen2.5-VL-7B & 24.36 & 17.71 & 1.21 & 1.24 & 87.74 & 23.39 \\
Qwen2.5-VL-32B & 54.16 & 54.11 & 32.79 & 35.71 & 54.65 & 51.91 \\
Qwen3-VL-2B-Thinking & 25.73 & 27.08 & 15.34 & 25.72 & 12.91 & 21.21 \\
Qwen3-VL-8B-Instruct & 51.04 & 48.86 & 31.69 & 49.13 & 71.78 & 45.90 \\
InternVL3.5-8B & 57.79 & 59.69 & 34.22 & 67.89 & 39.74 & 55.68 \\
InternVL3.5-14B & 60.51 & 63.19 & 40.59 & 67.96 & 35.04 & 62.19 \\
MiMo-VL-7B-RL & 38.68 & 35.44 & 17.61 & 21.11 & 69.57 & 36.57 \\
Qwen3.5-4B & 55.82 & 56.24 & 36.65 & 44.98 & 51.74 & 50.27 \\
Qwen3.5-9B & 59.26 & 60.06 & 43.30 & 53.23 & 51.70 & 52.30 \\
Qwen3.5-27B & 62.53 & 64.24 & 51.03 & 65.77 & 46.26 & 54.39 \\
\bottomrule
\end{tabular}
\end{adjustbox}
\end{table}

Table~\ref{tab:sc_e0_intervals} preserves the available overall confidence
intervals and the correct numerators for each metric. The world-macro
numerator is a sum of per-world accuracies and is therefore generally
non-integer. The absence of intervals for the other completed models in
Table~\ref{tab:sc_full_overall} reflects unavailable archived estimates; it
is not evidence of zero uncertainty.

% Source: supplied evidence report, lines 63--74.
\FloatBarrier
\begingroup
\footnotesize
\setlength{\tabcolsep}{3pt}
\renewcommand{\arraystretch}{1.08}
\begin{longtable}{llL{0.31\linewidth}rrl}
\caption{Available overall Qwen3.5 intervals. Percent estimates and intervals use their original estimands. The world-macro numerator is not a count of entirely correct worlds.}\label{tab:sc_e0_intervals}\\
\toprule
Qwen3.5 & Metric & Numerator/denominator & Worlds & Estimate & 95\% CI \\
\midrule
\endfirsthead
\multicolumn{6}{l}{\footnotesize Table~\thetable\ continued.}\\
\toprule
Qwen3.5 & Metric & Numerator/denominator & Worlds & Estimate & 95\% CI \\
\midrule
\endhead
\midrule\multicolumn{6}{r}{\footnotesize Continued on the next page.}\\
\endfoot
\bottomrule
\endlastfoot
4B & ClaimAcc & 13505/24396 & 4284 & 55.36 & [55.09, 56.56] \\
4B & BinaryClaimAcc & 12315/21896 & 4238 & 56.24 & [55.50, 56.97] \\
4B & PairAcc & 4012/10948 & 4238 & 36.65 & [35.60, 37.66] \\
4B & C recall & 4924/10948 & 4238 & 44.98 & [43.86, 46.12] \\
4B & U recall & 1190/2500 & 858 & 47.60 & [48.76, 54.55] \\
4B & World macro & 2153.7463005609006/4284 & 4284 & 50.27 & [49.50, 51.04] \\
9B & ClaimAcc & 14339/24396 & 4284 & 59.26 & [58.54, 60.01] \\
9B & BinaryClaimAcc & 13150/21896 & 4238 & 60.06 & [59.30, 60.80] \\
9B & PairAcc & 4741/10948 & 4238 & 43.30 & [42.32, 44.28] \\
9B & C recall & 5828/10948 & 4238 & 53.23 & [52.21, 54.23] \\
9B & U recall & 1189/2500 & 858 & 51.70 & [49.02, 54.37] \\
9B & World macro & 2240.523956461215/4284 & 4284 & 52.30 & [51.55, 53.05] \\
27B & ClaimAcc & 15130/24396 & 4284 & 62.53 & [61.75, 63.28] \\
27B & BinaryClaimAcc & 14066/21896 & 4238 & 64.24 & [63.48, 65.01] \\
27B & PairAcc & 5587/10948 & 4238 & 51.03 & [50.04, 51.98] \\
27B & C recall & 7201/10948 & 4238 & 65.77 & [64.83, 66.72] \\
27B & U recall & 1064/2500 & 858 & 46.26 & [43.64, 48.90] \\
27B & World macro & 2329.875445672674/4284 & 4284 & 54.39 & [53.68, 55.11] \\
\end{longtable}
\endgroup

Table~\ref{tab:sc_e0_levels} supplies the complete level-wise comparison.
Its non-monotonic profile should be interpreted alongside differences in
question and source composition rather than as a direct decomposition of
neural failure stages. In particular, a high L2 score does not imply that
all intermediate premises were correctly represented in the same run.

% Source: supplied evidence report, lines 85--96.
\begin{table}[htbp]
\centering
\caption{Level-wise scores (percent), shown as ClaimAcc/PairAcc. Level definitions are given in the benchmark chapter, not repeated here. Levels may share underlying worlds.}
\label{tab:sc_e0_levels}
\small
\setlength{\tabcolsep}{4pt}
\begin{adjustbox}{max width=\linewidth}
\begin{tabular}{lrrrr}
\toprule
Model & L1 & L2 & L3 & L4 \\
\midrule
Qwen2.5-VL-7B & 23.38/0.85 & 21.86/0.17 & 30.75/0.00 & 26.59/5.81 \\
Qwen2.5-VL-32B & 54.43/32.85 & 51.98/43.15 & 52.50/24.88 & 56.18/28.61 \\
Qwen3-VL-2B-Thinking & 28.74/15.37 & 52.27/40.92 & 4.57/1.92 & 2.72/0.09 \\
Qwen3-VL-8B-Instruct & 50.27/29.77 & 68.56/59.08 & 49.35/20.27 & 41.21/25.70 \\
InternVL3.5-8B & 57.28/33.77 & 71.86/48.93 & 51.96/21.81 & 53.73/32.83 \\
InternVL3.5-14B & 58.90/37.49 & 85.64/73.35 & 53.04/27.47 & 54.74/38.29 \\
MiMo-VL-7B-RL & 36.23/12.25 & 70.50/60.40 & 24.29/3.65 & 39.04/20.42 \\
Qwen3.5-4B & 52.60/34.53 & 83.66/71.20 & 47.08/20.56 & 59.18/28.61 \\
Qwen3.5-9B & 55.45/39.08 & 89.81/84.57 & 53.65/39.58 & 60.73/30.81 \\
Qwen3.5-27B & 58.62/46.41 & 91.71/87.71 & 57.84/49.47 & 65.05/44.10 \\
\bottomrule
\end{tabular}
\end{adjustbox}
\end{table}

The available level-specific Qwen3.5 intervals are retained in
Table~\ref{tab:sc_e0_level_intervals}. Each level uses its own input, pair,
and eligible-world denominator. These counts are not summed to obtain a
new total number of independent worlds.

% Source: supplied evidence report, lines 85--96.
\FloatBarrier
\begingroup
\footnotesize
\setlength{\tabcolsep}{3pt}
\renewcommand{\arraystretch}{1.08}
\begin{longtable}{llrL{0.25\linewidth}rL{0.25\linewidth}}
\caption{Level-specific counts and 95\% intervals for the three diagnostic-family checkpoints. Estimates and intervals are in percent.}\label{tab:sc_e0_level_intervals}\\
\toprule
Qwen3.5 & Level & Claim correct/N & Claim [CI] & Pairs correct/N & Pair [CI] \\
\midrule
\endfirsthead
\multicolumn{6}{l}{\footnotesize Table~\thetable\ continued.}\\
\toprule
Qwen3.5 & Level & Claim correct/N & Claim [CI] & Pairs correct/N & Pair [CI] \\
\midrule
\endhead
\midrule\multicolumn{6}{r}{\footnotesize Continued on the next page.}\\
\endfoot
\bottomrule
\endlastfoot
4B & L1 & 8730/16598 & 52.60\;[51.71, 53.45] & 2610/7559 & 34.53\;[33.45, 35.64] \\
4B & L2 & 2028/2624 & 83.66\;[80.88, 86.40] & 863/1212 & 71.20\;[66.56, 75.92] \\
4B & L3 & 1225/2602 & 47.08\;[45.07, 49.04] & 214/1041 & 20.56\;[17.82, 23.44] \\
4B & L4 & 1522/2572 & 59.18\;[57.70, 60.61] & 325/1136 & 28.61\;[25.94, 31.36] \\
9B & L1 & 9204/16598 & 55.45\;[54.57, 56.30] & 2954/7559 & 39.08\;[38.01, 40.16] \\
9B & L2 & 2177/2624 & 89.81\;[87.56, 91.97] & 1025/1212 & 84.57\;[81.49, 87.59] \\
9B & L3 & 1396/2602 & 53.65\;[51.32, 56.02] & 412/1041 & 39.58\;[36.26, 42.69] \\
9B & L4 & 1562/2572 & 60.73\;[59.13, 62.25] & 350/1136 & 30.81\;[27.97, 33.62] \\
27B & L1 & 9729/16598 & 58.62\;[57.70, 59.52] & 3508/7559 & 46.41\;[45.27, 47.54] \\
27B & L2 & 2223/2624 & 91.71\;[89.44, 93.79] & 1063/1212 & 87.71\;[84.92, 90.42] \\
27B & L3 & 1505/2602 & 57.84\;[55.37, 60.29] & 515/1041 & 49.47\;[46.09, 52.68] \\
27B & L4 & 1673/2572 & 65.05\;[63.21, 66.90] & 501/1136 & 44.10\;[40.98, 47.28] \\
\end{longtable}
\endgroup

Table~\ref{tab:sc_test_full} separates the existing test-only subset from
the all-split table. Test-only ClaimAcc is not strictly increasing across
Qwen3.5 sizes, whereas PairAcc increases on this particular subset. This
is a descriptive comparison of fixed checkpoints and evaluation conditions,
not a claim that parameter count causally determines state utilization.

% Source: supplied evidence report, lines 107--128.
\begin{table}[htbp]
\centering
\caption{Existing test-only results. ClaimAcc uses 5,608 inputs over 1,162 worlds; PairAcc uses 1,771 pairs over 1,114 eligible worlds. A dash denotes an unavailable archived confidence interval.}
\label{tab:sc_test_full}
\small
\setlength{\tabcolsep}{4pt}
\begin{adjustbox}{max width=\linewidth}
\begin{tabular}{lrL{0.22\linewidth}rL{0.22\linewidth}}
\toprule
Model & Claim correct/N & Claim [95\% CI] & Pairs correct/N & Pair [95\% CI] \\
\midrule
Qwen2.5-VL-7B & 2421/5608 & 43.17\;\textemdash & 27/1771 & 1.52\;\textemdash \\
Qwen2.5-VL-32B & 3011/5608 & 53.69\;\textemdash & 566/1771 & 31.96\;\textemdash \\
Qwen3-VL-2B-Thinking & 1143/5608 & 20.38\;\textemdash & 277/1771 & 15.64\;\textemdash \\
Qwen3-VL-8B-Instruct & 3098/5608 & 55.24\;\textemdash & 539/1771 & 30.43\;\textemdash \\
InternVL3.5-8B & 2864/5608 & 51.07\;\textemdash & 584/1771 & 32.98\;\textemdash \\
InternVL3.5-14B & 3087/5608 & 55.05\;\textemdash & 784/1771 & 44.27\;\textemdash \\
MiMo-VL-7B-RL & 2787/5608 & 49.70\;\textemdash & 403/1771 & 22.76\;\textemdash \\
Qwen3.5-4B & 3099/5608 & 55.26\;[53.80, 56.71] & 694/1771 & 39.19\;[37.01, 41.44] \\
Qwen3.5-9B & 3169/5608 & 56.51\;[55.10, 57.95] & 787/1771 & 44.44\;[42.23, 46.68] \\
Qwen3.5-27B & 3143/5608 & 56.04\;[54.65, 57.48] & 902/1771 & 50.93\;[48.79, 53.06] \\
\bottomrule
\end{tabular}
\end{adjustbox}
\end{table}

\FloatBarrier
\section{Behavioral bridge: conditions and complete results}
\label{app:sc_behavior}

Table~\ref{tab:sc_conditions} defines the diagnostic condition identifiers
used throughout Section~\ref{sec:sc_results}. These are interfaces to a fixed
state variable, not labels for observed internal computations. B00 is a
historical reused condition whose exact request is not present in the supplied
batch; its score is retained without reconstructing a prompt. The literal
texts for the remaining conditions are in Appendix~\ref{app:sc_prompts_count}.

% Source: supplied evidence report, lines 273--293.
\FloatBarrier
\begingroup
\footnotesize
\setlength{\tabcolsep}{3pt}
\renewcommand{\arraystretch}{1.08}
\begin{longtable}{L{0.28\linewidth}L{0.66\linewidth}}
\caption{Condition registry for the count diagnostics. The sample panel has 160 sequence-level entries per standard condition and an 80-world shared initial-protection supplement. B07 is not a matched control for B05.}\label{tab:sc_conditions}\\
\toprule
ID & Actual question or intervention \\
\midrule
\endfirsthead
\multicolumn{2}{l}{\footnotesize Table~\thetable\ continued.}\\
\toprule
ID & Actual question or intervention \\
\midrule
\endhead
\midrule\multicolumn{2}{r}{\footnotesize Continued on the next page.}\\
\endfoot
\bottomrule
\endlastfoot
\texttt{B00} & Historical full-task request (reused output) \\
\texttt{B01} & Canonical cumulative two-action visual task; query $s_2$ \\
\texttt{B02} & Independent visual initial-fact query; query $s_0$ \\
\texttt{B03} & Both actions retained; query $s_1$ \\
\texttt{B04} & Ask which named state is the input to action 2 \\
\texttt{B05} & B01 plus the correct initial count before both actions \\
\texttt{B06} & B01 plus the correct $s_1$ checkpoint after action 1 \\
\texttt{B07} & B01 plus an unrelated register carrying $s_1$ after action 1 \\
\texttt{B08} & Same cumulative count program with explicit $s_0$, no images \\
\texttt{B09} & Explicit $s_1$ and only the second action, no images \\
\texttt{B10} & Complete oracle state table; query $s_0$ \\
\texttt{B11} & Complete oracle state table; query $s_1$ \\
\texttt{B12} & Complete oracle state table; query $s_2$ \\
\texttt{B13} & Separate counterfactual symbolic suffix input (lower-valued instance) \\
\texttt{B14} & Separate counterfactual symbolic suffix input (higher-valued instance) \\
\texttt{B15} & Both actions retained; query the original $s_0$ \\
\texttt{B16} & Both actions retained; query an unchanged protected category \\
\texttt{B17} & Execute only action 1; query $s_1$ \\
\texttt{PROTECTED\_PRE\_SUPPLEMENT} & Independent initial query for the protected category \\
\end{longtable}
\endgroup

Tables~\ref{tab:sc_b1_4b}, \ref{tab:sc_b1_9b}, and~\ref{tab:sc_b1_27b}
report all archived B1 conditions for the three checkpoints. The 4B table
shows both substantial explicit-state improvements and remaining failures;
9B and 27B reach the panel ceiling for explicit initial state and symbolic
programs. However, state-table queries and protected-fact queries are not
uniformly correct. In particular, protection performance is already imperfect
at baseline, so a low post-action protection score alone cannot be read as
damage caused by updating the target.

% Source: supplied evidence report, lines 273--293.
\FloatBarrier
\begingroup
\footnotesize
\setlength{\tabcolsep}{3pt}
\renewcommand{\arraystretch}{1.08}
\begin{longtable}{L{0.36\linewidth}rrlr}
\caption{Complete Qwen3.5-4B behavioral bridge. Accuracy and 95\% world-cluster intervals are in percent. Counts follow the original condition-level scoring, including null/invalid decisions.}\label{tab:sc_b1_4b}\\
\toprule
Condition & Correct/N & Accuracy & 95\% CI & Worlds \\
\midrule
\endfirsthead
\multicolumn{5}{l}{\footnotesize Table~\thetable\ continued.}\\
\toprule
Condition & Correct/N & Accuracy & 95\% CI & Worlds \\
\midrule
\endhead
\midrule\multicolumn{5}{r}{\footnotesize Continued on the next page.}\\
\endfoot
\bottomrule
\endlastfoot
\texttt{B00} & 47/160 & 29.38 & [21.88, 37.50] & 80 \\
\texttt{B01} & 44/160 & 27.50 & [21.88, 33.12] & 80 \\
\texttt{B02} & 104/160 & 65.00 & [53.75, 75.00] & 80 \\
\texttt{B03} & 67/160 & 41.88 & [32.50, 51.25] & 80 \\
\texttt{B04} & 25/160 & 15.62 & [10.62, 21.25] & 80 \\
\texttt{B05} & 103/160 & 64.38 & [56.25, 72.50] & 80 \\
\texttt{B06} & 123/160 & 76.88 & [70.62, 83.12] & 80 \\
\texttt{B07} & 42/160 & 26.25 & [20.00, 33.75] & 80 \\
\texttt{B08} & 140/160 & 87.50 & [83.12, 91.88] & 80 \\
\texttt{B09} & 160/160 & 100.00 & [100.00, 100.00] & 80 \\
\texttt{B10} & 160/160 & 100.00 & [100.00, 100.00] & 80 \\
\texttt{B11} & 157/160 & 98.12 & [95.00, 100.00] & 80 \\
\texttt{B12} & 159/160 & 99.38 & [98.12, 100.00] & 80 \\
\texttt{B13} & 160/160 & 100.00 & [100.00, 100.00] & 80 \\
\texttt{B14} & 160/160 & 100.00 & [100.00, 100.00] & 80 \\
\texttt{B15} & 103/160 & 64.38 & [53.75, 75.00] & 80 \\
\texttt{B16} & 34/160 & 21.25 & [13.12, 30.00] & 80 \\
\texttt{B17} & 84/160 & 52.50 & [41.25, 63.75] & 80 \\
\texttt{PROTECTED\_PRE\_SUPPLEMENT} & 40/80 & 50.00 & [38.75, 61.25] & 80 \\
\end{longtable}
\endgroup

% Source: supplied evidence report, lines 273--293.
\FloatBarrier
\begingroup
\footnotesize
\setlength{\tabcolsep}{3pt}
\renewcommand{\arraystretch}{1.08}
\begin{longtable}{L{0.36\linewidth}rrlr}
\caption{Complete Qwen3.5-9B behavioral bridge. Accuracy and 95\% world-cluster intervals are in percent. Counts follow the original condition-level scoring, including null/invalid decisions.}\label{tab:sc_b1_9b}\\
\toprule
Condition & Correct/N & Accuracy & 95\% CI & Worlds \\
\midrule
\endfirsthead
\multicolumn{5}{l}{\footnotesize Table~\thetable\ continued.}\\
\toprule
Condition & Correct/N & Accuracy & 95\% CI & Worlds \\
\midrule
\endhead
\midrule\multicolumn{5}{r}{\footnotesize Continued on the next page.}\\
\endfoot
\bottomrule
\endlastfoot
\texttt{B00} & 53/160 & 33.12 & [26.25, 40.62] & 80 \\
\texttt{B01} & 56/160 & 35.00 & [26.88, 43.12] & 80 \\
\texttt{B02} & 100/160 & 62.50 & [51.25, 72.50] & 80 \\
\texttt{B03} & 46/160 & 28.75 & [20.62, 36.88] & 80 \\
\texttt{B04} & 160/160 & 100.00 & [100.00, 100.00] & 80 \\
\texttt{B05} & 160/160 & 100.00 & [100.00, 100.00] & 80 \\
\texttt{B06} & 104/160 & 65.00 & [56.88, 73.75] & 80 \\
\texttt{B07} & 48/160 & 30.00 & [22.50, 38.12] & 80 \\
\texttt{B08} & 160/160 & 100.00 & [100.00, 100.00] & 80 \\
\texttt{B09} & 160/160 & 100.00 & [100.00, 100.00] & 80 \\
\texttt{B10} & 154/160 & 96.25 & [91.88, 99.38] & 80 \\
\texttt{B11} & 159/160 & 99.38 & [98.12, 100.00] & 80 \\
\texttt{B12} & 150/160 & 93.75 & [90.00, 96.88] & 80 \\
\texttt{B13} & 160/160 & 100.00 & [100.00, 100.00] & 80 \\
\texttt{B14} & 160/160 & 100.00 & [100.00, 100.00] & 80 \\
\texttt{B15} & 99/160 & 61.88 & [51.25, 71.88] & 80 \\
\texttt{B16} & 49/160 & 30.63 & [21.25, 40.00] & 80 \\
\texttt{B17} & 80/160 & 50.00 & [38.75, 61.25] & 80 \\
\texttt{PROTECTED\_PRE\_SUPPLEMENT} & 29/80 & 36.25 & [26.25, 46.25] & 80 \\
\end{longtable}
\endgroup

% Source: supplied evidence report, lines 273--293.
\FloatBarrier
\begingroup
\footnotesize
\setlength{\tabcolsep}{3pt}
\renewcommand{\arraystretch}{1.08}
\begin{longtable}{L{0.36\linewidth}rrlr}
\caption{Complete Qwen3.5-27B behavioral bridge. Accuracy and 95\% world-cluster intervals are in percent. Counts follow the original condition-level scoring, including null/invalid decisions.}\label{tab:sc_b1_27b}\\
\toprule
Condition & Correct/N & Accuracy & 95\% CI & Worlds \\
\midrule
\endfirsthead
\multicolumn{5}{l}{\footnotesize Table~\thetable\ continued.}\\
\toprule
Condition & Correct/N & Accuracy & 95\% CI & Worlds \\
\midrule
\endhead
\midrule\multicolumn{5}{r}{\footnotesize Continued on the next page.}\\
\endfoot
\bottomrule
\endlastfoot
\texttt{B00} & 97/160 & 60.62 & [51.25, 70.00] & 80 \\
\texttt{B01} & 94/160 & 58.75 & [48.75, 68.75] & 80 \\
\texttt{B02} & 96/160 & 60.00 & [50.00, 70.00] & 80 \\
\texttt{B03} & 111/160 & 69.38 & [59.98, 78.77] & 80 \\
\texttt{B04} & 160/160 & 100.00 & [100.00, 100.00] & 80 \\
\texttt{B05} & 160/160 & 100.00 & [100.00, 100.00] & 80 \\
\texttt{B06} & 138/160 & 86.25 & [80.00, 91.88] & 80 \\
\texttt{B07} & 95/160 & 59.38 & [49.38, 68.75] & 80 \\
\texttt{B08} & 160/160 & 100.00 & [100.00, 100.00] & 80 \\
\texttt{B09} & 160/160 & 100.00 & [100.00, 100.00] & 80 \\
\texttt{B10} & 160/160 & 100.00 & [100.00, 100.00] & 80 \\
\texttt{B11} & 160/160 & 100.00 & [100.00, 100.00] & 80 \\
\texttt{B12} & 158/160 & 98.75 & [96.88, 100.00] & 80 \\
\texttt{B13} & 160/160 & 100.00 & [100.00, 100.00] & 80 \\
\texttt{B14} & 160/160 & 100.00 & [100.00, 100.00] & 80 \\
\texttt{B15} & 97/160 & 60.62 & [50.00, 71.25] & 80 \\
\texttt{B16} & 29/160 & 18.12 & [10.62, 26.25] & 80 \\
\texttt{B17} & 102/160 & 63.75 & [52.50, 73.75] & 80 \\
\texttt{PROTECTED\_PRE\_SUPPLEMENT} & 39/80 & 48.75 & [37.50, 60.00] & 80 \\
\end{longtable}
\endgroup

Table~\ref{tab:sc_b1_paired} provides the matched rescue/harm counts behind
the bridge comparisons. Explicit $s_0$ repairs every B01 error for 9B and 27B
in this panel, but not for 4B; the 4B condition also changes some originally
correct answers to incorrect ones. B06 has harmful transitions in all three
models. These observations motivate reporting net change and both conditional
event counts, rather than only the accuracy of the modified prompt.

% Source: supplied evidence report, lines 308--324.
\FloatBarrier
\begingroup
\footnotesize
\setlength{\tabcolsep}{3pt}
\renewcommand{\arraystretch}{1.08}
\begin{longtable}{llrrrl}
\caption{Paired changes relative to B01. Every row has 160 matched sequences over 80 worlds. Rescue is divided by baseline errors; harm is divided by baseline successes. Net differences and intervals are percentage points.}\label{tab:sc_b1_paired}\\
\toprule
Qwen3.5 & Contrast & Rescue/errors & Harm/successes & Net $\Delta$ & 95\% CI \\
\midrule
\endfirsthead
\multicolumn{6}{l}{\footnotesize Table~\thetable\ continued.}\\
\toprule
Qwen3.5 & Contrast & Rescue/errors & Harm/successes & Net $\Delta$ & 95\% CI \\
\midrule
\endhead
\midrule\multicolumn{6}{r}{\footnotesize Continued on the next page.}\\
\endfoot
\bottomrule
\endlastfoot
4B & B05--B01 & 74/116 & 15/44 & 36.88 & [25.62, 47.50] \\
4B & B06--B01 & 86/116 & 7/44 & 49.38 & [42.50, 56.25] \\
4B & B07--B01 & 5/116 & 7/44 & -1.25 & [-5.62, 3.12] \\
4B & B08--B01 & 96/116 & 0/44 & 60.00 & [53.75, 66.25] \\
4B & B09--B01 & 116/116 & 0/44 & 72.50 & [66.88, 78.12] \\
9B & B05--B01 & 104/104 & 0/56 & 65.00 & [56.88, 73.12] \\
9B & B06--B01 & 55/104 & 7/56 & 30.00 & [21.25, 38.75] \\
9B & B07--B01 & 5/104 & 13/56 & -5.00 & [-10.00, 0.62] \\
9B & B08--B01 & 104/104 & 0/56 & 65.00 & [56.88, 73.12] \\
9B & B09--B01 & 104/104 & 0/56 & 65.00 & [56.88, 72.50] \\
27B & B05--B01 & 66/66 & 0/94 & 41.25 & [31.25, 51.25] \\
27B & B06--B01 & 46/66 & 2/94 & 27.50 & [18.75, 36.25] \\
27B & B07--B01 & 4/66 & 3/94 & 0.62 & [-2.50, 3.75] \\
27B & B08--B01 & 66/66 & 0/94 & 41.25 & [31.88, 51.25] \\
27B & B09--B01 & 66/66 & 0/94 & 41.25 & [31.25, 51.25] \\
\end{longtable}
\endgroup

The count register comparison needs a separate qualification. B05 places
$s_0$ before the program, whereas legacy B07 places $s_1$ after action 1.
Table~\ref{tab:sc_legacy_sham} therefore reports their differences only as
post-hoc descriptive context contrasts. They cannot establish the effect of
information role while holding the state value and position fixed. The new
$s_0$ sham used by I1 is a different intervention condition
(Appendix~\ref{app:sc_i1}); the matched behavioral relation control is the
non-count experiment (Appendix~\ref{app:sc_noncount}).

% Source: supplied evidence report, lines 1899--1906.
\begin{table}[htbp]
\centering
\caption{Legacy, unmatched B05--B07 count contrast. These post-hoc descriptive differences do not provide a role-specific S0 control. Each contrast contains 160 sequences and 80 worlds.}
\label{tab:sc_legacy_sham}
\small
\setlength{\tabcolsep}{4pt}
\begin{adjustbox}{max width=\linewidth}
\begin{tabular}{lrrr}
\toprule
Qwen3.5 & Net count/N & $\Delta$ (pp) & 95\% CI \\
\midrule
4B & 61/160 & 38.12 & [26.88, 48.75] \\
9B & 112/160 & 70.00 & [61.88, 77.50] \\
27B & 65/160 & 40.62 & [30.62, 50.62] \\
\bottomrule
\end{tabular}
\end{adjustbox}
\end{table}

\FloatBarrier
\section{Case definitions and numerical error patterns}
\label{app:sc_cases}

Table~\ref{tab:sc_case_types} retains all five behavioral categories. Given
returned records with Boolean correctness, A and B distinguish oracle-rescued
full-task failures by whether the direct fact is correct; C and D distinguish
full-task successes by the same direct-fact criterion; E collects non-rescued
full-task failures. They form a partition of the present sequence records,
but their distinct-world sets can overlap. Returned null or invalid answers
may enter a behavioral error category, whereas the primary intervention
analysis applies its separate interface eligibility filter.

% Source: supplied evidence report, lines 238--242.
\begin{table}[htbp]
\centering
\caption{Behavioral categories, as sequence count/distinct affected worlds. A: direct correct, full wrong, oracle correct. B: direct wrong, full wrong, oracle correct. C: direct and full correct. D: direct wrong, full correct. E: full and oracle wrong.}
\label{tab:sc_case_types}
\small
\setlength{\tabcolsep}{4pt}
\begin{adjustbox}{max width=\linewidth}
\begin{tabular}{lrrrrr}
\toprule
Qwen3.5 & A & B & C & D & E \\
\midrule
4B & 35/30 & 39/26 & 40/39 & 4/4 & 42/40 \\
9B & 50/37 & 54/30 & 50/37 & 6/6 & 0/0 \\
27B & 11/8 & 55/31 & 85/45 & 9/8 & 0/0 \\
\bottomrule
\end{tabular}
\end{adjustbox}
\end{table}

Table~\ref{tab:sc_conditional_full} expands the conditional estimates in
Table~\ref{tab:sc_main_gap}. The relevant rescue denominator is the entire
direct-correct/full-wrong cohort, including non-rescued cases. For 4B this
yields 35/64, not the tautological 35/35 obtained by conditioning on Type A.
The world count is specific to the metric's eligible cohort: for example,
9B's first conditional rate uses 50 direct-correct worlds, while its rescue
rate uses the 37 worlds containing eligible full-task errors. These counts
must not be substituted for one another.

% Source: supplied evidence report, lines 250--260.
\FloatBarrier
\begingroup
\footnotesize
\setlength{\tabcolsep}{3pt}
\renewcommand{\arraystretch}{1.08}
\begin{longtable}{lL{0.40\linewidth}rrrl}
\caption{Complete archived conditional diagnostics. Percent estimates and intervals are preserved from the conditional-statistics artifact. The last condition concerns an independently queried intermediate state, not an internal trajectory.}\label{tab:sc_conditional_full}\\
\toprule
Qwen3.5 & Conditional quantity & Events/eligible & Worlds & Rate & 95\% CI \\
\midrule
\endfirsthead
\multicolumn{6}{l}{\footnotesize Table~\thetable\ continued.}\\
\toprule
Qwen3.5 & Conditional quantity & Events/eligible & Worlds & Rate & 95\% CI \\
\midrule
\endhead
\midrule\multicolumn{6}{r}{\footnotesize Continued on the next page.}\\
\endfoot
\bottomrule
\endlastfoot
4B & $\Pr(T=0\mid F=1)$ & 64/104 & 52 & 61.54 & [55.77, 68.27] \\
4B & $\Pr(O=1\mid F=1,T=0)$ & 35/64 & 51 & 54.69 & [42.19, 66.67] \\
4B & $\Pr(\mathrm{B06}=1\mid\mathrm{B03}=1,T=0)$ & 35/44 & 34 & 79.55 & [65.85, 91.11] \\
9B & $\Pr(T=0\mid F=1)$ & 50/100 & 50 & 50.00 & [40.00, 60.00] \\
9B & $\Pr(O=1\mid F=1,T=0)$ & 50/50 & 37 & 100.00 & [100.00, 100.00] \\
9B & $\Pr(\mathrm{B06}=1\mid\mathrm{B03}=1,T=0)$ & 14/23 & 23 & 60.87 & [39.13, 78.26] \\
27B & $\Pr(T=0\mid F=1)$ & 11/96 & 48 & 11.46 & [4.17, 19.79] \\
27B & $\Pr(O=1\mid F=1,T=0)$ & 11/11 & 8 & 100.00 & [100.00, 100.00] \\
27B & $\Pr(\mathrm{B06}=1\mid\mathrm{B03}=1,T=0)$ & 17/19 & 15 & 89.47 & [75.00, 100.00] \\
\end{longtable}
\endgroup

The exclusive numerical classifications in
Tables~\ref{tab:sc_error_b01} and~\ref{tab:sc_error_b06} describe returned
values, not identified algorithms. An ambiguous label is used whenever
multiple signatures yield the same number. For example, $s_1$ and the legacy
sham value are deliberately equal; a zero in the \emph{exclusive} $s_1$
category consequently does not mean the model never outputs $s_1$.
Table~\ref{tab:sc_error_signatures} preserves the corresponding non-exclusive
matches.

% Source: supplied evidence report, lines 339--343.
\begin{table}[htbp]
\centering
\caption{B01 exclusive error patterns. Fractions use the number of errors for that model as denominator, except the first row. The ambiguity category prevents numerical coincidences from being assigned to a unique internal state.}
\label{tab:sc_error_b01}
\small
\setlength{\tabcolsep}{4pt}
\begin{adjustbox}{max width=\linewidth}
\begin{tabular}{lrrr}
\toprule
Pattern & 4B & 9B & 27B \\
\midrule
All errors / sequences & 116/160 & 104/160 & 66/160 \\
Worlds with at least one error & 79 & 67 & 39 \\
$s_0$ & 7/116 & 6/104 & 0/66 \\
$s_1$ & 0/116 & 0/104 & 0/66 \\
Action-1 amount & 3/116 & 1/104 & 0/66 \\
Action-2 amount & 3/116 & 9/104 & 9/66 \\
$\tau_2(s_0)$ & 10/116 & 5/104 & 10/66 \\
$\tau_1(s_1)$ & 6/116 & 12/104 & 8/66 \\
Other integer & 27/116 & 15/104 & 11/66 \\
Null & 3/116 & 18/104 & 3/66 \\
Invalid & 0/116 & 0/104 & 0/66 \\
Ambiguous numerical pattern & 57/116 & 38/104 & 25/66 \\
\bottomrule
\end{tabular}
\end{adjustbox}
\end{table}

% Source: supplied evidence report, lines 370--374.
\begin{table}[htbp]
\centering
\caption{B06 exclusive error patterns. Fractions use the number of errors for that model as denominator, except the first row. The ambiguity category prevents numerical coincidences from being assigned to a unique internal state.}
\label{tab:sc_error_b06}
\small
\setlength{\tabcolsep}{4pt}
\begin{adjustbox}{max width=\linewidth}
\begin{tabular}{lrrr}
\toprule
Pattern & 4B & 9B & 27B \\
\midrule
All errors / sequences & 37/160 & 56/160 & 22/160 \\
Worlds with at least one error & 33 & 40 & 18 \\
$s_0$ & 0/37 & 0/56 & 0/22 \\
$s_1$ & 0/37 & 0/56 & 0/22 \\
Action-1 amount & 0/37 & 0/56 & 0/22 \\
Action-2 amount & 0/37 & 11/56 & 0/22 \\
$\tau_2(s_0)$ & 1/37 & 0/56 & 3/22 \\
$\tau_1(s_1)$ & 20/37 & 7/56 & 0/22 \\
Other integer & 0/37 & 18/56 & 5/22 \\
Null & 0/37 & 1/56 & 0/22 \\
Invalid & 0/37 & 0/56 & 0/22 \\
Ambiguous numerical pattern & 16/37 & 19/56 & 14/22 \\
\bottomrule
\end{tabular}
\end{adjustbox}
\end{table}

% Source: supplied evidence report, lines 346--365.
\FloatBarrier
\begingroup
\footnotesize
\setlength{\tabcolsep}{3pt}
\renewcommand{\arraystretch}{1.08}
\begin{longtable}{llL{0.21\linewidth}L{0.21\linewidth}L{0.21\linewidth}}
\caption{Non-exclusive numerical matches among errors. Each cell gives matches/errors and the number of affected worlds. Multiple signatures can match the same output and the rows must not be summed.}\label{tab:sc_error_signatures}\\
\toprule
Condition & Signature & 4B & 9B & 27B \\
\midrule
\endfirsthead
\multicolumn{5}{l}{\footnotesize Table~\thetable\ continued.}\\
\toprule
Condition & Signature & 4B & 9B & 27B \\
\midrule
\endhead
\midrule\multicolumn{5}{r}{\footnotesize Continued on the next page.}\\
\endfoot
\bottomrule
\endlastfoot
B01 & $s_0$ & 31/116; W=31 & 17/104; W=17 & 1/66; W=1 \\
B01 & $s_1$ & 35/116; W=31 & 17/104; W=17 & 19/66; W=19 \\
B01 & Action-1 amount & 30/116; W=27 & 17/104; W=16 & 11/66; W=11 \\
B01 & Action-2 amount & 21/116; W=20 & 33/104; W=30 & 22/66; W=15 \\
B01 & $\tau_2(s_0)$ & 10/116; W=10 & 5/104; W=5 & 10/66; W=6 \\
B01 & $\tau_1(s_1)$ & 17/116; W=16 & 21/104; W=21 & 9/66; W=9 \\
B06 & $s_0$ & 0/37; W=0 & 0/56; W=0 & 0/22; W=0 \\
B06 & $s_1$ & 16/37; W=14 & 18/56; W=18 & 14/22; W=14 \\
B06 & Action-1 amount & 2/37; W=2 & 6/56; W=6 & 6/22; W=6 \\
B06 & Action-2 amount & 1/37; W=1 & 17/56; W=12 & 6/22; W=6 \\
B06 & $\tau_2(s_0)$ & 1/37; W=1 & 0/56; W=0 & 3/22; W=3 \\
B06 & $\tau_1(s_1)$ & 20/37; W=20 & 7/56; W=7 & 0/22; W=0 \\
\end{longtable}
\endgroup

The sizeable ambiguity and other-integer categories prevent a one-to-one
mapping from these tables to ``stale state,'' ``wrong branch,'' or ``repeated
action'' mechanisms. Similarly, when inverse actions return the state to its
initial value, an answer equal to $s_0$ is not evidence that the actions were
ignored. These numerical patterns guide description but do not resolve
internal causal alternatives.

\FloatBarrier
\section{Target-semantic controls}
\label{app:sc_target}

The target experiment fixes 96 symbolic programs from 16 families and crosses
four wordings with three target states, for 1,152 requests per model and 3,456
in total. T1 uses a state index, T2 states the temporal target in words, T3
defines all states, and T4 adds a final target reminder. The literal versions
appear in Appendix~\ref{app:sc_prompts_target}.

Table~\ref{tab:sc_w1_full} reports every target and wording rather than only
the S1 repair. All explicit alternatives repair S0 and S1 in this panel, but
9B's S2 accuracy is lower under the longer definition/reminder prompts.
Thus, the experiment identifies strong target-wording sensitivity, not a
universal improvement from longer instructions. The archived W1 index-only
S2 score for 9B is 92/96; it is the value used here, without substituting an
earlier symbolic-run summary.

% Source: supplied evidence report, lines 409--419.
\FloatBarrier
\begingroup
\footnotesize
\setlength{\tabcolsep}{3pt}
\renewcommand{\arraystretch}{1.08}
\begin{longtable}{lllrrl}
\caption{Complete target-by-wording results. Every row contains 96 programs and 16 program-family clusters; accuracy and confidence intervals are percentages. Extreme empirical intervals are not population guarantees.}\label{tab:sc_w1_full}\\
\toprule
Qwen3.5 & Target & Wording & Correct/N & Accuracy & 95\% CI \\
\midrule
\endfirsthead
\multicolumn{6}{l}{\footnotesize Table~\thetable\ continued.}\\
\toprule
Qwen3.5 & Target & Wording & Correct/N & Accuracy & 95\% CI \\
\midrule
\endhead
\midrule\multicolumn{6}{r}{\footnotesize Continued on the next page.}\\
\endfoot
\bottomrule
\endlastfoot
4B & S0 & T1 & 15/96 & 15.62 & [12.50, 18.75] \\
4B & S0 & T2 & 96/96 & 100.00 & [100.00, 100.00] \\
4B & S0 & T3 & 96/96 & 100.00 & [100.00, 100.00] \\
4B & S0 & T4 & 96/96 & 100.00 & [100.00, 100.00] \\
4B & S1 & T1 & 37/96 & 38.54 & [31.25, 44.79] \\
4B & S1 & T2 & 96/96 & 100.00 & [100.00, 100.00] \\
4B & S1 & T3 & 96/96 & 100.00 & [100.00, 100.00] \\
4B & S1 & T4 & 96/96 & 100.00 & [100.00, 100.00] \\
4B & S2 & T1 & 68/96 & 70.83 & [62.50, 79.17] \\
4B & S2 & T2 & 69/96 & 71.88 & [65.62, 78.12] \\
4B & S2 & T3 & 66/96 & 68.75 & [61.46, 76.04] \\
4B & S2 & T4 & 66/96 & 68.75 & [61.46, 76.04] \\
9B & S0 & T1 & 47/96 & 48.96 & [45.83, 52.08] \\
9B & S0 & T2 & 96/96 & 100.00 & [100.00, 100.00] \\
9B & S0 & T3 & 96/96 & 100.00 & [100.00, 100.00] \\
9B & S0 & T4 & 96/96 & 100.00 & [100.00, 100.00] \\
9B & S1 & T1 & 0/96 & 0.00 & [0.00, 0.00] \\
9B & S1 & T2 & 96/96 & 100.00 & [100.00, 100.00] \\
9B & S1 & T3 & 96/96 & 100.00 & [100.00, 100.00] \\
9B & S1 & T4 & 96/96 & 100.00 & [100.00, 100.00] \\
9B & S2 & T1 & 92/96 & 95.83 & [92.71, 98.96] \\
9B & S2 & T2 & 88/96 & 91.67 & [87.50, 95.83] \\
9B & S2 & T3 & 75/96 & 78.12 & [72.92, 83.33] \\
9B & S2 & T4 & 74/96 & 77.08 & [71.88, 83.33] \\
27B & S0 & T1 & 33/96 & 34.38 & [33.33, 36.46] \\
27B & S0 & T2 & 96/96 & 100.00 & [100.00, 100.00] \\
27B & S0 & T3 & 96/96 & 100.00 & [100.00, 100.00] \\
27B & S0 & T4 & 96/96 & 100.00 & [100.00, 100.00] \\
27B & S1 & T1 & 0/96 & 0.00 & [0.00, 0.00] \\
27B & S1 & T2 & 96/96 & 100.00 & [100.00, 100.00] \\
27B & S1 & T3 & 96/96 & 100.00 & [100.00, 100.00] \\
27B & S1 & T4 & 96/96 & 100.00 & [100.00, 100.00] \\
27B & S2 & T1 & 96/96 & 100.00 & [100.00, 100.00] \\
27B & S2 & T2 & 96/96 & 100.00 & [100.00, 100.00] \\
27B & S2 & T3 & 96/96 & 100.00 & [100.00, 100.00] \\
27B & S2 & T4 & 96/96 & 100.00 & [100.00, 100.00] \\
\end{longtable}
\endgroup

Table~\ref{tab:sc_w1_paired} gives the archived paired changes relative to
T1. The large S1 effects coexist with negative S2 effects for 9B, including
-18.75 points for T4--T1. This counterevidence prevents describing semantic
clarification as a general repair of all state reasoning. All programs in the
reported non-degenerate subset satisfy $s_1\ne s_2$, so restricting to that
subset leaves the 96-program scores unchanged; $s_0$ may still coincide with
$s_2$.

% Source: supplied evidence report, lines 524--552.
\FloatBarrier
\begingroup
\footnotesize
\setlength{\tabcolsep}{3pt}
\renewcommand{\arraystretch}{1.08}
\begin{longtable}{lllrl}
\caption{Paired wording contrasts. Every comparison has 96 program pairs and 16 families. Differences and intervals are percentage points; all targets are retained.}\label{tab:sc_w1_paired}\\
\toprule
Qwen3.5 & Target & Contrast & $\Delta$ (pp) & 95\% CI \\
\midrule
\endfirsthead
\multicolumn{5}{l}{\footnotesize Table~\thetable\ continued.}\\
\toprule
Qwen3.5 & Target & Contrast & $\Delta$ (pp) & 95\% CI \\
\midrule
\endhead
\midrule\multicolumn{5}{r}{\footnotesize Continued on the next page.}\\
\endfoot
\bottomrule
\endlastfoot
4B & S0 & T2--T1 & 84.38 & [81.25, 87.50] \\
4B & S0 & T3--T1 & 84.38 & [81.25, 87.50] \\
4B & S0 & T4--T1 & 84.38 & [81.25, 87.50] \\
4B & S1 & T2--T1 & 61.46 & [55.21, 68.75] \\
4B & S1 & T3--T1 & 61.46 & [55.21, 68.75] \\
4B & S1 & T4--T1 & 61.46 & [55.21, 68.75] \\
4B & S2 & T2--T1 & 1.04 & [-5.21, 8.33] \\
4B & S2 & T3--T1 & -2.08 & [-7.29, 3.12] \\
4B & S2 & T4--T1 & -2.08 & [-7.29, 3.12] \\
9B & S0 & T2--T1 & 51.04 & [47.92, 54.17] \\
9B & S0 & T3--T1 & 51.04 & [47.92, 54.17] \\
9B & S0 & T4--T1 & 51.04 & [47.92, 54.17] \\
9B & S1 & T2--T1 & 100.00 & [100.00, 100.00] \\
9B & S1 & T3--T1 & 100.00 & [100.00, 100.00] \\
9B & S1 & T4--T1 & 100.00 & [100.00, 100.00] \\
9B & S2 & T2--T1 & -4.17 & [-8.33, 0.00] \\
9B & S2 & T3--T1 & -17.71 & [-22.92, -12.50] \\
9B & S2 & T4--T1 & -18.75 & [-23.96, -13.54] \\
27B & S0 & T2--T1 & 65.62 & [63.54, 66.67] \\
27B & S0 & T3--T1 & 65.62 & [63.54, 66.67] \\
27B & S0 & T4--T1 & 65.62 & [63.54, 66.67] \\
27B & S1 & T2--T1 & 100.00 & [100.00, 100.00] \\
27B & S1 & T3--T1 & 100.00 & [100.00, 100.00] \\
27B & S1 & T4--T1 & 100.00 & [100.00, 100.00] \\
27B & S2 & T2--T1 & 0.00 & [0.00, 0.00] \\
27B & S2 & T3--T1 & 0.00 & [0.00, 0.00] \\
27B & S2 & T4--T1 & 0.00 & [0.00, 0.00] \\
\end{longtable}
\endgroup

Table~\ref{tab:sc_w1_outputs} separates a raw terminal-value match from the
exclusive output categories. A terminal match can be correct or can coincide
with another state's value; it is not itself a wrong-state-selection label.
The exclusive columns instead allocate each response to correct, incorrect
but another legal-state value, another number, null, or invalid. Under T1/S1,
95/96 9B responses and 96/96 27B responses match the terminal value, while the
explicit target variants remove this pattern for S1.

% Source: supplied evidence report, lines 431--468.
\FloatBarrier
\begingroup
\footnotesize
\setlength{\tabcolsep}{3pt}
\renewcommand{\arraystretch}{1.08}
\begin{longtable}{llrrrrrr}
\caption{Output distributions for all target/wording conditions. All fractions have denominator 96. Terminal matching is non-exclusive; the remaining response categories are mutually exclusive. Each row has 16 program families.}\label{tab:sc_w1_outputs}\\
\toprule
Qwen3.5 & Condition & Terminal match & Correct & Wrong state & Other & Null & Invalid \\
\midrule
\endfirsthead
\multicolumn{8}{l}{\footnotesize Table~\thetable\ continued.}\\
\toprule
Qwen3.5 & Condition & Terminal match & Correct & Wrong state & Other & Null & Invalid \\
\midrule
\endhead
\midrule\multicolumn{8}{r}{\footnotesize Continued on the next page.}\\
\endfoot
\bottomrule
\endlastfoot
4B & T1\_S0 & 43/96 & 15/96 & 75/96 & 6/96 & 0/96 & 0/96 \\
4B & T1\_S1 & 48/96 & 37/96 & 48/96 & 11/96 & 0/96 & 0/96 \\
4B & T1\_S2 & 68/96 & 68/96 & 24/96 & 4/96 & 0/96 & 0/96 \\
4B & T2\_S0 & 33/96 & 96/96 & 0/96 & 0/96 & 0/96 & 0/96 \\
4B & T2\_S1 & 0/96 & 96/96 & 0/96 & 0/96 & 0/96 & 0/96 \\
4B & T2\_S2 & 69/96 & 69/96 & 11/96 & 16/96 & 0/96 & 0/96 \\
4B & T3\_S0 & 33/96 & 96/96 & 0/96 & 0/96 & 0/96 & 0/96 \\
4B & T3\_S1 & 0/96 & 96/96 & 0/96 & 0/96 & 0/96 & 0/96 \\
4B & T3\_S2 & 66/96 & 66/96 & 13/96 & 17/96 & 0/96 & 0/96 \\
4B & T4\_S0 & 33/96 & 96/96 & 0/96 & 0/96 & 0/96 & 0/96 \\
4B & T4\_S1 & 0/96 & 96/96 & 0/96 & 0/96 & 0/96 & 0/96 \\
4B & T4\_S2 & 66/96 & 66/96 & 13/96 & 17/96 & 0/96 & 0/96 \\
9B & T1\_S0 & 78/96 & 47/96 & 46/96 & 3/96 & 0/96 & 0/96 \\
9B & T1\_S1 & 95/96 & 0/96 & 95/96 & 1/96 & 0/96 & 0/96 \\
9B & T1\_S2 & 92/96 & 92/96 & 0/96 & 4/96 & 0/96 & 0/96 \\
9B & T2\_S0 & 33/96 & 96/96 & 0/96 & 0/96 & 0/96 & 0/96 \\
9B & T2\_S1 & 0/96 & 96/96 & 0/96 & 0/96 & 0/96 & 0/96 \\
9B & T2\_S2 & 88/96 & 88/96 & 0/96 & 8/96 & 0/96 & 0/96 \\
9B & T3\_S0 & 33/96 & 96/96 & 0/96 & 0/96 & 0/96 & 0/96 \\
9B & T3\_S1 & 0/96 & 96/96 & 0/96 & 0/96 & 0/96 & 0/96 \\
9B & T3\_S2 & 75/96 & 75/96 & 0/96 & 21/96 & 0/96 & 0/96 \\
9B & T4\_S0 & 33/96 & 96/96 & 0/96 & 0/96 & 0/96 & 0/96 \\
9B & T4\_S1 & 0/96 & 96/96 & 0/96 & 0/96 & 0/96 & 0/96 \\
9B & T4\_S2 & 74/96 & 74/96 & 0/96 & 22/96 & 0/96 & 0/96 \\
27B & T1\_S0 & 96/96 & 33/96 & 63/96 & 0/96 & 0/96 & 0/96 \\
27B & T1\_S1 & 96/96 & 0/96 & 96/96 & 0/96 & 0/96 & 0/96 \\
27B & T1\_S2 & 96/96 & 96/96 & 0/96 & 0/96 & 0/96 & 0/96 \\
27B & T2\_S0 & 33/96 & 96/96 & 0/96 & 0/96 & 0/96 & 0/96 \\
27B & T2\_S1 & 0/96 & 96/96 & 0/96 & 0/96 & 0/96 & 0/96 \\
27B & T2\_S2 & 96/96 & 96/96 & 0/96 & 0/96 & 0/96 & 0/96 \\
27B & T3\_S0 & 33/96 & 96/96 & 0/96 & 0/96 & 0/96 & 0/96 \\
27B & T3\_S1 & 0/96 & 96/96 & 0/96 & 0/96 & 0/96 & 0/96 \\
27B & T3\_S2 & 96/96 & 96/96 & 0/96 & 0/96 & 0/96 & 0/96 \\
27B & T4\_S0 & 33/96 & 96/96 & 0/96 & 0/96 & 0/96 & 0/96 \\
27B & T4\_S1 & 0/96 & 96/96 & 0/96 & 0/96 & 0/96 & 0/96 \\
27B & T4\_S2 & 96/96 & 96/96 & 0/96 & 0/96 & 0/96 & 0/96 \\
\end{longtable}
\endgroup

\FloatBarrier
\section{Non-count coordinate transformations}
\label{app:sc_noncount}

This experiment reuses the diagnostic logic, not the count action program.
The 47 worlds comprise 34 horizontal, six vertical, and seven depth relations.
The queried state is a binary relation, action 1 reflects the named coordinate
axis, and action 2 is identity. Accordingly, $s_1=s_2=\operatorname{flip}(s_0)$.
It tests neither two nontrivial transformations nor physical object motion.
Explicit and sham prompts contain the same relation code in the same context
region, but they are not claimed to have exactly equal token lengths.

Table~\ref{tab:sc_noncount_full} gives all four conditions and their archived
world intervals. The main matched comparison is in
Table~\ref{tab:sc_main_noncount}. Direct reporting remains substantially
better than the full transformation for 9B and 27B, while explicit state
changes the behavior of the two larger models much more than the sham.
The 4B run is an important negative boundary, not an omitted replication.

% Source: supplied evidence report, lines 567--571.
\FloatBarrier
\begingroup
\footnotesize
\setlength{\tabcolsep}{3pt}
\renewcommand{\arraystretch}{1.08}
\begin{longtable}{llrrl}
\caption{Complete non-count condition scores. Each row contains 47 worlds. Accuracy and confidence intervals are percentages.}\label{tab:sc_noncount_full}\\
\toprule
Qwen3.5 & Condition & Correct/worlds & Accuracy & 95\% CI \\
\midrule
\endfirsthead
\multicolumn{5}{l}{\footnotesize Table~\thetable\ continued.}\\
\toprule
Qwen3.5 & Condition & Correct/worlds & Accuracy & 95\% CI \\
\midrule
\endhead
\midrule\multicolumn{5}{r}{\footnotesize Continued on the next page.}\\
\endfoot
\bottomrule
\endlastfoot
4B & Direct $s_0$ & 35/47 & 74.47 & [61.70, 87.23] \\
4B & Full transformation & 12/47 & 25.53 & [12.77, 38.30] \\
4B & Explicit $s_0$ & 9/47 & 19.15 & [8.51, 31.91] \\
4B & Matched sham & 10/47 & 21.28 & [10.64, 34.04] \\
9B & Direct $s_0$ & 42/47 & 89.36 & [78.72, 97.87] \\
9B & Full transformation & 11/47 & 23.40 & [12.77, 36.17] \\
9B & Explicit $s_0$ & 37/47 & 78.72 & [65.96, 89.36] \\
9B & Matched sham & 9/47 & 19.15 & [8.51, 31.91] \\
27B & Direct $s_0$ & 41/47 & 87.23 & [76.60, 95.74] \\
27B & Full transformation & 4/47 & 8.51 & [2.13, 17.02] \\
27B & Explicit $s_0$ & 47/47 & 100.00 & [100.00, 100.00] \\
27B & Matched sham & 14/47 & 29.79 & [17.02, 42.55] \\
\end{longtable}
\endgroup

Table~\ref{tab:sc_noncount_family} retains the post-hoc family breakdown.
For 9B, the overall explicit-state advantage is driven by horizontal and
vertical relations; the depth subset instead gives 0/7 explicit answers
correct versus 2/7 sham answers. The small vertical and depth families have
wide or degenerate empirical intervals. These results qualify the scope of
``beyond counting'' and are not evidence for a uniform three-dimensional
state mechanism.

% Source: supplied evidence report, lines 588--598.
\FloatBarrier
\begingroup
\footnotesize
\setlength{\tabcolsep}{3pt}
\renewcommand{\arraystretch}{1.08}
\begin{longtable}{lllrrl}
\caption{Non-count family-by-condition results. The family analysis is post-hoc descriptive. Estimates and intervals are percentages; each row uses the family-specific world count in its denominator.}\label{tab:sc_noncount_family}\\
\toprule
Qwen3.5 & Family & Condition & Correct/worlds & Accuracy & 95\% CI \\
\midrule
\endfirsthead
\multicolumn{6}{l}{\footnotesize Table~\thetable\ continued.}\\
\toprule
Qwen3.5 & Family & Condition & Correct/worlds & Accuracy & 95\% CI \\
\midrule
\endhead
\midrule\multicolumn{6}{r}{\footnotesize Continued on the next page.}\\
\endfoot
\bottomrule
\endlastfoot
4B & Horizontal & Direct & 30/34 & 88.24 & [76.47, 97.06] \\
4B & Horizontal & Full & 5/34 & 14.71 & [2.94, 26.47] \\
4B & Horizontal & Explicit & 8/34 & 23.53 & [8.82, 38.24] \\
4B & Horizontal & Sham & 4/34 & 11.76 & [2.94, 23.53] \\
4B & Vertical & Direct & 5/6 & 83.33 & [50.00, 100.00] \\
4B & Vertical & Full & 1/6 & 16.67 & [0.00, 50.00] \\
4B & Vertical & Explicit & 0/6 & 0.00 & [0.00, 0.00] \\
4B & Vertical & Sham & 0/6 & 0.00 & [0.00, 0.00] \\
4B & Depth & Direct & 0/7 & 0.00 & [0.00, 0.00] \\
4B & Depth & Full & 6/7 & 85.71 & [57.14, 100.00] \\
4B & Depth & Explicit & 1/7 & 14.29 & [0.00, 42.86] \\
4B & Depth & Sham & 6/7 & 85.71 & [57.14, 100.00] \\
9B & Horizontal & Direct & 30/34 & 88.24 & [76.47, 97.06] \\
9B & Horizontal & Full & 6/34 & 17.65 & [5.88, 29.41] \\
9B & Horizontal & Explicit & 32/34 & 94.12 & [85.29, 100.00] \\
9B & Horizontal & Sham & 7/34 & 20.59 & [8.82, 35.29] \\
9B & Vertical & Direct & 6/6 & 100.00 & [100.00, 100.00] \\
9B & Vertical & Full & 0/6 & 0.00 & [0.00, 0.00] \\
9B & Vertical & Explicit & 5/6 & 83.33 & [50.00, 100.00] \\
9B & Vertical & Sham & 0/6 & 0.00 & [0.00, 0.00] \\
9B & Depth & Direct & 6/7 & 85.71 & [57.14, 100.00] \\
9B & Depth & Full & 5/7 & 71.43 & [42.86, 100.00] \\
9B & Depth & Explicit & 0/7 & 0.00 & [0.00, 0.00] \\
9B & Depth & Sham & 2/7 & 28.57 & [0.00, 57.14] \\
27B & Horizontal & Direct & 30/34 & 88.24 & [76.47, 97.06] \\
27B & Horizontal & Full & 2/34 & 5.88 & [0.00, 14.71] \\
27B & Horizontal & Explicit & 34/34 & 100.00 & [100.00, 100.00] \\
27B & Horizontal & Sham & 10/34 & 29.41 & [14.71, 44.12] \\
27B & Vertical & Direct & 5/6 & 83.33 & [50.00, 100.00] \\
27B & Vertical & Full & 2/6 & 33.33 & [0.00, 66.67] \\
27B & Vertical & Explicit & 6/6 & 100.00 & [100.00, 100.00] \\
27B & Vertical & Sham & 1/6 & 16.67 & [0.00, 50.00] \\
27B & Depth & Direct & 6/7 & 85.71 & [57.14, 100.00] \\
27B & Depth & Full & 0/7 & 0.00 & [0.00, 0.00] \\
27B & Depth & Explicit & 7/7 & 100.00 & [100.00, 100.00] \\
27B & Depth & Sham & 3/7 & 42.86 & [14.29, 85.71] \\
\end{longtable}
\endgroup

The family-specific explicit-minus-sham contrasts in
Table~\ref{tab:sc_noncount_family_effects} preserve zero and negative effects.
They were computed from existing predictions without selecting new worlds or
changing prompts. They should not be interpreted as multiple independently
confirmed positive replications.

% Source: supplied evidence report, lines 588--598.
\begin{table}[htbp]
\centering
\caption{Post-hoc family-specific paired effects. Net numerator/worlds is the sum of paired correctness differences, not the count of rescued worlds. Units are percentage points.}
\label{tab:sc_noncount_family_effects}
\small
\setlength{\tabcolsep}{4pt}
\begin{adjustbox}{max width=\linewidth}
\begin{tabular}{llrrrl}
\toprule
Qwen3.5 & Family & Worlds & Net/worlds & $\Delta$ & 95\% CI \\
\midrule
4B & Horizontal & 34 & 4/34 & 11.76 & [-8.82, 29.41] \\
4B & Vertical & 6 & 0/6 & 0.00 & [0.00, 0.00] \\
4B & Depth & 7 & -5/7 & -71.43 & [-100.00, -42.86] \\
9B & Horizontal & 34 & 25/34 & 73.53 & [55.88, 88.24] \\
9B & Vertical & 6 & 5/6 & 83.33 & [50.00, 100.00] \\
9B & Depth & 7 & -2/7 & -28.57 & [-57.14, 0.00] \\
27B & Horizontal & 34 & 24/34 & 70.59 & [55.88, 85.29] \\
27B & Vertical & 6 & 5/6 & 83.33 & [50.00, 100.00] \\
27B & Depth & 7 & 4/7 & 57.14 & [14.29, 85.71] \\
\bottomrule
\end{tabular}
\end{adjustbox}
\end{table}

Table~\ref{tab:sc_noncount_conditional} applies the same-world diagnostic
pattern after the experiment. The denominator for rescue is all worlds with
a correct direct report and an incorrect full transformation. The resulting
31/34 for 9B and 40/40 for 27B are consistent with the aggregate contrast,
but remain explicitly post-hoc. Conditioning on the already rescued
Type-A-like count would instead make the rescue rate tautological.

% Source: supplied evidence report, lines 607--611.
\begin{table}[htbp]
\centering
\caption{Post-hoc Type-A-like cases in the non-count experiment. Each world contributes one sequence. The conditional denominator includes non-rescued direct-correct/full-wrong cases.}
\label{tab:sc_noncount_conditional}
\small
\setlength{\tabcolsep}{4pt}
\begin{adjustbox}{max width=\linewidth}
\begin{tabular}{lrrrr}
\toprule
Qwen3.5 & Type-A-like/worlds & Rescued/eligible gap & Rate (\%) & 95\% CI \\
\midrule
4B & 8/47 & 8/33 & 24.24 & [9.09, 39.39] \\
9B & 31/47 & 31/34 & 91.18 & [79.41, 100.00] \\
27B & 40/47 & 40/40 & 100.00 & [100.00, 100.00] \\
\bottomrule
\end{tabular}
\end{adjustbox}
\end{table}

\FloatBarrier
\section{Representation probing}
\label{app:sc_probes}

R1 fits linear ridge readers to archived residual activations and compares
them with a majority baseline and a combined word/number-bag-plus-length
baseline. B1 is partitioned by world into 48 training, 16 selection, and
16 evaluation worlds, with two sequences per world. The symbolic B2 partition
uses 10/3/3 program families. Selection scores belong to the original
regularization selection stage; evaluation scores are not used to choose a
new layer. Separate length-only and bag-only baselines were not run, so the
combined baseline does not isolate the contribution of either component.

Table~\ref{tab:sc_probe_core} reports the prespecified B01/S1/checkpoint
slice. Even the best observed fixed-layer scores for 4B and 9B only match the
68.75\% surface baseline in this slice; the 27B scores are lower. This is a
comparison of observed point estimates, not a new paired significance test or
proof that state information is absent. Explicit-checkpoint contexts and
other variables can have different readout behavior. The full ledger in
Appendix~\ref{app:sc_probe_ledger} retains every variable, model, layer,
anchor, and unavailable-variable record rather than selecting only this
negative slice.

% Source: supplied evidence report, lines 629--660.
\FloatBarrier
\begingroup
\footnotesize
\setlength{\tabcolsep}{3pt}
\renewcommand{\arraystretch}{1.08}
\begin{longtable}{lllrrl}
\caption{Prespecified B01/S1/checkpoint probing slice. Every evaluated row uses 32 sequences from 16 worlds. Accuracy and intervals are percentages. Surface denotes the combined word/number-bag-plus-length baseline.}\label{tab:sc_probe_core}\\
\toprule
Qwen3.5 & Reader & Block & Correct/N & Accuracy & 95\% CI \\
\midrule
\endfirsthead
\multicolumn{6}{l}{\footnotesize Table~\thetable\ continued.}\\
\toprule
Qwen3.5 & Reader & Block & Correct/N & Accuracy & 95\% CI \\
\midrule
\endhead
\midrule\multicolumn{6}{r}{\footnotesize Continued on the next page.}\\
\endfoot
\bottomrule
\endlastfoot
4B & Majority & \scNA & 10/32 & 31.25 & [12.50, 56.25] \\
4B & Surface & \scNA & 22/32 & 68.75 & [46.88, 87.50] \\
4B & Residual & 0 & 12/32 & 37.50 & [12.50, 62.50] \\
4B & Residual & 4 & 18/32 & 56.25 & [31.25, 81.25] \\
4B & Residual & 9 & 22/32 & 68.75 & [43.75, 87.50] \\
4B & Residual & 13 & 20/32 & 62.50 & [37.50, 87.50] \\
4B & Residual & 18 & 20/32 & 62.50 & [37.50, 87.50] \\
4B & Residual & 22 & 22/32 & 68.75 & [43.75, 87.50] \\
4B & Residual & 27 & 22/32 & 68.75 & [43.75, 87.50] \\
4B & Residual & 31 & 22/32 & 68.75 & [43.75, 87.50] \\
9B & Majority & \scNA & 10/32 & 31.25 & [12.50, 56.25] \\
9B & Surface & \scNA & 22/32 & 68.75 & [46.88, 87.50] \\
9B & Residual & 0 & 16/32 & 50.00 & [25.00, 75.00] \\
9B & Residual & 4 & 18/32 & 56.25 & [31.25, 81.25] \\
9B & Residual & 9 & 20/32 & 62.50 & [37.50, 81.25] \\
9B & Residual & 13 & 16/32 & 50.00 & [25.00, 75.00] \\
9B & Residual & 18 & 18/32 & 56.25 & [31.25, 81.25] \\
9B & Residual & 22 & 20/32 & 62.50 & [37.50, 87.50] \\
9B & Residual & 27 & 22/32 & 68.75 & [43.75, 87.50] \\
9B & Residual & 31 & 20/32 & 62.50 & [37.50, 81.25] \\
27B & Majority & \scNA & 10/32 & 31.25 & [12.50, 56.25] \\
27B & Surface & \scNA & 22/32 & 68.75 & [46.88, 87.50] \\
27B & Residual & 0 & 14/32 & 43.75 & [18.75, 68.75] \\
27B & Residual & 9 & 14/32 & 43.75 & [18.75, 68.75] \\
27B & Residual & 18 & 18/32 & 56.25 & [31.25, 81.25] \\
27B & Residual & 27 & 18/32 & 56.25 & [31.25, 81.25] \\
27B & Residual & 36 & 18/32 & 56.25 & [31.25, 81.25] \\
27B & Residual & 45 & 16/32 & 50.00 & [25.00, 75.00] \\
27B & Residual & 54 & 18/32 & 56.25 & [31.25, 81.25] \\
27B & Residual & 63 & 18/32 & 56.25 & [31.25, 81.25] \\
\end{longtable}
\endgroup

The target-state label $Z$ is constant in the relevant training slices and is
not identifiable by these probes. B2 provides neither the entity nor the
protected-fact labels used by some visual probes. Additional slices have
unseen evaluation classes. These are limitations of the measurement panel,
not evidence that the corresponding information is absent from the model.
A correct probe answer, when obtained, likewise does not establish that the
model uses that representation to produce its own answer.

\FloatBarrier
\section{Counterfactual intermediate-state interchange}
\label{app:sc_i2}

I2 tests a specific high-level mapping: replacing one block-output token with
a donor activation is hypothesized to substitute the donor's intermediate
state into the recipient's subsequent computation. For donor $d$ and recipient
$r$, the target is
\begin{equation}
 y_{\mathrm{CF}}=\tau_{2,r}(s_{1,d}),
 \label{eq:sc_i2_target}
\end{equation}
not the donor's final answer. The actual symbolic example in
Appendix~\ref{app:sc_prompts_i2} has recipient $s_1=2$ and an add-three second
action, donor $s_1=12$ and final answer 11, and a counterfactual target of 15.
This separates the intended composition from direct donor-final copying.

The experiment uses 16 frozen pairs from ten program families, eight fixed
blocks, and three anchors. The symbolic checkpoint anchor aliases the end of
action 1 because the symbolic input has no explicit checkpoint slot.
Table~\ref{tab:sc_i2_grid} preserves the full grid: no cell produces the
specified counterfactual output. Its 384 rows are repeated interventions on
16 pairs, not 384 independent problems.

% Source: supplied evidence report, lines 684--693.
\begin{table}[htbp]
\centering
\caption{I2 primary interchange grid on 9B. Every cell has 16 pairs from ten program families. All CF-hit counts are retained; the empirical all-zero bootstrap interval is not a population certainty.}
\label{tab:sc_i2_grid}
\small
\setlength{\tabcolsep}{4pt}
\begin{adjustbox}{max width=\linewidth}
\begin{tabular}{lrrr}
\toprule
Block & Checkpoint & After action 2 & Query \\
\midrule
0 & 0/16 & 0/16 & 0/16 \\
4 & 0/16 & 0/16 & 0/16 \\
9 & 0/16 & 0/16 & 0/16 \\
13 & 0/16 & 0/16 & 0/16 \\
18 & 0/16 & 0/16 & 0/16 \\
22 & 0/16 & 0/16 & 0/16 \\
27 & 0/16 & 0/16 & 0/16 \\
31 & 0/16 & 0/16 & 0/16 \\
\bottomrule
\end{tabular}
\end{adjustbox}
\end{table}

Table~\ref{tab:sc_i2_continuous} shows that zero counterfactual hits do not
mean the interventions are completely inert. Thirteen free answers change,
and the mean original-gold log probability decreases, but the counterfactual
log-probability interval spans zero. Thus the observed changes do not provide
the predicted state-interchange evidence. Absolute donor-answer matches in
this table include pre-existing matches and are not rates of newly induced
copying.

% Source: supplied evidence report, lines 703--712.
\begin{table}[htbp]
\centering
\caption{I2 primary continuous and generation endpoints. LP changes are nats; event means are proportions. Every aggregate uses 384 interventions clustered into ten program families. The numerator for an LP change is a sum of changes, not a number of correct trials.}
\label{tab:sc_i2_continuous}
\small
\setlength{\tabcolsep}{4pt}
\begin{adjustbox}{max width=\linewidth}
\begin{tabular}{lrrl}
\toprule
Metric & Sum or events/N & Mean & 95\% family CI \\
\midrule
$\Delta\ell(y_{\mathrm{CF}})$ & 1.462551332/384 & 0.003809 & [-0.0345, 0.0369] \\
$\Delta\ell(y_{\mathrm{original}})$ & -20.209095186/384 & -0.052628 & [-0.0873, -0.0162] \\
$\Delta\ell(y_{\mathrm{donor\ final}})$ & 6.596495041/384 & 0.017178 & [-0.0151, 0.0467] \\
$\Delta\ell(s_{1,d})$ & 7.365594004/384 & 0.019181 & [-0.0219, 0.0551] \\
Free answer changed & 13.000000000/384 & 0.033854 & [0.0089, 0.0613] \\
Counterfactual hit & 0.000000000/384 & 0.000000 & [0.0000, 0.0000] \\
Donor-final value match & 24.000000000/384 & 0.062500 & [0.0000, 0.1875] \\
Donor-intermediate value match & 1.000000000/384 & 0.002604 & [0.0000, 0.0078] \\
\bottomrule
\end{tabular}
\end{adjustbox}
\end{table}

Tables~\ref{tab:sc_i2_controls} and~\ref{tab:sc_i2_control_matches}
distinguish execution coverage, each control's own counterfactual target,
and retained baseline answers. The self control preserves the baseline
answer on 128/128 trials while matching the original gold on 112/128; those
are different quantities. A random donor's own counterfactual-hit count is
not directly comparable to the primary target-hit count when the targets
differ.

% Source: supplied evidence report, lines 724--732.
\begin{table}[htbp]
\centering
\caption{I2 control coverage and primary generation counts. Each control has its own predefined counterfactual target. The columns must not be interpreted as identical-target comparisons across rows.}
\label{tab:sc_i2_controls}
\small
\setlength{\tabcolsep}{4pt}
\begin{adjustbox}{max width=\linewidth}
\begin{tabular}{lrrrr}
\toprule
Control & Returned/planned & Pairs/families & Own CF hit & Baseline preserved \\
\midrule
Random donor & 112/120 & 15/8 & 16/112 & 111/112 \\
Primary & 384/384 & 16/10 & 0/384 & 371/384 \\
Same donor, different $\tau_2$ & 120/120 & 15/10 & 0/120 & 117/120 \\
Same donor-final, different $s_1$ & 80/80 & 10/7 & 8/80 & 79/80 \\
Same $s_1$, different history & 48/56 & 7/5 & 0/48 & 48/48 \\
Self & 128/128 & 16/10 & 112/128 & 128/128 \\
Unrelated position & 128/128 & 16/10 & 0/128 & 128/128 \\
\bottomrule
\end{tabular}
\end{adjustbox}
\end{table}

% Source: supplied evidence report, lines 724--732.
\begin{table}[htbp]
\centering
\caption{Additional I2 control endpoints. Original-gold hits measure correctness, whereas donor-final matches are absolute numerical matches. No returned control response is null or invalid in this table.}
\label{tab:sc_i2_control_matches}
\small
\setlength{\tabcolsep}{4pt}
\begin{adjustbox}{max width=\linewidth}
\begin{tabular}{lrrrr}
\toprule
Control & Original gold & Donor-final match & Null & Invalid \\
\midrule
Random donor & 95/112 & 24/112 & 0/112 & 0/112 \\
Primary & 336/384 & 24/384 & 0/384 & 0/384 \\
Same donor, different $\tau_2$ & 117/120 & 32/120 & 0/120 & 0/120 \\
Same donor-final, different $s_1$ & 79/80 & 8/80 & 0/80 & 0/80 \\
Same $s_1$, different history & 48/48 & 0/48 & 0/48 & 0/48 \\
Self & 112/128 & 112/128 & 0/128 & 0/128 \\
Unrelated position & 112/128 & 8/128 & 0/128 & 0/128 \\
\bottomrule
\end{tabular}
\end{adjustbox}
\end{table}

Table~\ref{tab:sc_i2_same_target} restricts probability comparisons to the
same primary counterfactual candidate where that score was saved. None of the
available clustered intervals establishes a stable positive matched
advantage. The single-family comparison lacks a stored interval and cannot
substitute for broader evidence. The experiment does not include an
independent same-value unrelated-register control; I1's initial-state sham
must not be reused as if it were that missing I2 control.

% Source: supplied evidence report, lines 742--748.
\begin{table}[htbp]
\centering
\caption{Checkpoint primary-minus-control contrasts scored on the same primary-CF answer. Only saved, comparable candidate scores are included; differences are nats.}
\label{tab:sc_i2_same_target}
\small
\setlength{\tabcolsep}{4pt}
\begin{adjustbox}{max width=\linewidth}
\begin{tabular}{lrrl}
\toprule
Control & Paired N/families & Mean difference & 95\% CI \\
\midrule
Random donor & 64/6 & 0.012371 & [-0.1109, 0.0996] \\
Same donor-final, different $s_1$ & 8/1 & 0.205366 & \textemdash \\
Same $s_1$, different history & 48/5 & 0.015832 & [-0.0259, 0.0431] \\
Self & 24/3 & -0.076182 & [-0.2634, 0.0215] \\
Unrelated position & 128/10 & 0.023166 & [-0.0351, 0.0675] \\
\bottomrule
\end{tabular}
\end{adjustbox}
\end{table}

\textbf{Technical exclusions.}
A single fixed technical follow-up still found one context with cached versus
uncached outputs 11 versus 12; the earlier minimum top-two margin was zero.
Sixteen planned control interventions remain unrun under
\texttt{NOT\_RUN\_ENGINE\_EQUIVALENCE\_UNRESOLVED}. They do not reduce the
fully returned 384-primary denominator, but they limit control closure.
There are also 240 historical records without a comparable saved
primary-CF probability, including unrun or unrecorded cases. We do not
impute those values. The negative conclusion is limited to the tested
single-token mapping, not every possible state representation or causal
intervention.

\FloatBarrier
\section{Initial-state donor interventions: localization and validation}
\label{app:sc_i1}

I1 studies a different contrast from I2. The recipient is the original visual
full task, the informative donor is the same world and program with correct
$s_0$, and the matched sham donor places that same $s_0$ value before the
actions as an unrelated register. A success donor is a historical Type C case
from another world with the same new split, action type, and sequence;
a different final value is preferred before deterministic hash tie-breaking.
When no qualifying success donor exists, the control is not fabricated.
The exact donor texts appear in Appendix~\ref{app:sc_prompts_i1}.

\subsection{Scope and eligible comparison sets}
\label{app:sc_i1_design}

The 80 visual worlds are hash-ordered into 48 LOCALIZE and 32 SELECT worlds.
This is a window-selection split, not a claim that these worlds were never
used in earlier behavioral or representation experiments. LOCALIZE contains
64 eligible A/B sequences from 41 worlds. Eight actual blocks
$\{0,5,9,14,19,22,26,31\}$, four anchors, and three donor types produce
6,144 planned interventions. Of these, 5,088 return; 704 lack an eligible
success donor and 352 do not pass engine equivalence. The primary Type A
informative/sham matched set is 25 sequences from 20 worlds.

Table~\ref{tab:sc_anchors} specifies the anchors in the actual expanded token
sequence. Donor and recipient positions are matched by semantic boundary,
not by an assumed common absolute index. The checkpoint alias in the
symbolic I2 prompts is explicitly different from a written oracle checkpoint.

\begin{table}[htbp]
\centering
\caption{Actual token-boundary definitions. Positions are located after processor expansion and verified against the expanded token IDs; character offsets alone are not used as token indices.}
\label{tab:sc_anchors}
\footnotesize
\setlength{\tabcolsep}{4pt}
\begin{adjustbox}{max width=\linewidth}
\begin{tabular}{L{0.30\linewidth}L{0.62\linewidth}}
\toprule
Anchor & Definition \\
\midrule
\texttt{P\_CONTEXT\_END} & Last text token before the ACTION\_1 line (alias P\_PRE). \\
\texttt{P\_A1\_END} & Last token in the ACTION\_1 line. \\
\texttt{P\_A2\_END} & Last token in the ACTION\_2 line. \\
\texttt{P\_QUERY} & Last token before the OUTPUT CONTRACT line. \\
\texttt{P\_CHECKPOINT} & End of STATE\_NOTE; aliases P\_A1 when the symbolic prompt has no checkpoint slot. \\
\bottomrule
\end{tabular}
\end{adjustbox}
\end{table}

Each intervention replaces one actual block-output token and then continues
computation on the recipient. The donor vector is reapplied during full-prefix
recomputation for every generated token. Technical acceptance checks include
cached/uncached replay, no-op/self token behavior, and candidate-score
agreement. Full-block output replacement is not an intervention on a uniquely
identified state feature.

\subsection{All localization windows}
\label{app:sc_i1_localize}

Table~\ref{tab:sc_i1_windows} gives every informative, sham, and matched
specificity estimate. Its intervals are descriptive because the same
LOCALIZE data are used to select windows. The actual rule ranks positive
world-mean Type A informative-minus-sham effects, keeps at most two, and
breaks ties deterministically by depth and anchor. It does not impose the
later SELECT confidence-bound or world-count gate during localization.
The selected candidates are blocks 9 and 5 at context end, frozen before
SELECT outcomes are read.

% Source: supplied evidence report, lines 770--803.
\FloatBarrier
\begingroup
\footnotesize
\setlength{\tabcolsep}{3pt}
\renewcommand{\arraystretch}{1.08}
\begin{longtable}{rlrrrl}
\caption{All 32 I1 localization windows. LP differences are nats, world-averaged over the same 25 Type A sequences/20 worlds. No multiplicity or selection adjustment is applied to these exploratory intervals.}\label{tab:sc_i1_windows}\\
\toprule
Block & Anchor & Info $\Delta$LP & Sham $\Delta$LP & Specific $\Delta$LP & 95\% CI \\
\midrule
\endfirsthead
\multicolumn{6}{l}{\footnotesize Table~\thetable\ continued.}\\
\toprule
Block & Anchor & Info $\Delta$LP & Sham $\Delta$LP & Specific $\Delta$LP & 95\% CI \\
\midrule
\endhead
\midrule\multicolumn{6}{r}{\footnotesize Continued on the next page.}\\
\endfoot
\bottomrule
\endlastfoot
0 & A1 end & -0.001610 & -0.004306 & 0.002696 & [-0.0290, 0.0312] \\
0 & A2 end & -0.001107 & 0.015729 & -0.016835 & [-0.0394, 0.0051] \\
0 & Context end & -0.007092 & -0.012296 & 0.005204 & [-0.0211, 0.0328] \\
0 & Query & -0.012741 & 0.010724 & -0.023465 & [-0.0569, 0.0078] \\
5 & A1 end & 0.000041 & -0.013612 & 0.013653 & [-0.0121, 0.0376] \\
5 & A2 end & -0.009554 & -0.007460 & -0.002094 & [-0.0347, 0.0301] \\
5 & Context end & 0.002909 & -0.015594 & 0.018502 & [-0.0115, 0.0512] \\
5 & Query & -0.007725 & -0.010906 & 0.003182 & [-0.0195, 0.0269] \\
9 & A1 end & 0.009425 & 0.004335 & 0.005090 & [-0.0235, 0.0349] \\
9 & A2 end & -0.047823 & -0.026203 & -0.021620 & [-0.0613, 0.0102] \\
9 & Context end & 0.033206 & -0.007099 & 0.040305 & [0.0109, 0.0700] \\
9 & Query & -0.006226 & -0.001790 & -0.004436 & [-0.0241, 0.0161] \\
14 & A1 end & 0.002873 & 0.029332 & -0.026460 & [-0.0519, -0.0003] \\
14 & A2 end & -0.152398 & -0.075489 & -0.076910 & [-0.1623, -0.0048] \\
14 & Context end & -0.011004 & 0.016452 & -0.027456 & [-0.0498, -0.0070] \\
14 & Query & 0.013651 & 0.008213 & 0.005438 & [-0.0175, 0.0281] \\
19 & A1 end & -0.007951 & -0.009769 & 0.001818 & [-0.0126, 0.0160] \\
19 & A2 end & -0.018943 & -0.005286 & -0.013657 & [-0.0308, 0.0025] \\
19 & Context end & -0.000241 & -0.002615 & 0.002374 & [-0.0191, 0.0227] \\
19 & Query & -0.004797 & 0.002579 & -0.007376 & [-0.0284, 0.0147] \\
22 & A1 end & -0.008295 & 0.005784 & -0.014079 & [-0.0278, -0.0024] \\
22 & A2 end & 0.006183 & -0.002741 & 0.008924 & [-0.0104, 0.0273] \\
22 & Context end & -0.008228 & -0.001462 & -0.006766 & [-0.0211, 0.0089] \\
22 & Query & 0.005037 & 0.001665 & 0.003372 & [-0.0129, 0.0186] \\
26 & A1 end & 0.000604 & -0.007602 & 0.008206 & [-0.0096, 0.0278] \\
26 & A2 end & 0.005336 & -0.000164 & 0.005500 & [-0.0148, 0.0246] \\
26 & Context end & -0.001792 & 0.009439 & -0.011231 & [-0.0261, 0.0023] \\
26 & Query & -0.000405 & -0.011473 & 0.011068 & [-0.0014, 0.0272] \\
31 & A1 end & 0.000000 & 0.000000 & 0.000000 & [0.0000, 0.0000] \\
31 & A2 end & 0.000000 & 0.000000 & 0.000000 & [0.0000, 0.0000] \\
31 & Context end & 0.000000 & 0.000000 & 0.000000 & [0.0000, 0.0000] \\
31 & Query & 0.000000 & 0.000000 & 0.000000 & [0.0000, 0.0000] \\
\end{longtable}
\endgroup

Table~\ref{tab:sc_i1_localize_events} complements probability changes with
free-generation outcomes for every window. Informative-donor rescue and
null/invalid counts use the returned Type A cases. The success-donor value
matches instead use their separately eligible A/B controls. There are no
baseline-correct primary Type A recipients, so their harm denominator is
inapplicable rather than zero. These differences in eligibility prevent
combining all columns into one denominator or declaring a window selective
from a positive mean alone.

% Source: supplied evidence report, lines 770--803.
\FloatBarrier
\begingroup
\footnotesize
\setlength{\tabcolsep}{3pt}
\renewcommand{\arraystretch}{1.08}
\begin{longtable}{rlrrrr}
\caption{Generation endpoints for all localization windows. Informative rescue/null/invalid use 25 sequences/20 worlds; success-donor absolute matching uses 34 sequences/27 worlds. Absolute matches need not be induced by patching.}\label{tab:sc_i1_localize_events}\\
\toprule
Block & Anchor & Info rescue & Success-donor match & Info null & Info invalid \\
\midrule
\endfirsthead
\multicolumn{6}{l}{\footnotesize Table~\thetable\ continued.}\\
\toprule
Block & Anchor & Info rescue & Success-donor match & Info null & Info invalid \\
\midrule
\endhead
\midrule\multicolumn{6}{r}{\footnotesize Continued on the next page.}\\
\endfoot
\bottomrule
\endlastfoot
0 & A1 end & 1/25 & 10/34 & 1/25 & 0/25 \\
0 & A2 end & 1/25 & 10/34 & 0/25 & 0/25 \\
0 & Context end & 1/25 & 11/34 & 1/25 & 0/25 \\
0 & Query & 0/25 & 10/34 & 0/25 & 0/25 \\
5 & A1 end & 1/25 & 10/34 & 1/25 & 0/25 \\
5 & A2 end & 1/25 & 10/34 & 1/25 & 0/25 \\
5 & Context end & 1/25 & 10/34 & 1/25 & 0/25 \\
5 & Query & 1/25 & 10/34 & 0/25 & 0/25 \\
9 & A1 end & 2/25 & 11/34 & 1/25 & 0/25 \\
9 & A2 end & 0/25 & 11/34 & 1/25 & 0/25 \\
9 & Context end & 1/25 & 10/34 & 1/25 & 0/25 \\
9 & Query & 1/25 & 10/34 & 0/25 & 0/25 \\
14 & A1 end & 1/25 & 12/34 & 1/25 & 0/25 \\
14 & A2 end & 0/25 & 9/34 & 0/25 & 0/25 \\
14 & Context end & 1/25 & 10/34 & 2/25 & 0/25 \\
14 & Query & 0/25 & 10/34 & 0/25 & 0/25 \\
19 & A1 end & 1/25 & 10/34 & 0/25 & 0/25 \\
19 & A2 end & 0/25 & 11/34 & 0/25 & 0/25 \\
19 & Context end & 1/25 & 10/34 & 0/25 & 0/25 \\
19 & Query & 1/25 & 10/34 & 0/25 & 0/25 \\
22 & A1 end & 0/25 & 10/34 & 0/25 & 0/25 \\
22 & A2 end & 1/25 & 10/34 & 0/25 & 0/25 \\
22 & Context end & 0/25 & 10/34 & 1/25 & 0/25 \\
22 & Query & 0/25 & 10/34 & 1/25 & 0/25 \\
26 & A1 end & 0/25 & 10/34 & 1/25 & 0/25 \\
26 & A2 end & 0/25 & 10/34 & 0/25 & 0/25 \\
26 & Context end & 0/25 & 10/34 & 0/25 & 0/25 \\
26 & Query & 0/25 & 10/34 & 0/25 & 0/25 \\
31 & A1 end & 0/25 & 10/34 & 0/25 & 0/25 \\
31 & A2 end & 0/25 & 10/34 & 0/25 & 0/25 \\
31 & Context end & 0/25 & 10/34 & 0/25 & 0/25 \\
31 & Query & 0/25 & 10/34 & 0/25 & 0/25 \\
\end{longtable}
\endgroup

The localize-only positive interval at block 9 is therefore a candidate,
not a confirmed mechanism. The two fixed candidates are assessed below
without changing their positions in response to SELECT.

\subsection{SELECT specificity and generation}
\label{app:sc_i1_select}

Table~\ref{tab:sc_select_detail} reports the candidate-validation results.
Both ordinary intervals span zero, as do the corresponding Bonferroni-adjusted
lower bounds. Each window has zero rescues among the 17 interface-valid
primary Type A sequences from 14 worlds. There are 20 returned A-labeled
sequences from 15 worlds before the three interface cases are separated;
these two denominators are not interchangeable.

% Source: supplied evidence report, lines 850--853.
\begin{table}[htbp]
\centering
\caption{Frozen I1 candidate-validation results. All effects are informative-minus-sham correct-answer LP changes in nats. Candidates were selected using LOCALIZE, not SELECT.}
\label{tab:sc_select_detail}
\small
\setlength{\tabcolsep}{4pt}
\begin{adjustbox}{max width=\linewidth}
\begin{tabular}{lrrrrr}
\toprule
Block/anchor & Seq./worlds & Specific $\Delta$LP & 95\% CI & Adjusted lower & Rescue \\
\midrule
9/context & 17/14 & 0.035659 & [-0.0034, 0.0792] & -0.008274 & 0/17 \\
5/context & 17/14 & 0.019488 & [-0.0118, 0.0506] & -0.015685 & 0/17 \\
\bottomrule
\end{tabular}
\end{adjustbox}
\end{table}

Table~\ref{tab:sc_select_protection} records unchanged-state, unrelated-fact,
success-case, and reverse-patch checks. No damage or reverse answer change is
observed, but forward rescue is also absent. Consequently, these zeros do
not demonstrate a safe, selective repair. The reverse endpoint is any answer
change, not automatically a correct-to-wrong transition, and has no separately
precomputed interval.

% Source: supplied evidence report, lines 863--872.
\begin{table}[htbp]
\centering
\caption{SELECT protection and reverse checks. Denominators are the separately eligible sequence sets, with distinct-world counts stated explicitly. Zero-width empirical damage intervals do not establish zero population risk.}
\label{tab:sc_select_protection}
\small
\setlength{\tabcolsep}{4pt}
\begin{adjustbox}{max width=\linewidth}
\begin{tabular}{llrrl}
\toprule
Block & Check & Events/N & Worlds & Archived 95\% CI \\
\midrule
9 & Initial fact & 0/42 & 22 & [0.00, 0.00] \\
9 & Protected fact & 0/14 & 10 & [0.00, 0.00] \\
9 & Originally successful & 0/22 & 16 & [0.00, 0.00] \\
9 & Reverse answer change & 0/40 & 26 & \textemdash \\
5 & Initial fact & 0/42 & 22 & [0.00, 0.00] \\
5 & Protected fact & 0/14 & 10 & [0.00, 0.00] \\
5 & Originally successful & 0/22 & 16 & [0.00, 0.00] \\
5 & Reverse answer change & 0/40 & 26 & \textemdash \\
\bottomrule
\end{tabular}
\end{adjustbox}
\end{table}

Table~\ref{tab:sc_select_copy} separates absolute donor-value matching from
new matching caused by the comparison. The frozen gate uses the absolute
post-match proportion and fails for both candidates. However, nine matches
were already present at baseline; only one new match is observed for block 9
and none for block 5. It would therefore be incorrect to claim that the
intervention induced donor copying in roughly a third of cases. Even a newly
matching numerical value is not by itself a mechanistic proof of copying.

% Source: supplied evidence report, lines 877--880.
\begin{table}[htbp]
\centering
\caption{Donor-value matches in the eligible success-donor comparison. Both rows contain 26 sequences from 20 worlds. The gate uses absolute post matches; the interpretation also reports baseline and newly induced matches.}
\label{tab:sc_select_copy}
\small
\setlength{\tabcolsep}{4pt}
\begin{adjustbox}{max width=\linewidth}
\begin{tabular}{lrrrr}
\toprule
Block & Post matches & Baseline matches & New matches & Worlds \\
\midrule
9 & 10/26 & 9/26 & 1/26 & 20 \\
5 & 9/26 & 9/26 & 0/26 & 20 \\
\bottomrule
\end{tabular}
\end{adjustbox}
\end{table}

Table~\ref{tab:sc_select_coverage} retains the selected-window execution
coverage, including the supplementary protection/reverse queries on both
partitions. These totals differ from the initial 6,144-item localization grid.
A completed submission is not the same as complete matched-control coverage;
ineligible donors and unresolved engine checks remain separate categories.
Returned null values are included as observed outputs, not converted into
missingness or numerical zero.

% Source: supplied evidence report, lines 885--889.
\begin{table}[htbp]
\centering
\caption{Selected-window and supplementary-query execution coverage. Invalid and null entries are counts among returned trials. The rows retain the source partition identifiers; the total is not an independent sample size.}
\label{tab:sc_select_coverage}
\small
\setlength{\tabcolsep}{4pt}
\begin{adjustbox}{max width=\linewidth}
\begin{tabular}{lrrrrrrrr}
\toprule
Partition & Plan & Return & No donor & Engine & Invalid & Null & Cases & Worlds \\
\midrule
ALL & 2596 & 2392 & 142 & 62 & 0 & 331 & 154 & 80 \\
LOCALIZE & 1400 & 1288 & 88 & 24 & 0 & 183 & 92 & 48 \\
SELECT & 1196 & 1104 & 54 & 38 & 0 & 148 & 62 & 32 \\
\bottomrule
\end{tabular}
\end{adjustbox}
\end{table}

\subsection{Frozen decision rule and unrun branches}
\label{app:sc_i1_gate}

Table~\ref{tab:sc_select_gate} reproduces the gate's actual outcomes. The
requirements are a positive informative mean, a positive two-window-adjusted
specificity lower bound, at least eight primary worlds, at least half of
worlds having positive effects, all matched controls returned, each damage
rate at most 10\%, and absolute identifiable donor matching at most 25\%.
The frozen implementation does not have a separate Boolean requirement
\texttt{rescue>0}; the zero-rescue result is reported as an important measured
endpoint, not retrospectively added to the frozen rule.

% Source: supplied evidence report, lines 902--923.
\begin{table}[htbp]
\centering
\caption{Actual frozen I1 continuation checks. The execution-completion flag is reproduced as recorded and does not override the separately failed matched-control completeness check.}
\label{tab:sc_select_gate}
\small
\setlength{\tabcolsep}{4pt}
\begin{adjustbox}{max width=\linewidth}
\begin{tabular}{lcc}
\toprule
Check & Block 9 & Block 5 \\
\midrule
All matched controls returned & False & False \\
Execution-completion flag & True & True \\
Absolute donor match at most 25\% & False & False \\
At least 8 primary worlds & True & True \\
Initial-fact damage at most 10\% & True & True \\
Informative mean greater than zero & True & True \\
Adjusted specificity lower bound above zero & False & False \\
At least 50\% positive-effect worlds & True & True \\
Protected-fact damage at most 10\% & True & True \\
Success-case damage at most 10\% & True & True \\
\bottomrule
\end{tabular}
\end{adjustbox}
\end{table}

Both candidates receive \texttt{NO\_GO}. The interpretation-limited copy
check is not the only obstacle: specificity and matched-control completeness
also fail. We do not change thresholds after seeing these outcomes. I3
component/path localization is therefore
\texttt{NOT\_RUN\_GATE\_NOT\_MET}, and cross-scale internal validation is
\texttt{NOT\_RUN\_MECHANISM\_LOCK\_REQUIRED}. Three-model behavioral results
and technical acceptance cannot substitute for those unrun causal
experiments. The decision constrains the tested mapping and interventions;
it is neither proof that no local mechanism exists nor evidence for a
specific distributed mechanism.

\FloatBarrier
\section{Representative matched cases}
\label{app:sc_qualitative}

The cases in Table~\ref{tab:sc_qualitative} are drawn by the fixed hash rules
recorded in the evidence report, not by selecting the largest improvements.
Count cases are selected within the indicated Type A category; the non-count
case is selected within the 9B Type-A-like category; the target-wording
program is selected without conditioning on its improvement; and the I1 case
comes from the primary SELECT cohort. These cases illustrate already
quantified patterns and are not additional independent experiments. Full
image data are not reproduced here, so these transcripts are not a new
visual re-audit of the gold values.

\begin{table}[htbp]
\centering
\caption{Fixed-rule representative outputs. The first three rows compare independent direct, full, and explicit-state requests. The fourth compares target wordings, and the fifth shows frozen activation interventions. Exact prompt instances appear in Appendix~\ref{app:sc_prompts}.}
\label{tab:sc_qualitative}
\footnotesize
\setlength{\tabcolsep}{4pt}
\begin{adjustbox}{max width=\linewidth}
\begin{tabular}{L{0.14\linewidth}L{0.19\linewidth}L{0.39\linewidth}L{0.20\linewidth}}
\toprule
Case & Gold state(s) & Observed outputs & Interpretation \\
\midrule
9B count & $s_0,s_1,s_2=1,3,6$ & Direct 1; full 5; explicit 6; legacy register 5 & Type A \\
27B count & $s_0,s_1,s_2=2,1,2$ & Direct 2; full 1; explicit 2; legacy register 1 & Type A; inverse program \\
9B horizontal & Left $\to$ right & Direct left; full left; explicit right; sham left & Matched relation contrast \\
Target wording & $s_0,s_1,s_2=9,13,9$ & S1 under T1: 9; under T2/T3/T4: 13 & Target wording matters \\
9B I1 SELECT & $s_2=1$ & Baseline 2; block-5 patch 2; block-9 patch 2 & No rescue \\
\bottomrule
\end{tabular}
\end{adjustbox}
\end{table}

In the first count case, one cabinet followed by add-two and add-three gives
$s_2=6$, but the full answer is 5 despite a correct separate initial answer.
The second case returns to its initial count after a remove/add inverse;
correctly answering it does not establish that both intermediate transitions
were represented. In the relation case, the image describes $s_0$, while the
question asks for the reflected coordinate frame; an unchanged visual
relation is therefore incorrect for the transformed query. For the target
program, $s_0=s_2=9$, so the wrong answer 9 cannot be uniquely attributed to
one of those states. Finally, unchanged I1 outputs are consistent with the
aggregate lack of validated rescue and must not be presented as an effective
intervention.

\FloatBarrier
\section{Interpretation boundaries and untested alternatives}
\label{app:sc_limits}

The experimental findings support a distinction between independently
reportable facts and success on matched transformation queries. They do not
show that the correct initial state was present in the failed run's internal
computation, nor do they locate a unique perception, binding, initialization,
transition, or readout failure. Oracle input changes information availability
and linguistic task organization at the same time. W1 demonstrates this
sensitivity for one symbolic query anomaly; it does not eliminate every
prompt-related explanation for B1 or the non-count tasks.

The explicit-state contrast is not uniform across models and relations.
4B includes harmed count successes and lacks the overall positive non-count
explicit-minus-sham result; 9B's depth subset is a counterexample to its
aggregate non-count improvement. The non-count program contains only one
nontrivial reflection. Consequently, neither the test scale nor the state
operations support a claim about arbitrary three-dimensional dynamics or
embodied planning.

Representation readability and causal efficacy are different measurements.
The prespecified negative probe comparison coexists with more readable
variables and oracle contexts in the full ledger. I2 does not validate the
specified intermediate-state interchange, and I1 does not pass its frozen
validation gate. Non-significant intervals are not equivalence tests proving
zero effect. Limited donor coverage, engine-equivalence exclusions, small
cluster counts, and the chosen single-token intervention family remain
material limits. Conversely, no damage without successful rescue is not
positive evidence for selective repair.

The evidence report distinguishes automatic provisional review from human
verification, and no new human verification is claimed for these diagnostics.
All-split coverage, existing test-only scores, diagnostic discovery worlds,
and window-validation partitions have different roles. Post-hoc descriptive
family and case summaries remain labeled as such. The results below retain
negative, zero, and unrun outcomes; they do not infer a compact state circuit
or a distributed replacement merely to supply a unified mechanism story.

\FloatBarrier
\section{Exact prompt instances and presentation}
\label{app:sc_prompts}

The prompt instances below are transcribed from the recorded requests, not
rewritten from an intended experimental design. They retain the original
English wording, including grammatical irregularities and literal state
labels. They are instantiated examples rather than newly invented placeholder
templates. Formatting line wraps in the typeset listings are for readability;
the supplied text is otherwise preserved. Ground-truth values printed outside
the boxes are explanatory metadata and were not appended to ordinary model
inputs. Oracle values inside an oracle-condition box were deliberately part
of that condition.

Table~\ref{tab:sc_prompt_index} links the prompt groups to their corresponding
result sections. The shared system instruction is used by the diagnostic
value tasks; E0 has its own rendered label/confidence/reason instruction.
This separation is essential because changing an output contract can change
what is scoreable without establishing a change in spatial ability.

\begin{table}[htbp]
\centering
\caption{Prompt and result cross-reference. All groups below contain actual run text; the historical B00 request is explicitly unavailable and is not reconstructed.}
\label{tab:sc_prompt_index}
\small
\setlength{\tabcolsep}{4pt}
\begin{adjustbox}{max width=\linewidth}
\begin{tabular}{lll}
\toprule
Prompt group & Prompt appendix & Corresponding results \\
\midrule
Overall evaluation & Appendix~\ref{app:sc_prompts_e0} & Appendix~\ref{app:sc_full_results} \\
B1 count diagnostics & Appendix~\ref{app:sc_prompts_count} & Appendices~\ref{app:sc_behavior}, \ref{app:sc_cases} \\
Target-semantic controls & Appendix~\ref{app:sc_prompts_target} & Appendix~\ref{app:sc_target} \\
Non-count relations & Appendix~\ref{app:sc_prompts_noncount} & Appendix~\ref{app:sc_noncount} \\
Intermediate-state interchange & Appendix~\ref{app:sc_prompts_i2} & Appendix~\ref{app:sc_i2} \\
Initial-state donor interventions & Appendix~\ref{app:sc_prompts_i1} & Appendix~\ref{app:sc_i1} \\
\bottomrule
\end{tabular}
\end{adjustbox}
\end{table}

\subsection{Shared diagnostic system instruction}
\label{app:sc_prompt_system}

The diagnostic system instruction is the following literal text.

\begin{scprompt}
You are evaluating spatial facts from supplied observations. Use the specified reference image, scope and named state. Candidate statements are propositions to check, not observations. An irrelevant register describes a separate bookkeeping entry, not the scene. Hypothetical actions change only their named branch. Preserve the initial state and unrelated facts. If the supplied evidence does not determine the requested value, return null. Return only the requested JSON, with no reasoning or markdown, within 512 output tokens.
\end{scprompt}

In each user message, media are inserted in their recorded order before the
payload text. The shared construction prefixes media with
\texttt{Evidence image $i$ (role):} and then inserts the image block.
The diagnostic pipeline replaces the final text block with its own payload,
so the E0 label/confidence/reason suffix is not accidentally appended to a
value-only request. Thinking is disabled by the diagnostic chat-template call.
The original reference/support roles specify counting scope, not multiple
independent counts. These prompt transcripts do not constitute an additional visual audit
of the underlying examples.

\subsection{Overall evaluation: rendered prompt examples}
\label{app:sc_prompts_e0}

The following four rendered examples expose the actual E0 input contract at
each level. The visible chat and image markers are retained. They illustrate
the evaluation interface, not a new definition of the benchmark levels.
All use the same 512-token retained-output instruction. Empty reasoning tags
in a rendered example are not a generated reasoning trace.

\subsubsection{E0 rendered example: L1}
\label{app:sc_prompt_e0_l1}

\begin{scprompt}
<|im_start|>system
You are evaluating a spatial claim using only the supplied visual evidence and, when present, the stated hypothetical intervention. Classify the claim as exactly one of SUPPORTED, CONTRADICTORY, or UNKNOWN. SUPPORTED means the accessible evidence entails the claim. CONTRADICTORY means it explicitly refutes the claim. UNKNOWN means both the claim and its negation remain possible from the accessible evidence. Missing information is not negative evidence. Return JSON only.<|im_end|>
<|im_start|>user
Evidence image 1 (frame_0):<|vision_start|><|image_pad|><|vision_end|>Source bounding-box coordinates, when written in the claim, use the normalized 0–1000 xyxy scale. Colored outlines identify only the source-annotated targets in each supplied view. frame_0 is the primary view; other frame numbers retain their original source indices.

Spatial claim:
Let entity A denote the blue bbox target, and let entity B denote the red bbox target. In the current image of the actual scene, from the perspective declared in the source question, the entity A is below the entity B.

Accessible evidence: 1 media item(s) supplied above.

Return exactly one JSON object with this schema:
{"label":"SUPPORTED|CONTRADICTORY|UNKNOWN","confidence":0.0,"reason":"brief explanation"}

Output formatting requirement: Return only the JSON object itself. Do not use Markdown, backticks, code fences, or text outside the object. Start with { and end with }. Use exactly the keys label, confidence, and reason. Your entire response, including the answer and explanation, must not exceed 512 output tokens. Be concise and finish the JSON within this limit. Output label first, then confidence, then reason. Generation stops at 512 tokens; only the retained output is scored, and any content beyond the limit is ignored.<|im_end|>
<|im_start|>assistant
<think>

</think>
\end{scprompt}

\subsubsection{E0 rendered example: L2}
\label{app:sc_prompt_e0_l2}

\begin{scprompt}
<|im_start|>system
You are evaluating a spatial claim using only the supplied visual evidence and, when present, the stated hypothetical intervention. Classify the claim as exactly one of SUPPORTED, CONTRADICTORY, or UNKNOWN. SUPPORTED means the accessible evidence entails the claim. CONTRADICTORY means it explicitly refutes the claim. UNKNOWN means both the claim and its negation remain possible from the accessible evidence. Missing information is not negative evidence. Return JSON only.<|im_end|>
<|im_start|>user
Evidence image 1 (frame_0):<|vision_start|><|image_pad|><|vision_end|>Source bounding-box coordinates, when written in the claim, use the normalized 0–1000 xyxy scale. Colored outlines identify only the source-annotated targets in each supplied view. frame_0 is the primary view; other frame numbers retain their original source indices.

Spatial claim:
Let entity A denote the object marked by bounding box [472, 171, 999, 857], and let entity B denote the object marked by bounding box [0, 2, 382, 993]. The entity A is to the left of the entity B in the current image of the actual scene, from the perspective declared in the source question.

Accessible evidence: 1 media item(s) supplied above.

Return exactly one JSON object with this schema:
{"label":"SUPPORTED|CONTRADICTORY|UNKNOWN","confidence":0.0,"reason":"brief explanation"}

Output formatting requirement: Return only the JSON object itself. Do not use Markdown, backticks, code fences, or text outside the object. Start with { and end with }. Use exactly the keys label, confidence, and reason. Your entire response, including the answer and explanation, must not exceed 512 output tokens. Be concise and finish the JSON within this limit. Output label first, then confidence, then reason. Generation stops at 512 tokens; only the retained output is scored, and any content beyond the limit is ignored.<|im_end|>
<|im_start|>assistant
<think>

</think>
\end{scprompt}

\subsubsection{E0 rendered example: L3}
\label{app:sc_prompt_e0_l3}

\begin{scprompt}
<|im_start|>system
You are evaluating a spatial claim using only the supplied visual evidence and, when present, the stated hypothetical intervention. Classify the claim as exactly one of SUPPORTED, CONTRADICTORY, or UNKNOWN. SUPPORTED means the accessible evidence entails the claim. CONTRADICTORY means it explicitly refutes the claim. UNKNOWN means both the claim and its negation remain possible from the accessible evidence. Missing information is not negative evidence. Return JSON only.<|im_end|>
<|im_start|>user
Evidence image 1 (frame_0):<|vision_start|><|image_pad|><|vision_end|>Evidence image 2 (frame_1):<|vision_start|><|image_pad|><|vision_end|>Evidence image 3 (frame_2):<|vision_start|><|image_pad|><|vision_end|>Source bounding-box coordinates, when written in the claim, use the normalized 0–1000 xyxy scale. Colored outlines identify only the source-annotated targets in each supplied view. frame_0 is the primary view; other frame numbers retain their original source indices.

Spatial claim:
Let entity A denote the red bbox target, and let entity B denote the observer primary view. In the current image of the actual scene, from the perspective declared in the source question, the entity B is in front of the entity A.

Accessible evidence: 3 media item(s) supplied above.

Return exactly one JSON object with this schema:
{"label":"SUPPORTED|CONTRADICTORY|UNKNOWN","confidence":0.0,"reason":"brief explanation"}

Output formatting requirement: Return only the JSON object itself. Do not use Markdown, backticks, code fences, or text outside the object. Start with { and end with }. Use exactly the keys label, confidence, and reason. Your entire response, including the answer and explanation, must not exceed 512 output tokens. Be concise and finish the JSON within this limit. Output label first, then confidence, then reason. Generation stops at 512 tokens; only the retained output is scored, and any content beyond the limit is ignored.<|im_end|>
<|im_start|>assistant
<think>

</think>
\end{scprompt}

\subsubsection{E0 rendered example: L4}
\label{app:sc_prompt_e0_l4}

\begin{scprompt}
<|im_start|>system
You are evaluating a spatial claim using only the supplied visual evidence and, when present, the stated hypothetical intervention. Classify the claim as exactly one of SUPPORTED, CONTRADICTORY, or UNKNOWN. SUPPORTED means the accessible evidence entails the claim. CONTRADICTORY means it explicitly refutes the claim. UNKNOWN means both the claim and its negation remain possible from the accessible evidence. Missing information is not negative evidence. Return JSON only.<|im_end|>
<|im_start|>user
Evidence image 1 (camera_view):<|vision_start|><|image_pad|><|vision_end|>Evidence image 2 (top_view_label):<|vision_start|><|image_pad|><|vision_end|>Evidence image 3 (top_view_no_label):<|vision_start|><|image_pad|><|vision_end|>Evidence image 4 (top_view_no_label_rotated):<|vision_start|><|image_pad|><|vision_end|>Evidence image 5 (top_view_with_label_rotated):<|vision_start|><|image_pad|><|vision_end|>Hypothetical intervention:
The towel beside the light has been removed.

Spatial claim:
After the change, there are 1 towel in the scene.

Accessible visual evidence: 5 image(s) supplied above.

Return exactly one JSON object with this schema:
{"label":"SUPPORTED|CONTRADICTORY|UNKNOWN","confidence":0.0,"reason":"brief explanation"}

Output formatting requirement: Return only the JSON object itself. Do not use Markdown, backticks, code fences, or text outside the object. Start with { and end with }. Use exactly the keys label, confidence, and reason. Your entire response, including the answer and explanation, must not exceed 512 output tokens. Be concise and finish the JSON within this limit. Output label first, then confidence, then reason. Generation stops at 512 tokens; only the retained output is scored, and any content beyond the limit is ignored.<|im_end|>
<|im_start|>assistant
<think>

</think>
\end{scprompt}

\subsection{Count diagnostic conditions}
\label{app:sc_prompts_count}

These instances use the same recorded cabinet-count world as the first case
in Table~\ref{tab:sc_qualitative}. Its queried gold states are
$s_0=1$, $s_1=3$, and $s_2=6$, with add-two followed by add-three. Visual
conditions present five ordered images: one reference frame and four support
frames; symbolic conditions contain no images. The system instruction is
Appendix~\ref{app:sc_prompt_system}. Each condition's expected value is
reported outside the prompt box. The protected category in these examples
is mirrors, whose recorded initial count is one.

\subsubsection{B00: historical reused request}
\label{app:sc_prompt_b00}

The historical score is retained in the B1 tables, but its exact request is unavailable in the archived public batch. No prompt is reconstructed.

\subsubsection{B01}
\label{app:sc_prompt_b01}

Media blocks: 5. Recorded expected value: 6. The condition is defined in Table~\ref{tab:sc_conditions}.

\begin{scprompt}
OBSERVED STATE S0: the supplied reference image. Support views do not add members to the counting scope.
IMAGE ORDER: 1=reference_frame, 2=support_frame_1, 3=support_frame_2, 4=support_frame_3, 5=support_frame_4
OBJECT: cabinets. COUNTING SCOPE: the cabinets counted in the reference image, not a whole-room census.
The following two actions are cumulative on one hypothetical counted set. A1 maps S0 to S1; A2 acts on S1 to produce S2, not independently on S0. The recorded S0 and all other categories remain unchanged. Only quantities are queried; do not recount occlusion after the edit.
ACTION_1: add exactly 2 new cabinets.
STATE_NOTE: No state value is supplied here.
ACTION_2: add exactly 3 new cabinets.
TARGET: S2
QUERY: How many cabinets are in the reference-image counting scope in state S2?
OUTPUT CONTRACT: Return exactly one JSON object with key "value". Its value is a nonnegative JSON integer, NOT a quoted string; use the JSON literal null, not the string "null", when undetermined. Even an undetermined answer must use the object {"value":null}, not bare null.
\end{scprompt}

\subsubsection{B02}
\label{app:sc_prompt_b02}

Media blocks: 5. Recorded expected value: 1. The condition is defined in Table~\ref{tab:sc_conditions}.

\begin{scprompt}
OBSERVED STATE S0: the supplied reference image. Support views do not add members to the counting scope.
IMAGE ORDER: 1=reference_frame, 2=support_frame_1, 3=support_frame_2, 4=support_frame_3, 5=support_frame_4
OBJECT: cabinets. COUNTING SCOPE: the cabinets counted in the reference image, not a whole-room census.
TARGET: S0
QUERY: How many cabinets are in the reference-image counting scope in state S0?
OUTPUT CONTRACT: Return exactly one JSON object with key "value". Its value is a nonnegative JSON integer, NOT a quoted string; use the JSON literal null, not the string "null", when undetermined. Even an undetermined answer must use the object {"value":null}, not bare null.
\end{scprompt}

\subsubsection{B03}
\label{app:sc_prompt_b03}

Media blocks: 5. Recorded expected value: 3. The condition is defined in Table~\ref{tab:sc_conditions}.

\begin{scprompt}
OBSERVED STATE S0: the supplied reference image. Support views do not add members to the counting scope.
IMAGE ORDER: 1=reference_frame, 2=support_frame_1, 3=support_frame_2, 4=support_frame_3, 5=support_frame_4
OBJECT: cabinets. COUNTING SCOPE: the cabinets counted in the reference image, not a whole-room census.
The following two actions are cumulative on one hypothetical counted set. A1 maps S0 to S1; A2 acts on S1 to produce S2, not independently on S0. The recorded S0 and all other categories remain unchanged. Only quantities are queried; do not recount occlusion after the edit.
ACTION_1: add exactly 2 new cabinets.
STATE_NOTE: No state value is supplied here.
ACTION_2: add exactly 3 new cabinets.
TARGET: S1
QUERY: How many cabinets are in the reference-image counting scope in state S1?
OUTPUT CONTRACT: Return exactly one JSON object with key "value". Its value is a nonnegative JSON integer, NOT a quoted string; use the JSON literal null, not the string "null", when undetermined. Even an undetermined answer must use the object {"value":null}, not bare null.
\end{scprompt}

\subsubsection{B04}
\label{app:sc_prompt_b04}

Media blocks: 5. Recorded expected value: \texttt{S1}. The condition is defined in Table~\ref{tab:sc_conditions}.

\begin{scprompt}
OBSERVED STATE S0: the supplied reference image. Support views do not add members to the counting scope.
IMAGE ORDER: 1=reference_frame, 2=support_frame_1, 3=support_frame_2, 4=support_frame_3, 5=support_frame_4
OBJECT: cabinets. COUNTING SCOPE: the cabinets counted in the reference image, not a whole-room census.
The following two actions are cumulative on one hypothetical counted set. A1 maps S0 to S1; A2 acts on S1 to produce S2, not independently on S0. The recorded S0 and all other categories remain unchanged. Only quantities are queried; do not recount occlusion after the edit.
ACTION_1: add exactly 2 new cabinets.
STATE_NOTE: No state value is supplied here.
ACTION_2: add exactly 3 new cabinets.
QUERY: Which named state is the input to ACTION_2? Return its state name.
OUTPUT CONTRACT: Return exactly one JSON object with key "value". Its value is one of these exact strings: ["S0", "S1", "S2"]; use the JSON literal null, not the string "null", when undetermined. Even an undetermined answer must use the object {"value":null}, not bare null.
\end{scprompt}

\subsubsection{B05}
\label{app:sc_prompt_b05}

Media blocks: 5. Recorded expected value: 6. The condition is defined in Table~\ref{tab:sc_conditions}.

\begin{scprompt}
OBSERVED STATE S0: the supplied reference image. Support views do not add members to the counting scope.
IMAGE ORDER: 1=reference_frame, 2=support_frame_1, 3=support_frame_2, 4=support_frame_3, 5=support_frame_4
OBJECT: cabinets. COUNTING SCOPE: the cabinets counted in the reference image, not a whole-room census.
ORACLE INITIAL FACT: the target count in S0 is exactly 1.
The following two actions are cumulative on one hypothetical counted set. A1 maps S0 to S1; A2 acts on S1 to produce S2, not independently on S0. The recorded S0 and all other categories remain unchanged. Only quantities are queried; do not recount occlusion after the edit.
ACTION_1: add exactly 2 new cabinets.
STATE_NOTE: No state value is supplied here.
ACTION_2: add exactly 3 new cabinets.
TARGET: S2
QUERY: How many cabinets are in the reference-image counting scope in state S2?
OUTPUT CONTRACT: Return exactly one JSON object with key "value". Its value is a nonnegative JSON integer, NOT a quoted string; use the JSON literal null, not the string "null", when undetermined. Even an undetermined answer must use the object {"value":null}, not bare null.
\end{scprompt}

\subsubsection{B06}
\label{app:sc_prompt_b06}

Media blocks: 5. Recorded expected value: 6. The condition is defined in Table~\ref{tab:sc_conditions}.

\begin{scprompt}
OBSERVED STATE S0: the supplied reference image. Support views do not add members to the counting scope.
IMAGE ORDER: 1=reference_frame, 2=support_frame_1, 3=support_frame_2, 4=support_frame_3, 5=support_frame_4
OBJECT: cabinets. COUNTING SCOPE: the cabinets counted in the reference image, not a whole-room census.
The following two actions are cumulative on one hypothetical counted set. A1 maps S0 to S1; A2 acts on S1 to produce S2, not independently on S0. The recorded S0 and all other categories remain unchanged. Only quantities are queried; do not recount occlusion after the edit.
ACTION_1: add exactly 2 new cabinets.
STATE_NOTE: ORACLE CHECKPOINT: the target count in S1 is exactly 3.
ACTION_2: add exactly 3 new cabinets.
TARGET: S2
QUERY: How many cabinets are in the reference-image counting scope in state S2?
OUTPUT CONTRACT: Return exactly one JSON object with key "value". Its value is a nonnegative JSON integer, NOT a quoted string; use the JSON literal null, not the string "null", when undetermined. Even an undetermined answer must use the object {"value":null}, not bare null.
\end{scprompt}

\subsubsection{B07}
\label{app:sc_prompt_b07}

Media blocks: 5. Recorded expected value: 6. The condition is defined in Table~\ref{tab:sc_conditions}.

\begin{scprompt}
OBSERVED STATE S0: the supplied reference image. Support views do not add members to the counting scope.
IMAGE ORDER: 1=reference_frame, 2=support_frame_1, 3=support_frame_2, 4=support_frame_3, 5=support_frame_4
OBJECT: cabinets. COUNTING SCOPE: the cabinets counted in the reference image, not a whole-room census.
The following two actions are cumulative on one hypothetical counted set. A1 maps S0 to S1; A2 acts on S1 to produce S2, not independently on S0. The recorded S0 and all other categories remain unchanged. Only quantities are queried; do not recount occlusion after the edit.
ACTION_1: add exactly 2 new cabinets.
STATE_NOTE: IRRELEVANT REGISTER: an unrelated bookkeeping entry is exactly 3. This is not a scene count or state update.
ACTION_2: add exactly 3 new cabinets.
TARGET: S2
QUERY: How many cabinets are in the reference-image counting scope in state S2?
OUTPUT CONTRACT: Return exactly one JSON object with key "value". Its value is a nonnegative JSON integer, NOT a quoted string; use the JSON literal null, not the string "null", when undetermined. Even an undetermined answer must use the object {"value":null}, not bare null.
\end{scprompt}

\subsubsection{B08}
\label{app:sc_prompt_b08}

Media blocks: 0. Recorded expected value: 6. The condition is defined in Table~\ref{tab:sc_conditions}.

\begin{scprompt}
SYMBOLIC PROGRAM ONLY; no image evidence is required. The target counted set initially has exactly 1 members in S0.
The following two actions are cumulative on one hypothetical counted set. A1 maps S0 to S1; A2 acts on S1 to produce S2, not independently on S0. The recorded S0 and all other categories remain unchanged. Only quantities are queried; do not recount occlusion after the edit.
ACTION_1: add exactly 2 new members.
ACTION_2: add exactly 3 new members.
QUERY: What is the count in S2?
OUTPUT CONTRACT: Return exactly one JSON object with key "value". Its value is a nonnegative JSON integer, NOT a quoted string; use the JSON literal null, not the string "null", when undetermined. Even an undetermined answer must use the object {"value":null}, not bare null.
\end{scprompt}

\subsubsection{B09}
\label{app:sc_prompt_b09}

Media blocks: 0. Recorded expected value: 6. The condition is defined in Table~\ref{tab:sc_conditions}.

\begin{scprompt}
EXPLICIT SYMBOLIC INPUT: state X has exactly 3 members. Apply one action to X: add exactly 3 new members.
QUERY: What is the resulting count?
OUTPUT CONTRACT: Return exactly one JSON object with key "value". Its value is a nonnegative JSON integer, NOT a quoted string; use the JSON literal null, not the string "null", when undetermined. Even an undetermined answer must use the object {"value":null}, not bare null.
\end{scprompt}

\subsubsection{B10}
\label{app:sc_prompt_b10}

Media blocks: 5. Recorded expected value: 1. The condition is defined in Table~\ref{tab:sc_conditions}.

\begin{scprompt}
OBSERVED STATE S0: the supplied reference image. Support views do not add members to the counting scope.
IMAGE ORDER: 1=reference_frame, 2=support_frame_1, 3=support_frame_2, 4=support_frame_3, 5=support_frame_4
OBJECT: cabinets. COUNTING SCOPE: the cabinets counted in the reference image, not a whole-room census.
ORACLE COMPLETE STATE TABLE: S0=1; S1=3; S2=6.
TARGET: S0
QUERY: How many cabinets are in the reference-image counting scope in state S0?
OUTPUT CONTRACT: Return exactly one JSON object with key "value". Its value is a nonnegative JSON integer, NOT a quoted string; use the JSON literal null, not the string "null", when undetermined. Even an undetermined answer must use the object {"value":null}, not bare null.
\end{scprompt}

\subsubsection{B11}
\label{app:sc_prompt_b11}

Media blocks: 5. Recorded expected value: 3. The condition is defined in Table~\ref{tab:sc_conditions}.

\begin{scprompt}
OBSERVED STATE S0: the supplied reference image. Support views do not add members to the counting scope.
IMAGE ORDER: 1=reference_frame, 2=support_frame_1, 3=support_frame_2, 4=support_frame_3, 5=support_frame_4
OBJECT: cabinets. COUNTING SCOPE: the cabinets counted in the reference image, not a whole-room census.
ORACLE COMPLETE STATE TABLE: S0=1; S1=3; S2=6.
TARGET: S1
QUERY: How many cabinets are in the reference-image counting scope in state S1?
OUTPUT CONTRACT: Return exactly one JSON object with key "value". Its value is a nonnegative JSON integer, NOT a quoted string; use the JSON literal null, not the string "null", when undetermined. Even an undetermined answer must use the object {"value":null}, not bare null.
\end{scprompt}

\subsubsection{B12}
\label{app:sc_prompt_b12}

Media blocks: 5. Recorded expected value: 6. The condition is defined in Table~\ref{tab:sc_conditions}.

\begin{scprompt}
OBSERVED STATE S0: the supplied reference image. Support views do not add members to the counting scope.
IMAGE ORDER: 1=reference_frame, 2=support_frame_1, 3=support_frame_2, 4=support_frame_3, 5=support_frame_4
OBJECT: cabinets. COUNTING SCOPE: the cabinets counted in the reference image, not a whole-room census.
ORACLE COMPLETE STATE TABLE: S0=1; S1=3; S2=6.
TARGET: S2
QUERY: How many cabinets are in the reference-image counting scope in state S2?
OUTPUT CONTRACT: Return exactly one JSON object with key "value". Its value is a nonnegative JSON integer, NOT a quoted string; use the JSON literal null, not the string "null", when undetermined. Even an undetermined answer must use the object {"value":null}, not bare null.
\end{scprompt}

\subsubsection{B13}
\label{app:sc_prompt_b13}

Media blocks: 0. Recorded expected value: 5. The condition is defined in Table~\ref{tab:sc_conditions}.

\begin{scprompt}
COUNTERFACTUAL SYMBOLIC INPUT: This is a separate hypothetical input, not an observation of the image.
EXPLICIT SYMBOLIC INPUT: state X has exactly 2 members. Apply one action to X: add exactly 3 new members.
QUERY: What is the resulting count?
OUTPUT CONTRACT: Return exactly one JSON object with key "value". Its value is a nonnegative JSON integer, NOT a quoted string; use the JSON literal null, not the string "null", when undetermined. Even an undetermined answer must use the object {"value":null}, not bare null.
\end{scprompt}

\subsubsection{B14}
\label{app:sc_prompt_b14}

Media blocks: 0. Recorded expected value: 7. The condition is defined in Table~\ref{tab:sc_conditions}.

\begin{scprompt}
COUNTERFACTUAL SYMBOLIC INPUT: This is a separate hypothetical input, not an observation of the image.
EXPLICIT SYMBOLIC INPUT: state X has exactly 4 members. Apply one action to X: add exactly 3 new members.
QUERY: What is the resulting count?
OUTPUT CONTRACT: Return exactly one JSON object with key "value". Its value is a nonnegative JSON integer, NOT a quoted string; use the JSON literal null, not the string "null", when undetermined. Even an undetermined answer must use the object {"value":null}, not bare null.
\end{scprompt}

\subsubsection{B15}
\label{app:sc_prompt_b15}

Media blocks: 5. Recorded expected value: 1. The condition is defined in Table~\ref{tab:sc_conditions}.

\begin{scprompt}
OBSERVED STATE S0: the supplied reference image. Support views do not add members to the counting scope.
IMAGE ORDER: 1=reference_frame, 2=support_frame_1, 3=support_frame_2, 4=support_frame_3, 5=support_frame_4
OBJECT: cabinets. COUNTING SCOPE: the cabinets counted in the reference image, not a whole-room census.
The following two actions are cumulative on one hypothetical counted set. A1 maps S0 to S1; A2 acts on S1 to produce S2, not independently on S0. The recorded S0 and all other categories remain unchanged. Only quantities are queried; do not recount occlusion after the edit.
ACTION_1: add exactly 2 new cabinets.
STATE_NOTE: No state value is supplied here.
ACTION_2: add exactly 3 new cabinets.
TARGET: S0
QUERY: How many cabinets are in the reference-image counting scope in state S0?
OUTPUT CONTRACT: Return exactly one JSON object with key "value". Its value is a nonnegative JSON integer, NOT a quoted string; use the JSON literal null, not the string "null", when undetermined. Even an undetermined answer must use the object {"value":null}, not bare null.
\end{scprompt}

\subsubsection{B16}
\label{app:sc_prompt_b16}

Media blocks: 5. Recorded expected value: 1. The condition is defined in Table~\ref{tab:sc_conditions}.

\begin{scprompt}
OBSERVED STATE S0: the supplied reference image. Support views do not add members to the counting scope.
IMAGE ORDER: 1=reference_frame, 2=support_frame_1, 3=support_frame_2, 4=support_frame_3, 5=support_frame_4
OBJECT: cabinets. COUNTING SCOPE: the cabinets counted in the reference image, not a whole-room census.
The following two actions are cumulative on one hypothetical counted set. A1 maps S0 to S1; A2 acts on S1 to produce S2, not independently on S0. The recorded S0 and all other categories remain unchanged. Only quantities are queried; do not recount occlusion after the edit.
ACTION_1: add exactly 2 new cabinets.
STATE_NOTE: No state value is supplied here.
ACTION_2: add exactly 3 new cabinets.
TARGET: S2
QUERY: How many mirrors are in the reference-image counting scope in state S2?
OUTPUT CONTRACT: Return exactly one JSON object with key "value". Its value is a nonnegative JSON integer, NOT a quoted string; use the JSON literal null, not the string "null", when undetermined. Even an undetermined answer must use the object {"value":null}, not bare null.
\end{scprompt}

\subsubsection{B17}
\label{app:sc_prompt_b17}

Media blocks: 5. Recorded expected value: 3. The condition is defined in Table~\ref{tab:sc_conditions}.

\begin{scprompt}
OBSERVED STATE S0: the supplied reference image. Support views do not add members to the counting scope.
IMAGE ORDER: 1=reference_frame, 2=support_frame_1, 3=support_frame_2, 4=support_frame_3, 5=support_frame_4
OBJECT: cabinets. COUNTING SCOPE: the cabinets counted in the reference image, not a whole-room census.
Apply only the following first action to S0 to obtain S1. All other categories remain unchanged; do not recount occlusion.
ACTION_1: add exactly 2 new cabinets.
TARGET: S1
QUERY: How many cabinets are in the reference-image counting scope in state S1?
OUTPUT CONTRACT: Return exactly one JSON object with key "value". Its value is a nonnegative JSON integer, NOT a quoted string; use the JSON literal null, not the string "null", when undetermined. Even an undetermined answer must use the object {"value":null}, not bare null.
\end{scprompt}

\subsubsection{PROTECTED\_PRE\_SUPPLEMENT}
\label{app:sc_prompt_protected_pre_supplement}

Media blocks: 5. Recorded expected value: 1. The condition is defined in Table~\ref{tab:sc_conditions}.

\begin{scprompt}
OBSERVED STATE S0: the supplied reference image. Support views do not add members to the counting scope.
IMAGE ORDER: 1=reference_frame, 2=support_frame_1, 3=support_frame_2, 4=support_frame_3, 5=support_frame_4
OBJECT: cabinets. COUNTING SCOPE: the cabinets counted in the reference image, not a whole-room census.
TARGET: S0
QUERY: How many mirrors are in the reference-image counting scope in state S0?
OUTPUT CONTRACT: Return exactly one JSON object with key "value". Its value is a nonnegative JSON integer, NOT a quoted string; use the JSON literal null, not the string "null", when undetermined. Even an undetermined answer must use the object {"value":null}, not bare null.
\end{scprompt}

\subsection{Target wording: all twelve query variants}
\label{app:sc_prompts_target}

All twelve examples below use one fixed symbolic program with
$s_0=9$, $s_1=13$, and $s_2=9$: add four, then remove four. They contain no
media and share the system instruction in
Appendix~\ref{app:sc_prompt_system}. T2 removes the index-only target in
favor of temporal wording; T3 defines the states; T4 adds the final reminder.
The output schema is unchanged. The result tables in
Appendix~\ref{app:sc_target} quantify these changes over all 96 programs,
not just this example.

\subsubsection{T1\_S0}
\label{app:sc_prompt_w1_t1_s0}

Recorded expected value: 9.

\begin{scprompt}
NON-SPATIAL SYMBOLIC REGISTER R. Its initial value in S0 is 9. Actions are cumulative; A2 acts on the result of A1.
ACTION_1: add exactly 4 new register units.
ACTION_2: remove exactly 4 counted register units.
TARGET: S0. QUERY: What is the value of register R at the target state?
OUTPUT CONTRACT: Return exactly one JSON object with key "value". Its value is a nonnegative JSON integer, NOT a quoted string; use the JSON literal null, not the string "null", when undetermined. Even an undetermined answer must use the object {"value":null}, not bare null.
\end{scprompt}

\subsubsection{T2\_S0}
\label{app:sc_prompt_w1_t2_s0}

Recorded expected value: 9.

\begin{scprompt}
NON-SPATIAL SYMBOLIC REGISTER R. Its initial value in S0 is 9. Actions are cumulative; A2 acts on the result of A1.
ACTION_1: add exactly 4 new register units.
ACTION_2: remove exactly 4 counted register units.
QUERY: What is the value of register R before Action 1 and before Action 2?
OUTPUT CONTRACT: Return exactly one JSON object with key "value". Its value is a nonnegative JSON integer, NOT a quoted string; use the JSON literal null, not the string "null", when undetermined. Even an undetermined answer must use the object {"value":null}, not bare null.
\end{scprompt}

\subsubsection{T3\_S0}
\label{app:sc_prompt_w1_t3_s0}

Recorded expected value: 9.

\begin{scprompt}
NON-SPATIAL SYMBOLIC REGISTER R. Its initial value in S0 is 9. Actions are cumulative; A2 acts on the result of A1.
ACTION_1: add exactly 4 new register units.
ACTION_2: remove exactly 4 counted register units.
STATE DEFINITIONS: S0 is before both actions. S1 is immediately after Action 1 and before Action 2. S2 is immediately after Action 2, applied to S1.
TARGET: S0. QUERY: What is the value of register R at the target state?
OUTPUT CONTRACT: Return exactly one JSON object with key "value". Its value is a nonnegative JSON integer, NOT a quoted string; use the JSON literal null, not the string "null", when undetermined. Even an undetermined answer must use the object {"value":null}, not bare null.
\end{scprompt}

\subsubsection{T4\_S0}
\label{app:sc_prompt_w1_t4_s0}

Recorded expected value: 9.

\begin{scprompt}
NON-SPATIAL SYMBOLIC REGISTER R. Its initial value in S0 is 9. Actions are cumulative; A2 acts on the result of A1.
ACTION_1: add exactly 4 new register units.
ACTION_2: remove exactly 4 counted register units.
STATE DEFINITIONS: S0 is before both actions. S1 is immediately after Action 1 and before Action 2. S2 is immediately after Action 2, applied to S1.
TARGET: S0. QUERY: What is the value of register R at the target state?
Do not return S2. Return S0 only.
OUTPUT CONTRACT: Return exactly one JSON object with key "value". Its value is a nonnegative JSON integer, NOT a quoted string; use the JSON literal null, not the string "null", when undetermined. Even an undetermined answer must use the object {"value":null}, not bare null.
\end{scprompt}

\subsubsection{T1\_S1}
\label{app:sc_prompt_w1_t1_s1}

Recorded expected value: 13.

\begin{scprompt}
NON-SPATIAL SYMBOLIC REGISTER R. Its initial value in S0 is 9. Actions are cumulative; A2 acts on the result of A1.
ACTION_1: add exactly 4 new register units.
ACTION_2: remove exactly 4 counted register units.
TARGET: S1. QUERY: What is the value of register R at the target state?
OUTPUT CONTRACT: Return exactly one JSON object with key "value". Its value is a nonnegative JSON integer, NOT a quoted string; use the JSON literal null, not the string "null", when undetermined. Even an undetermined answer must use the object {"value":null}, not bare null.
\end{scprompt}

\subsubsection{T2\_S1}
\label{app:sc_prompt_w1_t2_s1}

Recorded expected value: 13.

\begin{scprompt}
NON-SPATIAL SYMBOLIC REGISTER R. Its initial value in S0 is 9. Actions are cumulative; A2 acts on the result of A1.
ACTION_1: add exactly 4 new register units.
ACTION_2: remove exactly 4 counted register units.
QUERY: What is the value of register R immediately after Action 1 but before Action 2?
OUTPUT CONTRACT: Return exactly one JSON object with key "value". Its value is a nonnegative JSON integer, NOT a quoted string; use the JSON literal null, not the string "null", when undetermined. Even an undetermined answer must use the object {"value":null}, not bare null.
\end{scprompt}

\subsubsection{T3\_S1}
\label{app:sc_prompt_w1_t3_s1}

Recorded expected value: 13.

\begin{scprompt}
NON-SPATIAL SYMBOLIC REGISTER R. Its initial value in S0 is 9. Actions are cumulative; A2 acts on the result of A1.
ACTION_1: add exactly 4 new register units.
ACTION_2: remove exactly 4 counted register units.
STATE DEFINITIONS: S0 is before both actions. S1 is immediately after Action 1 and before Action 2. S2 is immediately after Action 2, applied to S1.
TARGET: S1. QUERY: What is the value of register R at the target state?
OUTPUT CONTRACT: Return exactly one JSON object with key "value". Its value is a nonnegative JSON integer, NOT a quoted string; use the JSON literal null, not the string "null", when undetermined. Even an undetermined answer must use the object {"value":null}, not bare null.
\end{scprompt}

\subsubsection{T4\_S1}
\label{app:sc_prompt_w1_t4_s1}

Recorded expected value: 13.

\begin{scprompt}
NON-SPATIAL SYMBOLIC REGISTER R. Its initial value in S0 is 9. Actions are cumulative; A2 acts on the result of A1.
ACTION_1: add exactly 4 new register units.
ACTION_2: remove exactly 4 counted register units.
STATE DEFINITIONS: S0 is before both actions. S1 is immediately after Action 1 and before Action 2. S2 is immediately after Action 2, applied to S1.
TARGET: S1. QUERY: What is the value of register R at the target state?
Do not return S2. Return S1 only.
OUTPUT CONTRACT: Return exactly one JSON object with key "value". Its value is a nonnegative JSON integer, NOT a quoted string; use the JSON literal null, not the string "null", when undetermined. Even an undetermined answer must use the object {"value":null}, not bare null.
\end{scprompt}

\subsubsection{T1\_S2}
\label{app:sc_prompt_w1_t1_s2}

Recorded expected value: 9.

\begin{scprompt}
NON-SPATIAL SYMBOLIC REGISTER R. Its initial value in S0 is 9. Actions are cumulative; A2 acts on the result of A1.
ACTION_1: add exactly 4 new register units.
ACTION_2: remove exactly 4 counted register units.
TARGET: S2. QUERY: What is the value of register R at the target state?
OUTPUT CONTRACT: Return exactly one JSON object with key "value". Its value is a nonnegative JSON integer, NOT a quoted string; use the JSON literal null, not the string "null", when undetermined. Even an undetermined answer must use the object {"value":null}, not bare null.
\end{scprompt}

\subsubsection{T2\_S2}
\label{app:sc_prompt_w1_t2_s2}

Recorded expected value: 9.

\begin{scprompt}
NON-SPATIAL SYMBOLIC REGISTER R. Its initial value in S0 is 9. Actions are cumulative; A2 acts on the result of A1.
ACTION_1: add exactly 4 new register units.
ACTION_2: remove exactly 4 counted register units.
QUERY: What is the value of register R immediately after Action 2 has acted on the result of Action 1?
OUTPUT CONTRACT: Return exactly one JSON object with key "value". Its value is a nonnegative JSON integer, NOT a quoted string; use the JSON literal null, not the string "null", when undetermined. Even an undetermined answer must use the object {"value":null}, not bare null.
\end{scprompt}

\subsubsection{T3\_S2}
\label{app:sc_prompt_w1_t3_s2}

Recorded expected value: 9.

\begin{scprompt}
NON-SPATIAL SYMBOLIC REGISTER R. Its initial value in S0 is 9. Actions are cumulative; A2 acts on the result of A1.
ACTION_1: add exactly 4 new register units.
ACTION_2: remove exactly 4 counted register units.
STATE DEFINITIONS: S0 is before both actions. S1 is immediately after Action 1 and before Action 2. S2 is immediately after Action 2, applied to S1.
TARGET: S2. QUERY: What is the value of register R at the target state?
OUTPUT CONTRACT: Return exactly one JSON object with key "value". Its value is a nonnegative JSON integer, NOT a quoted string; use the JSON literal null, not the string "null", when undetermined. Even an undetermined answer must use the object {"value":null}, not bare null.
\end{scprompt}

\subsubsection{T4\_S2}
\label{app:sc_prompt_w1_t4_s2}

Recorded expected value: 9.

\begin{scprompt}
NON-SPATIAL SYMBOLIC REGISTER R. Its initial value in S0 is 9. Actions are cumulative; A2 acts on the result of A1.
ACTION_1: add exactly 4 new register units.
ACTION_2: remove exactly 4 counted register units.
STATE DEFINITIONS: S0 is before both actions. S1 is immediately after Action 1 and before Action 2. S2 is immediately after Action 2, applied to S1.
TARGET: S2. QUERY: What is the value of register R at the target state?
Do not return S1. Return S2 only.
OUTPUT CONTRACT: Return exactly one JSON object with key "value". Its value is a nonnegative JSON integer, NOT a quoted string; use the JSON literal null, not the string "null", when undetermined. Even an undetermined answer must use the object {"value":null}, not bare null.
\end{scprompt}

\subsection{Non-count relations: all conditions in three families}
\label{app:sc_prompts_noncount}

Within each family, the reported instance is selected by the fixed world-hash
rule rather than by its prediction. Each group gives all four conditions.
Horizontal and vertical examples use one image; the depth example uses a
reference image and four support views. The relation code in the explicit
prompt and the sham is the same. The sham negates its relevance to the
queried objects; the two sentences are not claimed to be token-identical or
equal in length. The coordinate transformation is a reflection followed by
identity, not a rendered physical edit.

\subsubsection{HORIZONTAL DIRECT S0}
\label{app:sc_prompt_nc_horizontal_direct_s0}

Media blocks: 1. Recorded expected relation: \texttt{LEFT\_OF}.

\begin{scprompt}
SCOPE: current_media. REFERENCE FRAME: source_question_declared_perspective. TARGET VIEW: primary_view. Source bounding-box coordinates, when written in the claim, use the normalized 0–1000 xyxy scale. Colored outlines identify only the source-annotated targets in each supplied view. frame_0 is the primary view; other frame numbers retain their original source indices.
All images show S0, the original observed scene.
IMAGE ORDER: 1=frame_0
TARGET: S0
Use the named state and reference frame. QUERY: Along the horizontal (left/right) axis, what is the relation of the red bbox target to the blue bbox target?
OUTPUT CONTRACT: Return one JSON object with only "value": one of ["LEFT_OF", "RIGHT_OF"] or null if not determined.
Maximum 512 output tokens.
\end{scprompt}

\subsubsection{HORIZONTAL FULL TRANSITION}
\label{app:sc_prompt_nc_horizontal_full_transition}

Media blocks: 1. Recorded expected relation: \texttt{RIGHT\_OF}.

\begin{scprompt}
SCOPE: current_media. REFERENCE FRAME: source_question_declared_perspective. TARGET VIEW: primary_view. Source bounding-box coordinates, when written in the claim, use the normalized 0–1000 xyxy scale. Colored outlines identify only the source-annotated targets in each supplied view. frame_0 is the primary view; other frame numbers retain their original source indices.
All images show S0, the original observed scene.
IMAGE ORDER: 1=frame_0
S0 is the original frame. Action 1 produces S1. Action 2 acts on S1 to produce S2. These are coordinate-frame transformations, not new images or physical movements.
ACTION_1: Reverse only the horizontal coordinate axis of the named reference frame; preserve the objects and their identities.
ACTION_2: Leave the transformed frame and objects unchanged.
TARGET: S2
Use the named state and reference frame. QUERY: Along the horizontal (left/right) axis, what is the relation of the red bbox target to the blue bbox target?
OUTPUT CONTRACT: Return one JSON object with only "value": one of ["LEFT_OF", "RIGHT_OF"] or null if not determined.
Maximum 512 output tokens.
\end{scprompt}

\subsubsection{HORIZONTAL EXPLICIT S0}
\label{app:sc_prompt_nc_horizontal_explicit_s0}

Media blocks: 1. Recorded expected relation: \texttt{RIGHT\_OF}.

\begin{scprompt}
SCOPE: current_media. REFERENCE FRAME: source_question_declared_perspective. TARGET VIEW: primary_view. Source bounding-box coordinates, when written in the claim, use the normalized 0–1000 xyxy scale. Colored outlines identify only the source-annotated targets in each supplied view. frame_0 is the primary view; other frame numbers retain their original source indices.
All images show S0, the original observed scene.
IMAGE ORDER: 1=frame_0
ORACLE INITIAL FACT: In S0, the red bbox target has relation LEFT_OF to the blue bbox target.
S0 is the original frame. Action 1 produces S1. Action 2 acts on S1 to produce S2. These are coordinate-frame transformations, not new images or physical movements.
ACTION_1: Reverse only the horizontal coordinate axis of the named reference frame; preserve the objects and their identities.
ACTION_2: Leave the transformed frame and objects unchanged.
TARGET: S2
Use the named state and reference frame. QUERY: Along the horizontal (left/right) axis, what is the relation of the red bbox target to the blue bbox target?
OUTPUT CONTRACT: Return one JSON object with only "value": one of ["LEFT_OF", "RIGHT_OF"] or null if not determined.
Maximum 512 output tokens.
\end{scprompt}

\subsubsection{HORIZONTAL MATCHED SHAM}
\label{app:sc_prompt_nc_horizontal_matched_sham}

Media blocks: 1. Recorded expected relation: \texttt{RIGHT\_OF}.

\begin{scprompt}
SCOPE: current_media. REFERENCE FRAME: source_question_declared_perspective. TARGET VIEW: primary_view. Source bounding-box coordinates, when written in the claim, use the normalized 0–1000 xyxy scale. Colored outlines identify only the source-annotated targets in each supplied view. frame_0 is the primary view; other frame numbers retain their original source indices.
All images show S0, the original observed scene.
IMAGE ORDER: 1=frame_0
UNRELATED REGISTER: A separate bookkeeping entry stores the relation code LEFT_OF; it is not a fact about these objects.
S0 is the original frame. Action 1 produces S1. Action 2 acts on S1 to produce S2. These are coordinate-frame transformations, not new images or physical movements.
ACTION_1: Reverse only the horizontal coordinate axis of the named reference frame; preserve the objects and their identities.
ACTION_2: Leave the transformed frame and objects unchanged.
TARGET: S2
Use the named state and reference frame. QUERY: Along the horizontal (left/right) axis, what is the relation of the red bbox target to the blue bbox target?
OUTPUT CONTRACT: Return one JSON object with only "value": one of ["LEFT_OF", "RIGHT_OF"] or null if not determined.
Maximum 512 output tokens.
\end{scprompt}

\subsubsection{VERTICAL DIRECT S0}
\label{app:sc_prompt_nc_vertical_direct_s0}

Media blocks: 1. Recorded expected relation: \texttt{ABOVE}.

\begin{scprompt}
SCOPE: current_media. REFERENCE FRAME: source_question_declared_perspective. TARGET VIEW: primary_view. Source bounding-box coordinates, when written in the claim, use the normalized 0–1000 xyxy scale. Colored outlines identify only the source-annotated targets in each supplied view. frame_0 is the primary view; other frame numbers retain their original source indices.
All images show S0, the original observed scene.
IMAGE ORDER: 1=frame_0
TARGET: S0
Use the named state and reference frame. QUERY: Along the vertical (above/below) axis, what is the relation of the red bbox target to the blue bbox target?
OUTPUT CONTRACT: Return one JSON object with only "value": one of ["ABOVE", "BELOW"] or null if not determined.
Maximum 512 output tokens.
\end{scprompt}

\subsubsection{VERTICAL FULL TRANSITION}
\label{app:sc_prompt_nc_vertical_full_transition}

Media blocks: 1. Recorded expected relation: \texttt{BELOW}.

\begin{scprompt}
SCOPE: current_media. REFERENCE FRAME: source_question_declared_perspective. TARGET VIEW: primary_view. Source bounding-box coordinates, when written in the claim, use the normalized 0–1000 xyxy scale. Colored outlines identify only the source-annotated targets in each supplied view. frame_0 is the primary view; other frame numbers retain their original source indices.
All images show S0, the original observed scene.
IMAGE ORDER: 1=frame_0
S0 is the original frame. Action 1 produces S1. Action 2 acts on S1 to produce S2. These are coordinate-frame transformations, not new images or physical movements.
ACTION_1: Reverse only the vertical coordinate axis of the named reference frame; preserve the objects and their identities.
ACTION_2: Leave the transformed frame and objects unchanged.
TARGET: S2
Use the named state and reference frame. QUERY: Along the vertical (above/below) axis, what is the relation of the red bbox target to the blue bbox target?
OUTPUT CONTRACT: Return one JSON object with only "value": one of ["ABOVE", "BELOW"] or null if not determined.
Maximum 512 output tokens.
\end{scprompt}

\subsubsection{VERTICAL EXPLICIT S0}
\label{app:sc_prompt_nc_vertical_explicit_s0}

Media blocks: 1. Recorded expected relation: \texttt{BELOW}.

\begin{scprompt}
SCOPE: current_media. REFERENCE FRAME: source_question_declared_perspective. TARGET VIEW: primary_view. Source bounding-box coordinates, when written in the claim, use the normalized 0–1000 xyxy scale. Colored outlines identify only the source-annotated targets in each supplied view. frame_0 is the primary view; other frame numbers retain their original source indices.
All images show S0, the original observed scene.
IMAGE ORDER: 1=frame_0
ORACLE INITIAL FACT: In S0, the red bbox target has relation ABOVE to the blue bbox target.
S0 is the original frame. Action 1 produces S1. Action 2 acts on S1 to produce S2. These are coordinate-frame transformations, not new images or physical movements.
ACTION_1: Reverse only the vertical coordinate axis of the named reference frame; preserve the objects and their identities.
ACTION_2: Leave the transformed frame and objects unchanged.
TARGET: S2
Use the named state and reference frame. QUERY: Along the vertical (above/below) axis, what is the relation of the red bbox target to the blue bbox target?
OUTPUT CONTRACT: Return one JSON object with only "value": one of ["ABOVE", "BELOW"] or null if not determined.
Maximum 512 output tokens.
\end{scprompt}

\subsubsection{VERTICAL MATCHED SHAM}
\label{app:sc_prompt_nc_vertical_matched_sham}

Media blocks: 1. Recorded expected relation: \texttt{BELOW}.

\begin{scprompt}
SCOPE: current_media. REFERENCE FRAME: source_question_declared_perspective. TARGET VIEW: primary_view. Source bounding-box coordinates, when written in the claim, use the normalized 0–1000 xyxy scale. Colored outlines identify only the source-annotated targets in each supplied view. frame_0 is the primary view; other frame numbers retain their original source indices.
All images show S0, the original observed scene.
IMAGE ORDER: 1=frame_0
UNRELATED REGISTER: A separate bookkeeping entry stores the relation code ABOVE; it is not a fact about these objects.
S0 is the original frame. Action 1 produces S1. Action 2 acts on S1 to produce S2. These are coordinate-frame transformations, not new images or physical movements.
ACTION_1: Reverse only the vertical coordinate axis of the named reference frame; preserve the objects and their identities.
ACTION_2: Leave the transformed frame and objects unchanged.
TARGET: S2
Use the named state and reference frame. QUERY: Along the vertical (above/below) axis, what is the relation of the red bbox target to the blue bbox target?
OUTPUT CONTRACT: Return one JSON object with only "value": one of ["ABOVE", "BELOW"] or null if not determined.
Maximum 512 output tokens.
\end{scprompt}

\subsubsection{DEPTH DIRECT S0}
\label{app:sc_prompt_nc_depth_direct_s0}

Media blocks: 5. Recorded expected relation: \texttt{BEHIND}.

\begin{scprompt}
SCOPE: reference_frame. REFERENCE FRAME: ca_vqa_reference_frame. TARGET VIEW: reference_frame:train_bundle_29289db3edf69519410d8855:record_214_frame_4. 
All images show S0, the original observed scene.
IMAGE ORDER: 1=reference_frame, 2=support_frame_1, 3=support_frame_2, 4=support_frame_3, 5=support_frame_4
TARGET: S0
Use the named state and reference frame. QUERY: Along the depth (front/behind) axis, what is the relation of the flower vase to the table?
OUTPUT CONTRACT: Return one JSON object with only "value": one of ["BEHIND", "FRONT_OF"] or null if not determined.
Maximum 512 output tokens.
\end{scprompt}

\subsubsection{DEPTH FULL TRANSITION}
\label{app:sc_prompt_nc_depth_full_transition}

Media blocks: 5. Recorded expected relation: \texttt{FRONT\_OF}.

\begin{scprompt}
SCOPE: reference_frame. REFERENCE FRAME: ca_vqa_reference_frame. TARGET VIEW: reference_frame:train_bundle_29289db3edf69519410d8855:record_214_frame_4. 
All images show S0, the original observed scene.
IMAGE ORDER: 1=reference_frame, 2=support_frame_1, 3=support_frame_2, 4=support_frame_3, 5=support_frame_4
S0 is the original frame. Action 1 produces S1. Action 2 acts on S1 to produce S2. These are coordinate-frame transformations, not new images or physical movements.
ACTION_1: Reverse only the depth coordinate axis of the named reference frame; preserve the objects and their identities.
ACTION_2: Leave the transformed frame and objects unchanged.
TARGET: S2
Use the named state and reference frame. QUERY: Along the depth (front/behind) axis, what is the relation of the flower vase to the table?
OUTPUT CONTRACT: Return one JSON object with only "value": one of ["BEHIND", "FRONT_OF"] or null if not determined.
Maximum 512 output tokens.
\end{scprompt}

\subsubsection{DEPTH EXPLICIT S0}
\label{app:sc_prompt_nc_depth_explicit_s0}

Media blocks: 5. Recorded expected relation: \texttt{FRONT\_OF}.

\begin{scprompt}
SCOPE: reference_frame. REFERENCE FRAME: ca_vqa_reference_frame. TARGET VIEW: reference_frame:train_bundle_29289db3edf69519410d8855:record_214_frame_4. 
All images show S0, the original observed scene.
IMAGE ORDER: 1=reference_frame, 2=support_frame_1, 3=support_frame_2, 4=support_frame_3, 5=support_frame_4
ORACLE INITIAL FACT: In S0, the flower vase has relation BEHIND to the table.
S0 is the original frame. Action 1 produces S1. Action 2 acts on S1 to produce S2. These are coordinate-frame transformations, not new images or physical movements.
ACTION_1: Reverse only the depth coordinate axis of the named reference frame; preserve the objects and their identities.
ACTION_2: Leave the transformed frame and objects unchanged.
TARGET: S2
Use the named state and reference frame. QUERY: Along the depth (front/behind) axis, what is the relation of the flower vase to the table?
OUTPUT CONTRACT: Return one JSON object with only "value": one of ["BEHIND", "FRONT_OF"] or null if not determined.
Maximum 512 output tokens.
\end{scprompt}

\subsubsection{DEPTH MATCHED SHAM}
\label{app:sc_prompt_nc_depth_matched_sham}

Media blocks: 5. Recorded expected relation: \texttt{FRONT\_OF}.

\begin{scprompt}
SCOPE: reference_frame. REFERENCE FRAME: ca_vqa_reference_frame. TARGET VIEW: reference_frame:train_bundle_29289db3edf69519410d8855:record_214_frame_4. 
All images show S0, the original observed scene.
IMAGE ORDER: 1=reference_frame, 2=support_frame_1, 3=support_frame_2, 4=support_frame_3, 5=support_frame_4
UNRELATED REGISTER: A separate bookkeeping entry stores the relation code BEHIND; it is not a fact about these objects.
S0 is the original frame. Action 1 produces S1. Action 2 acts on S1 to produce S2. These are coordinate-frame transformations, not new images or physical movements.
ACTION_1: Reverse only the depth coordinate axis of the named reference frame; preserve the objects and their identities.
ACTION_2: Leave the transformed frame and objects unchanged.
TARGET: S2
Use the named state and reference frame. QUERY: Along the depth (front/behind) axis, what is the relation of the flower vase to the table?
OUTPUT CONTRACT: Return one JSON object with only "value": one of ["BEHIND", "FRONT_OF"] or null if not determined.
Maximum 512 output tokens.
\end{scprompt}

\subsection{I2 symbolic recipient and donor}
\label{app:sc_prompts_i2}

The recipient starts at 10, sets the register to 2, and then adds 3, giving
original final value 5. The donor starts at 10, adds 2, and then removes 1,
giving intermediate value 12 and final value 11. Under
Eq.~\eqref{eq:sc_i2_target}, replacing the recipient intermediate value with
the donor intermediate value predicts 15. These gold values define the
interchange target outside the actual input; they are not appended to either
prompt.

\subsubsection{I2 Recipient}
\label{app:sc_prompt_i2_recipient}

No media. Shared diagnostic system instruction.

\begin{scprompt}
NON-SPATIAL SYMBOLIC REGISTER R. Its initial value in S0 is 10. Actions are cumulative; A2 acts on the result of A1.
ACTION_1: set the register value to exactly 2, replacing its previous value.
ACTION_2: add exactly 3 new register units.
TARGET: S2. QUERY: What is the value of register R at this state?
OUTPUT CONTRACT: Return exactly one JSON object with key "value". Its value is a nonnegative JSON integer, NOT a quoted string; use the JSON literal null, not the string "null", when undetermined. Even an undetermined answer must use the object {"value":null}, not bare null.
\end{scprompt}

\subsubsection{I2 Donor}
\label{app:sc_prompt_i2_donor}

No media. Shared diagnostic system instruction.

\begin{scprompt}
NON-SPATIAL SYMBOLIC REGISTER R. Its initial value in S0 is 10. Actions are cumulative; A2 acts on the result of A1.
ACTION_1: add exactly 2 new register units.
ACTION_2: remove exactly 1 counted register units.
TARGET: S2. QUERY: What is the value of register R at this state?
OUTPUT CONTRACT: Return exactly one JSON object with key "value". Its value is a nonnegative JSON integer, NOT a quoted string; use the JSON literal null, not the string "null", when undetermined. Even an undetermined answer must use the object {"value":null}, not bare null.
\end{scprompt}

\subsection{I1 actual recipient, donor, and protection inputs}
\label{app:sc_prompts_i1}

The following instances expose the actual same-world initial-state donor
construction for final-state, initial-state, and protected-fact questions.
The painting count is the transformed target, while lamps are protected.
The matched register carries the initial painting count before action 1.
An independent query-specific baseline and donor are used for protection;
we do not reuse an S2 baseline as the answer to a different question.
The final instance is the recorded structurally matched success donor from
another world. All examples use five media blocks in the reference/support
order and the shared diagnostic system instruction.

\subsubsection{PROTECTED BASE}
\label{app:sc_prompt_i1_1}

Actual condition: \texttt{BASE\_PROTECTED}. Recorded expected value: 2. Activation directions and eligible controls are described in Appendix~\ref{app:sc_i1}.

\begin{scprompt}
OBSERVED STATE S0: the supplied reference image. Support views do not add members to the counting scope.
IMAGE ORDER: 1=reference_frame, 2=support_frame_1, 3=support_frame_2, 4=support_frame_3, 5=support_frame_4
OBJECT: paintings. COUNTING SCOPE: the paintings counted in the reference image, not a whole-room census.
The following two actions are cumulative on one hypothetical counted set. A1 maps S0 to S1; A2 acts on S1 to produce S2, not independently on S0. The recorded S0 and all other categories remain unchanged. Only quantities are queried; do not recount occlusion after the edit.
ACTION_1: remove exactly 1 counted paintings.
STATE_NOTE: No state value is supplied here.
ACTION_2: add exactly 1 new paintings.
TARGET: S2
QUERY: How many lamps are in the reference-image counting scope in state S2?
OUTPUT CONTRACT: Return exactly one JSON object with key "value". Its value is a nonnegative JSON integer, NOT a quoted string; use the JSON literal null, not the string "null", when undetermined. Even an undetermined answer must use the object {"value":null}, not bare null.
\end{scprompt}

\subsubsection{PROTECTED INFORMATIVE S0}
\label{app:sc_prompt_i1_2}

Actual condition: \texttt{INFORMATIVE\_S0\_PROTECTED}. Recorded expected value: 2. Activation directions and eligible controls are described in Appendix~\ref{app:sc_i1}.

\begin{scprompt}
OBSERVED STATE S0: the supplied reference image. Support views do not add members to the counting scope.
IMAGE ORDER: 1=reference_frame, 2=support_frame_1, 3=support_frame_2, 4=support_frame_3, 5=support_frame_4
OBJECT: paintings. COUNTING SCOPE: the paintings counted in the reference image, not a whole-room census.
ORACLE INITIAL FACT: the target count in S0 is exactly 1.
The following two actions are cumulative on one hypothetical counted set. A1 maps S0 to S1; A2 acts on S1 to produce S2, not independently on S0. The recorded S0 and all other categories remain unchanged. Only quantities are queried; do not recount occlusion after the edit.
ACTION_1: remove exactly 1 counted paintings.
STATE_NOTE: No state value is supplied here.
ACTION_2: add exactly 1 new paintings.
TARGET: S2
QUERY: How many lamps are in the reference-image counting scope in state S2?
OUTPUT CONTRACT: Return exactly one JSON object with key "value". Its value is a nonnegative JSON integer, NOT a quoted string; use the JSON literal null, not the string "null", when undetermined. Even an undetermined answer must use the object {"value":null}, not bare null.
\end{scprompt}

\subsubsection{PROTECTED SAME VALUE SHAM}
\label{app:sc_prompt_i1_3}

Actual condition: \texttt{SAME\_VALUE\_SHAM\_PROTECTED}. Recorded expected value: 2. Activation directions and eligible controls are described in Appendix~\ref{app:sc_i1}.

\begin{scprompt}
OBSERVED STATE S0: the supplied reference image. Support views do not add members to the counting scope.
IMAGE ORDER: 1=reference_frame, 2=support_frame_1, 3=support_frame_2, 4=support_frame_3, 5=support_frame_4
OBJECT: paintings. COUNTING SCOPE: the paintings counted in the reference image, not a whole-room census.
UNRELATED REGISTER: its bookkeeping value is exactly 1, not the scene state.
The following two actions are cumulative on one hypothetical counted set. A1 maps S0 to S1; A2 acts on S1 to produce S2, not independently on S0. The recorded S0 and all other categories remain unchanged. Only quantities are queried; do not recount occlusion after the edit.
ACTION_1: remove exactly 1 counted paintings.
STATE_NOTE: No state value is supplied here.
ACTION_2: add exactly 1 new paintings.
TARGET: S2
QUERY: How many lamps are in the reference-image counting scope in state S2?
OUTPUT CONTRACT: Return exactly one JSON object with key "value". Its value is a nonnegative JSON integer, NOT a quoted string; use the JSON literal null, not the string "null", when undetermined. Even an undetermined answer must use the object {"value":null}, not bare null.
\end{scprompt}

\subsubsection{S0 BASE}
\label{app:sc_prompt_i1_4}

Actual condition: \texttt{BASE\_S0}. Recorded expected value: 1. Activation directions and eligible controls are described in Appendix~\ref{app:sc_i1}.

\begin{scprompt}
OBSERVED STATE S0: the supplied reference image. Support views do not add members to the counting scope.
IMAGE ORDER: 1=reference_frame, 2=support_frame_1, 3=support_frame_2, 4=support_frame_3, 5=support_frame_4
OBJECT: paintings. COUNTING SCOPE: the paintings counted in the reference image, not a whole-room census.
The following two actions are cumulative on one hypothetical counted set. A1 maps S0 to S1; A2 acts on S1 to produce S2, not independently on S0. The recorded S0 and all other categories remain unchanged. Only quantities are queried; do not recount occlusion after the edit.
ACTION_1: remove exactly 1 counted paintings.
STATE_NOTE: No state value is supplied here.
ACTION_2: add exactly 1 new paintings.
TARGET: S0
QUERY: How many paintings are in the reference-image counting scope in state S0?
OUTPUT CONTRACT: Return exactly one JSON object with key "value". Its value is a nonnegative JSON integer, NOT a quoted string; use the JSON literal null, not the string "null", when undetermined. Even an undetermined answer must use the object {"value":null}, not bare null.
\end{scprompt}

\subsubsection{S0 INFORMATIVE S0}
\label{app:sc_prompt_i1_5}

Actual condition: \texttt{INFORMATIVE\_S0\_S0}. Recorded expected value: 1. Activation directions and eligible controls are described in Appendix~\ref{app:sc_i1}.

\begin{scprompt}
OBSERVED STATE S0: the supplied reference image. Support views do not add members to the counting scope.
IMAGE ORDER: 1=reference_frame, 2=support_frame_1, 3=support_frame_2, 4=support_frame_3, 5=support_frame_4
OBJECT: paintings. COUNTING SCOPE: the paintings counted in the reference image, not a whole-room census.
ORACLE INITIAL FACT: the target count in S0 is exactly 1.
The following two actions are cumulative on one hypothetical counted set. A1 maps S0 to S1; A2 acts on S1 to produce S2, not independently on S0. The recorded S0 and all other categories remain unchanged. Only quantities are queried; do not recount occlusion after the edit.
ACTION_1: remove exactly 1 counted paintings.
STATE_NOTE: No state value is supplied here.
ACTION_2: add exactly 1 new paintings.
TARGET: S0
QUERY: How many paintings are in the reference-image counting scope in state S0?
OUTPUT CONTRACT: Return exactly one JSON object with key "value". Its value is a nonnegative JSON integer, NOT a quoted string; use the JSON literal null, not the string "null", when undetermined. Even an undetermined answer must use the object {"value":null}, not bare null.
\end{scprompt}

\subsubsection{S0 SAME VALUE SHAM}
\label{app:sc_prompt_i1_6}

Actual condition: \texttt{SAME\_VALUE\_SHAM\_S0}. Recorded expected value: 1. Activation directions and eligible controls are described in Appendix~\ref{app:sc_i1}.

\begin{scprompt}
OBSERVED STATE S0: the supplied reference image. Support views do not add members to the counting scope.
IMAGE ORDER: 1=reference_frame, 2=support_frame_1, 3=support_frame_2, 4=support_frame_3, 5=support_frame_4
OBJECT: paintings. COUNTING SCOPE: the paintings counted in the reference image, not a whole-room census.
UNRELATED REGISTER: its bookkeeping value is exactly 1, not the scene state.
The following two actions are cumulative on one hypothetical counted set. A1 maps S0 to S1; A2 acts on S1 to produce S2, not independently on S0. The recorded S0 and all other categories remain unchanged. Only quantities are queried; do not recount occlusion after the edit.
ACTION_1: remove exactly 1 counted paintings.
STATE_NOTE: No state value is supplied here.
ACTION_2: add exactly 1 new paintings.
TARGET: S0
QUERY: How many paintings are in the reference-image counting scope in state S0?
OUTPUT CONTRACT: Return exactly one JSON object with key "value". Its value is a nonnegative JSON integer, NOT a quoted string; use the JSON literal null, not the string "null", when undetermined. Even an undetermined answer must use the object {"value":null}, not bare null.
\end{scprompt}

\subsubsection{S2 BASE}
\label{app:sc_prompt_i1_7}

Actual condition: \texttt{BASE\_S2}. Recorded expected value: 1. Activation directions and eligible controls are described in Appendix~\ref{app:sc_i1}.

\begin{scprompt}
OBSERVED STATE S0: the supplied reference image. Support views do not add members to the counting scope.
IMAGE ORDER: 1=reference_frame, 2=support_frame_1, 3=support_frame_2, 4=support_frame_3, 5=support_frame_4
OBJECT: paintings. COUNTING SCOPE: the paintings counted in the reference image, not a whole-room census.
The following two actions are cumulative on one hypothetical counted set. A1 maps S0 to S1; A2 acts on S1 to produce S2, not independently on S0. The recorded S0 and all other categories remain unchanged. Only quantities are queried; do not recount occlusion after the edit.
ACTION_1: remove exactly 1 counted paintings.
STATE_NOTE: No state value is supplied here.
ACTION_2: add exactly 1 new paintings.
TARGET: S2
QUERY: How many paintings are in the reference-image counting scope in state S2?
OUTPUT CONTRACT: Return exactly one JSON object with key "value". Its value is a nonnegative JSON integer, NOT a quoted string; use the JSON literal null, not the string "null", when undetermined. Even an undetermined answer must use the object {"value":null}, not bare null.
\end{scprompt}

\subsubsection{S2 INFORMATIVE S0}
\label{app:sc_prompt_i1_8}

Actual condition: \texttt{INFORMATIVE\_S0\_S2}. Recorded expected value: 1. Activation directions and eligible controls are described in Appendix~\ref{app:sc_i1}.

\begin{scprompt}
OBSERVED STATE S0: the supplied reference image. Support views do not add members to the counting scope.
IMAGE ORDER: 1=reference_frame, 2=support_frame_1, 3=support_frame_2, 4=support_frame_3, 5=support_frame_4
OBJECT: paintings. COUNTING SCOPE: the paintings counted in the reference image, not a whole-room census.
ORACLE INITIAL FACT: the target count in S0 is exactly 1.
The following two actions are cumulative on one hypothetical counted set. A1 maps S0 to S1; A2 acts on S1 to produce S2, not independently on S0. The recorded S0 and all other categories remain unchanged. Only quantities are queried; do not recount occlusion after the edit.
ACTION_1: remove exactly 1 counted paintings.
STATE_NOTE: No state value is supplied here.
ACTION_2: add exactly 1 new paintings.
TARGET: S2
QUERY: How many paintings are in the reference-image counting scope in state S2?
OUTPUT CONTRACT: Return exactly one JSON object with key "value". Its value is a nonnegative JSON integer, NOT a quoted string; use the JSON literal null, not the string "null", when undetermined. Even an undetermined answer must use the object {"value":null}, not bare null.
\end{scprompt}

\subsubsection{S2 SAME VALUE SHAM}
\label{app:sc_prompt_i1_9}

Actual condition: \texttt{SAME\_VALUE\_SHAM\_S2}. Recorded expected value: 1. Activation directions and eligible controls are described in Appendix~\ref{app:sc_i1}.

\begin{scprompt}
OBSERVED STATE S0: the supplied reference image. Support views do not add members to the counting scope.
IMAGE ORDER: 1=reference_frame, 2=support_frame_1, 3=support_frame_2, 4=support_frame_3, 5=support_frame_4
OBJECT: paintings. COUNTING SCOPE: the paintings counted in the reference image, not a whole-room census.
UNRELATED REGISTER: its bookkeeping value is exactly 1, not the scene state.
The following two actions are cumulative on one hypothetical counted set. A1 maps S0 to S1; A2 acts on S1 to produce S2, not independently on S0. The recorded S0 and all other categories remain unchanged. Only quantities are queried; do not recount occlusion after the edit.
ACTION_1: remove exactly 1 counted paintings.
STATE_NOTE: No state value is supplied here.
ACTION_2: add exactly 1 new paintings.
TARGET: S2
QUERY: How many paintings are in the reference-image counting scope in state S2?
OUTPUT CONTRACT: Return exactly one JSON object with key "value". Its value is a nonnegative JSON integer, NOT a quoted string; use the JSON literal null, not the string "null", when undetermined. Even an undetermined answer must use the object {"value":null}, not bare null.
\end{scprompt}

\subsubsection{Recorded success donor for S2}
\label{app:sc_prompt_i1_10}

Actual condition: \texttt{BASE\_S2}. Recorded expected value: 2. Activation directions and eligible controls are described in Appendix~\ref{app:sc_i1}.

\begin{scprompt}
OBSERVED STATE S0: the supplied reference image. Support views do not add members to the counting scope.
IMAGE ORDER: 1=reference_frame, 2=support_frame_1, 3=support_frame_2, 4=support_frame_3, 5=support_frame_4
OBJECT: doors. COUNTING SCOPE: the doors counted in the reference image, not a whole-room census.
The following two actions are cumulative on one hypothetical counted set. A1 maps S0 to S1; A2 acts on S1 to produce S2, not independently on S0. The recorded S0 and all other categories remain unchanged. Only quantities are queried; do not recount occlusion after the edit.
ACTION_1: add exactly 2 new doors.
STATE_NOTE: No state value is supplied here.
ACTION_2: remove exactly 2 counted doors.
TARGET: S2
QUERY: How many doors are in the reference-image counting scope in state S2?
OUTPUT CONTRACT: Return exactly one JSON object with key "value". Its value is a nonnegative JSON integer, NOT a quoted string; use the JSON literal null, not the string "null", when undetermined. Even an undetermined answer must use the object {"value":null}, not bare null.
\end{scprompt}

\FloatBarrier
\section{Complete fixed-layer probing ledger}
\label{app:sc_probe_ledger}

\ifSCFullProbeTables
This ledger retains all 1,365 source result records in a two-anchor layout.
A residual row has separate checkpoint and query cells; baseline rows contain
one result, not two replicated evaluations. Each cell gives correct/N and
its archived accuracy confidence interval in percent. ``Sel.'' is the
original selection accuracy (percent), not a score used to select a new
reporting layer. $U$ is the number of evaluation rows with a class absent
from the training class set. A nonzero $U$ marks the original
class-coverage-limited estimate. Dashes identify unavailable measurements,
not zero accuracy. The value fractions preserve the underlying event counts;
intervals and selection percentages are rounded only for presentation.

For visual B1 panels, the train/selection/evaluation split is 48/16/16 worlds
and evaluation contains 32 sequences. For symbolic B2, the split is 10/3/3
program families and evaluation contains 18 sequences from three families.
The combined surface baseline is the actual word/number-bag-plus-length
reader. The archived evidence for this appendix does not specify a numerical ridge
regularization grid; we do not assert unrecorded hyperparameter values.

\subsection{Qwen3.5-4B: B1/B01}
\label{app:sc_r1_full_4b_b1_b01}

Table~\ref{tab:sc_r1_full_4b_b1_b01} includes initial and final state variables, both actions, entity identity, and the protected fact for the original full visual context. The complete table prevents the prespecified S1 summary from being read as a claim that every variable is equally unreadable. Class-coverage limitations and the constant target label are retained.

% Source: supplied evidence report, lines 4295--5661.
\FloatBarrier
\begingroup
\scriptsize
\setlength{\tabcolsep}{3pt}
\renewcommand{\arraystretch}{1.08}
\begin{longtable}{L{0.105\linewidth}L{0.09\linewidth}L{0.25\linewidth}rL{0.25\linewidth}rr}
\caption{Qwen3.5-4B B1/B01 complete fixed-layer ledger. Residual rows are indexed by actual block; baseline rows have no block. Correct/N and confidence intervals are retained for each available anchor; selection values are descriptive. Evaluation: 32 sequences/16 worlds.}\label{tab:sc_r1_full_4b_b1_b01}\\
\toprule
Variable & Reader/block & Checkpoint: count [CI] & Sel. & Query: count [CI] & Sel. & $U$ \\
\midrule
\endfirsthead
\multicolumn{7}{l}{\footnotesize Table~\thetable\ continued.}\\
\toprule
Variable & Reader/block & Checkpoint: count [CI] & Sel. & Query: count [CI] & Sel. & $U$ \\
\midrule
\endhead
\midrule\multicolumn{7}{r}{\footnotesize Continued on the next page.}\\
\endfoot
\bottomrule
\endlastfoot
$Z$ & \multicolumn{6}{l}{Constant training label} \\
$s_0$ & Majority & 22/32 [43.75, 87.50] & 62.50 & \scNA & \scNA & 2 \\
$s_0$ & Surface & 21/32 [43.75, 87.50] & 40.62 & \scNA & \scNA & 2 \\
$s_0$ & 0 & 16/32 [25.00, 75.00] & 37.50 & 19/32 [37.50, 81.25] & 43.75 & 2 \\
$s_0$ & 4 & 18/32 [31.25, 81.25] & 37.50 & 15/32 [25.00, 68.75] & 43.75 & 2 \\
$s_0$ & 9 & 22/32 [43.75, 87.50] & 43.75 & 19/32 [37.50, 81.25] & 43.75 & 2 \\
$s_0$ & 13 & 22/32 [43.75, 87.50] & 43.75 & 20/32 [40.62, 84.38] & 43.75 & 2 \\
$s_0$ & 18 & 24/32 [56.25, 93.75] & 43.75 & 22/32 [43.75, 87.50] & 28.12 & 2 \\
$s_0$ & 22 & 24/32 [50.00, 93.75] & 50.00 & 20/32 [37.50, 87.50] & 31.25 & 2 \\
$s_0$ & 27 & 24/32 [50.00, 93.75] & 37.50 & 20/32 [37.50, 87.50] & 31.25 & 2 \\
$s_0$ & 31 & 24/32 [50.00, 93.75] & 43.75 & 20/32 [37.50, 87.50] & 34.38 & 2 \\
$s_1$ & Majority & 10/32 [12.50, 56.25] & 43.75 & \scNA & \scNA & 0 \\
$s_1$ & Surface & 22/32 [46.88, 87.50] & 65.62 & \scNA & \scNA & 0 \\
$s_1$ & 0 & 12/32 [12.50, 62.50] & 31.25 & 7/32 [6.25, 40.70] & 21.88 & 0 \\
$s_1$ & 4 & 18/32 [31.25, 81.25] & 31.25 & 21/32 [43.75, 87.50] & 28.12 & 0 \\
$s_1$ & 9 & 22/32 [43.75, 87.50] & 37.50 & 18/32 [31.25, 81.25] & 31.25 & 0 \\
$s_1$ & 13 & 20/32 [37.50, 87.50] & 50.00 & 21/32 [43.75, 87.50] & 43.75 & 0 \\
$s_1$ & 18 & 20/32 [37.50, 87.50] & 37.50 & 21/32 [43.75, 87.50] & 43.75 & 0 \\
$s_1$ & 22 & 22/32 [43.75, 87.50] & 31.25 & 20/32 [37.50, 87.50] & 40.62 & 0 \\
$s_1$ & 27 & 22/32 [43.75, 87.50] & 37.50 & 20/32 [37.50, 87.50] & 40.62 & 0 \\
$s_1$ & 31 & 22/32 [43.75, 87.50] & 37.50 & 22/32 [43.75, 87.50] & 37.50 & 0 \\
$s_2$ & Majority & 11/32 [21.88, 43.75] & 31.25 & \scNA & \scNA & 0 \\
$s_2$ & Surface & 23/32 [50.00, 90.62] & 56.25 & \scNA & \scNA & 0 \\
$s_2$ & 0 & 9/32 [15.62, 40.62] & 21.88 & 10/32 [18.75, 46.88] & 40.62 & 0 \\
$s_2$ & 4 & 9/32 [15.62, 40.62] & 21.88 & 18/32 [37.50, 75.00] & 50.00 & 0 \\
$s_2$ & 9 & 11/32 [21.88, 43.75] & 21.88 & 21/32 [43.75, 84.38] & 46.88 & 0 \\
$s_2$ & 13 & 10/32 [18.75, 43.75] & 28.12 & 22/32 [46.88, 87.50] & 46.88 & 0 \\
$s_2$ & 18 & 12/32 [28.12, 46.88] & 25.00 & 23/32 [53.12, 87.50] & 43.75 & 0 \\
$s_2$ & 22 & 14/32 [34.38, 50.00] & 25.00 & 24/32 [56.25, 90.62] & 56.25 & 0 \\
$s_2$ & 27 & 13/32 [31.25, 50.00] & 21.88 & 19/32 [40.62, 78.12] & 50.00 & 0 \\
$s_2$ & 31 & 13/32 [31.25, 50.00] & 25.00 & 17/32 [34.38, 71.88] & 46.88 & 0 \\
A1 amt. & Majority & 16/32 [25.00, 75.00] & 50.00 & \scNA & \scNA & 0 \\
A1 amt. & Surface & 32/32 [100.00, 100.00] & 100.00 & \scNA & \scNA & 0 \\
A1 amt. & 0 & 32/32 [100.00, 100.00] & 93.75 & 24/32 [50.00, 93.75] & 75.00 & 0 \\
A1 amt. & 4 & 32/32 [100.00, 100.00] & 100.00 & 32/32 [100.00, 100.00] & 87.50 & 0 \\
A1 amt. & 9 & 32/32 [100.00, 100.00] & 100.00 & 32/32 [100.00, 100.00] & 100.00 & 0 \\
A1 amt. & 13 & 32/32 [100.00, 100.00] & 100.00 & 32/32 [100.00, 100.00] & 100.00 & 0 \\
A1 amt. & 18 & 32/32 [100.00, 100.00] & 100.00 & 32/32 [100.00, 100.00] & 100.00 & 0 \\
A1 amt. & 22 & 32/32 [100.00, 100.00] & 100.00 & 32/32 [100.00, 100.00] & 100.00 & 0 \\
A1 amt. & 27 & 32/32 [100.00, 100.00] & 100.00 & 32/32 [100.00, 100.00] & 100.00 & 0 \\
A1 amt. & 31 & 32/32 [100.00, 100.00] & 100.00 & 32/32 [100.00, 100.00] & 100.00 & 0 \\
A2 amt. & Majority & 16/32 [50.00, 50.00] & 50.00 & \scNA & \scNA & 0 \\
A2 amt. & Surface & 32/32 [100.00, 100.00] & 100.00 & \scNA & \scNA & 0 \\
A2 amt. & 0 & 16/32 [50.00, 50.00] & 50.00 & 23/32 [59.38, 84.38] & 68.75 & 0 \\
A2 amt. & 4 & 16/32 [50.00, 50.00] & 50.00 & 32/32 [100.00, 100.00] & 100.00 & 0 \\
A2 amt. & 9 & 16/32 [50.00, 50.00] & 50.00 & 32/32 [100.00, 100.00] & 100.00 & 0 \\
A2 amt. & 13 & 16/32 [50.00, 50.00] & 50.00 & 32/32 [100.00, 100.00] & 100.00 & 0 \\
A2 amt. & 18 & 16/32 [50.00, 50.00] & 50.00 & 32/32 [100.00, 100.00] & 100.00 & 0 \\
A2 amt. & 22 & 16/32 [50.00, 50.00] & 50.00 & 32/32 [100.00, 100.00] & 100.00 & 0 \\
A2 amt. & 27 & 16/32 [50.00, 50.00] & 50.00 & 31/32 [90.62, 100.00] & 100.00 & 0 \\
A2 amt. & 31 & 16/32 [50.00, 50.00] & 50.00 & 30/32 [84.38, 100.00] & 100.00 & 0 \\
A1 type & Majority & 16/32 [25.00, 75.00] & 50.00 & \scNA & \scNA & 0 \\
A1 type & Surface & 32/32 [100.00, 100.00] & 100.00 & \scNA & \scNA & 0 \\
A1 type & 0 & 32/32 [100.00, 100.00] & 93.75 & 24/32 [50.00, 93.75] & 75.00 & 0 \\
A1 type & 4 & 32/32 [100.00, 100.00] & 100.00 & 32/32 [100.00, 100.00] & 87.50 & 0 \\
A1 type & 9 & 32/32 [100.00, 100.00] & 100.00 & 32/32 [100.00, 100.00] & 100.00 & 0 \\
A1 type & 13 & 32/32 [100.00, 100.00] & 100.00 & 32/32 [100.00, 100.00] & 100.00 & 0 \\
A1 type & 18 & 32/32 [100.00, 100.00] & 100.00 & 32/32 [100.00, 100.00] & 100.00 & 0 \\
A1 type & 22 & 32/32 [100.00, 100.00] & 100.00 & 32/32 [100.00, 100.00] & 100.00 & 0 \\
A1 type & 27 & 32/32 [100.00, 100.00] & 100.00 & 32/32 [100.00, 100.00] & 100.00 & 0 \\
A1 type & 31 & 32/32 [100.00, 100.00] & 100.00 & 32/32 [100.00, 100.00] & 100.00 & 0 \\
A2 type & Majority & 24/32 [62.50, 87.50] & 75.00 & \scNA & \scNA & 0 \\
A2 type & Surface & 32/32 [100.00, 100.00] & 100.00 & \scNA & \scNA & 0 \\
A2 type & 0 & 24/32 [62.50, 87.50] & 75.00 & 31/32 [90.62, 100.00] & 93.75 & 0 \\
A2 type & 4 & 24/32 [62.50, 87.50] & 75.00 & 32/32 [100.00, 100.00] & 100.00 & 0 \\
A2 type & 9 & 24/32 [62.50, 87.50] & 75.00 & 32/32 [100.00, 100.00] & 100.00 & 0 \\
A2 type & 13 & 24/32 [62.50, 87.50] & 75.00 & 32/32 [100.00, 100.00] & 100.00 & 0 \\
A2 type & 18 & 24/32 [62.50, 87.50] & 75.00 & 32/32 [100.00, 100.00] & 100.00 & 0 \\
A2 type & 22 & 24/32 [62.50, 87.50] & 75.00 & 32/32 [100.00, 100.00] & 100.00 & 0 \\
A2 type & 27 & 24/32 [62.50, 87.50] & 75.00 & 32/32 [100.00, 100.00] & 100.00 & 0 \\
A2 type & 31 & 24/32 [62.50, 87.50] & 75.00 & 32/32 [100.00, 100.00] & 93.75 & 0 \\
Entity & Majority & 4/32 [0.00, 31.25] & 25.00 & \scNA & \scNA & 2 \\
Entity & Surface & 30/32 [81.25, 100.00] & 87.50 & \scNA & \scNA & 2 \\
Entity & 0 & 20/32 [37.50, 87.50] & 56.25 & 27/32 [65.62, 100.00] & 87.50 & 2 \\
Entity & 4 & 24/32 [50.00, 93.75] & 68.75 & 28/32 [68.75, 100.00] & 87.50 & 2 \\
Entity & 9 & 26/32 [62.50, 100.00] & 62.50 & 30/32 [81.25, 100.00] & 87.50 & 2 \\
Entity & 13 & 28/32 [68.75, 100.00] & 62.50 & 30/32 [81.25, 100.00] & 87.50 & 2 \\
Entity & 18 & 28/32 [68.75, 100.00] & 75.00 & 30/32 [81.25, 100.00] & 81.25 & 2 \\
Entity & 22 & 30/32 [81.25, 100.00] & 62.50 & 30/32 [81.25, 100.00] & 81.25 & 2 \\
Entity & 27 & 30/32 [81.25, 100.00] & 62.50 & 30/32 [81.25, 100.00] & 87.50 & 2 \\
Entity & 31 & 30/32 [81.25, 100.00] & 75.00 & 30/32 [81.25, 100.00] & 87.50 & 2 \\
Protected & Majority & 20/32 [37.50, 87.50] & 81.25 & \scNA & \scNA & 0 \\
Protected & Surface & 16/32 [25.00, 75.00] & 78.12 & \scNA & \scNA & 0 \\
Protected & 0 & 14/32 [18.75, 68.75] & 56.25 & 16/32 [25.00, 75.00] & 75.00 & 0 \\
Protected & 4 & 22/32 [43.75, 87.50] & 68.75 & 18/32 [34.38, 78.12] & 75.00 & 0 \\
Protected & 9 & 18/32 [31.25, 81.25] & 75.00 & 19/32 [34.38, 81.25] & 78.12 & 0 \\
Protected & 13 & 16/32 [25.00, 75.00] & 62.50 & 20/32 [37.50, 87.50] & 71.88 & 0 \\
Protected & 18 & 20/32 [37.50, 87.50] & 75.00 & 16/32 [25.00, 75.00] & 75.00 & 0 \\
Protected & 22 & 16/32 [25.00, 75.00] & 75.00 & 19/32 [34.38, 81.25] & 71.88 & 0 \\
Protected & 27 & 16/32 [25.00, 75.00] & 75.00 & 17/32 [28.12, 75.00] & 81.25 & 0 \\
Protected & 31 & 18/32 [31.25, 81.25] & 75.00 & 16/32 [25.00, 75.00] & 81.25 & 0 \\
\end{longtable}
\endgroup

\subsection{Qwen3.5-4B: B1/B06}
\label{app:sc_r1_full_4b_b1_b06}

Table~\ref{tab:sc_r1_full_4b_b1_b06} uses the explicit intermediate-checkpoint context. High checkpoint readability can reflect a value that is directly supplied in the input; it is not independent evidence that the unassisted model constructs or causally uses that state. All query-anchor results remain visible as a contrasting readout location.

% Source: supplied evidence report, lines 4295--5661.
\FloatBarrier
\begingroup
\scriptsize
\setlength{\tabcolsep}{3pt}
\renewcommand{\arraystretch}{1.08}
\begin{longtable}{L{0.105\linewidth}L{0.09\linewidth}L{0.25\linewidth}rL{0.25\linewidth}rr}
\caption{Qwen3.5-4B B1/B06 complete fixed-layer ledger. Residual rows are indexed by actual block; baseline rows have no block. Correct/N and confidence intervals are retained for each available anchor; selection values are descriptive. Evaluation: 32 sequences/16 worlds.}\label{tab:sc_r1_full_4b_b1_b06}\\
\toprule
Variable & Reader/block & Checkpoint: count [CI] & Sel. & Query: count [CI] & Sel. & $U$ \\
\midrule
\endfirsthead
\multicolumn{7}{l}{\footnotesize Table~\thetable\ continued.}\\
\toprule
Variable & Reader/block & Checkpoint: count [CI] & Sel. & Query: count [CI] & Sel. & $U$ \\
\midrule
\endhead
\midrule\multicolumn{7}{r}{\footnotesize Continued on the next page.}\\
\endfoot
\bottomrule
\endlastfoot
$Z$ & \multicolumn{6}{l}{Constant training label} \\
$s_0$ & Majority & 22/32 [43.75, 87.50] & 62.50 & \scNA & \scNA & 2 \\
$s_0$ & Surface & 21/32 [43.75, 87.50] & 71.88 & \scNA & \scNA & 2 \\
$s_0$ & 0 & 26/32 [62.50, 100.00] & 75.00 & 19/32 [34.38, 81.25] & 43.75 & 2 \\
$s_0$ & 4 & 30/32 [81.25, 100.00] & 93.75 & 15/32 [25.00, 68.75] & 40.62 & 2 \\
$s_0$ & 9 & 30/32 [81.25, 100.00] & 93.75 & 24/32 [56.25, 93.75] & 56.25 & 2 \\
$s_0$ & 13 & 30/32 [81.25, 100.00] & 93.75 & 20/32 [37.50, 87.50] & 78.12 & 2 \\
$s_0$ & 18 & 30/32 [81.25, 100.00] & 93.75 & 26/32 [62.50, 100.00] & 78.12 & 2 \\
$s_0$ & 22 & 30/32 [81.25, 100.00] & 81.25 & 21/32 [43.75, 87.50] & 56.25 & 2 \\
$s_0$ & 27 & 30/32 [81.25, 100.00] & 75.00 & 22/32 [43.75, 87.50] & 46.88 & 2 \\
$s_0$ & 31 & 30/32 [81.25, 100.00] & 87.50 & 22/32 [43.75, 87.50] & 50.00 & 2 \\
$s_1$ & Majority & 10/32 [12.50, 56.25] & 43.75 & \scNA & \scNA & 0 \\
$s_1$ & Surface & 26/32 [62.50, 96.88] & 96.88 & \scNA & \scNA & 0 \\
$s_1$ & 0 & 32/32 [100.00, 100.00] & 100.00 & 12/32 [12.50, 62.50] & 37.50 & 0 \\
$s_1$ & 4 & 32/32 [100.00, 100.00] & 100.00 & 24/32 [50.00, 93.75] & 56.25 & 0 \\
$s_1$ & 9 & 32/32 [100.00, 100.00] & 100.00 & 26/32 [62.50, 100.00] & 81.25 & 0 \\
$s_1$ & 13 & 32/32 [100.00, 100.00] & 100.00 & 28/32 [68.75, 100.00] & 75.00 & 0 \\
$s_1$ & 18 & 32/32 [100.00, 100.00] & 93.75 & 25/32 [59.38, 93.75] & 81.25 & 0 \\
$s_1$ & 22 & 30/32 [81.25, 100.00] & 93.75 & 24/32 [50.00, 93.75] & 78.12 & 0 \\
$s_1$ & 27 & 30/32 [81.25, 100.00] & 93.75 & 22/32 [43.75, 87.50] & 87.50 & 0 \\
$s_1$ & 31 & 30/32 [81.25, 100.00] & 93.75 & 25/32 [59.38, 93.75] & 75.00 & 0 \\
$s_2$ & Majority & 11/32 [21.88, 43.75] & 31.25 & \scNA & \scNA & 0 \\
$s_2$ & Surface & 24/32 [56.25, 93.75] & 78.12 & \scNA & \scNA & 0 \\
$s_2$ & 0 & 15/32 [40.62, 50.00] & 50.00 & 7/32 [6.25, 43.75] & 50.00 & 0 \\
$s_2$ & 4 & 16/32 [50.00, 50.00] & 50.00 & 15/32 [31.25, 65.62] & 53.12 & 0 \\
$s_2$ & 9 & 15/32 [40.62, 50.00] & 50.00 & 24/32 [56.25, 93.75] & 75.00 & 0 \\
$s_2$ & 13 & 15/32 [40.62, 50.00] & 46.88 & 26/32 [65.62, 93.75] & 81.25 & 0 \\
$s_2$ & 18 & 15/32 [40.62, 50.00] & 46.88 & 21/32 [43.75, 84.38] & 68.75 & 0 \\
$s_2$ & 22 & 15/32 [40.62, 50.00] & 46.88 & 24/32 [56.25, 93.75] & 71.88 & 0 \\
$s_2$ & 27 & 15/32 [40.62, 50.00] & 46.88 & 21/32 [46.88, 84.38] & 68.75 & 0 \\
$s_2$ & 31 & 15/32 [40.62, 50.00] & 46.88 & 20/32 [43.75, 81.25] & 46.88 & 0 \\
A1 amt. & Majority & 16/32 [25.00, 75.00] & 50.00 & \scNA & \scNA & 0 \\
A1 amt. & Surface & 31/32 [90.62, 100.00] & 100.00 & \scNA & \scNA & 0 \\
A1 amt. & 0 & 30/32 [81.25, 100.00] & 93.75 & 23/32 [50.00, 90.62] & 71.88 & 0 \\
A1 amt. & 4 & 32/32 [100.00, 100.00] & 100.00 & 32/32 [100.00, 100.00] & 93.75 & 0 \\
A1 amt. & 9 & 32/32 [100.00, 100.00] & 100.00 & 32/32 [100.00, 100.00] & 100.00 & 0 \\
A1 amt. & 13 & 32/32 [100.00, 100.00] & 100.00 & 32/32 [100.00, 100.00] & 100.00 & 0 \\
A1 amt. & 18 & 32/32 [100.00, 100.00] & 100.00 & 32/32 [100.00, 100.00] & 100.00 & 0 \\
A1 amt. & 22 & 32/32 [100.00, 100.00] & 100.00 & 32/32 [100.00, 100.00] & 100.00 & 0 \\
A1 amt. & 27 & 32/32 [100.00, 100.00] & 100.00 & 32/32 [100.00, 100.00] & 100.00 & 0 \\
A1 amt. & 31 & 32/32 [100.00, 100.00] & 100.00 & 32/32 [100.00, 100.00] & 100.00 & 0 \\
A2 amt. & Majority & 16/32 [50.00, 50.00] & 50.00 & \scNA & \scNA & 0 \\
A2 amt. & Surface & 30/32 [84.38, 100.00] & 96.88 & \scNA & \scNA & 0 \\
A2 amt. & 0 & 16/32 [50.00, 50.00] & 50.00 & 25/32 [65.62, 90.62] & 75.00 & 0 \\
A2 amt. & 4 & 16/32 [50.00, 50.00] & 50.00 & 31/32 [90.62, 100.00] & 96.88 & 0 \\
A2 amt. & 9 & 16/32 [50.00, 50.00] & 50.00 & 32/32 [100.00, 100.00] & 100.00 & 0 \\
A2 amt. & 13 & 16/32 [50.00, 50.00] & 50.00 & 32/32 [100.00, 100.00] & 100.00 & 0 \\
A2 amt. & 18 & 16/32 [50.00, 50.00] & 50.00 & 32/32 [100.00, 100.00] & 100.00 & 0 \\
A2 amt. & 22 & 16/32 [50.00, 50.00] & 50.00 & 32/32 [100.00, 100.00] & 100.00 & 0 \\
A2 amt. & 27 & 16/32 [50.00, 50.00] & 50.00 & 32/32 [100.00, 100.00] & 100.00 & 0 \\
A2 amt. & 31 & 16/32 [50.00, 50.00] & 50.00 & 32/32 [100.00, 100.00] & 100.00 & 0 \\
A1 type & Majority & 16/32 [25.00, 75.00] & 50.00 & \scNA & \scNA & 0 \\
A1 type & Surface & 31/32 [90.62, 100.00] & 100.00 & \scNA & \scNA & 0 \\
A1 type & 0 & 30/32 [81.25, 100.00] & 93.75 & 23/32 [50.00, 90.62] & 71.88 & 0 \\
A1 type & 4 & 32/32 [100.00, 100.00] & 100.00 & 32/32 [100.00, 100.00] & 93.75 & 0 \\
A1 type & 9 & 32/32 [100.00, 100.00] & 100.00 & 32/32 [100.00, 100.00] & 100.00 & 0 \\
A1 type & 13 & 32/32 [100.00, 100.00] & 100.00 & 32/32 [100.00, 100.00] & 100.00 & 0 \\
A1 type & 18 & 32/32 [100.00, 100.00] & 100.00 & 32/32 [100.00, 100.00] & 100.00 & 0 \\
A1 type & 22 & 32/32 [100.00, 100.00] & 100.00 & 32/32 [100.00, 100.00] & 100.00 & 0 \\
A1 type & 27 & 32/32 [100.00, 100.00] & 100.00 & 32/32 [100.00, 100.00] & 100.00 & 0 \\
A1 type & 31 & 32/32 [100.00, 100.00] & 100.00 & 32/32 [100.00, 100.00] & 100.00 & 0 \\
A2 type & Majority & 24/32 [62.50, 87.50] & 75.00 & \scNA & \scNA & 0 \\
A2 type & Surface & 31/32 [90.62, 100.00] & 100.00 & \scNA & \scNA & 0 \\
A2 type & 0 & 24/32 [62.50, 87.50] & 75.00 & 31/32 [90.62, 100.00] & 90.62 & 0 \\
A2 type & 4 & 24/32 [62.50, 87.50] & 75.00 & 32/32 [100.00, 100.00] & 100.00 & 0 \\
A2 type & 9 & 24/32 [62.50, 87.50] & 75.00 & 32/32 [100.00, 100.00] & 100.00 & 0 \\
A2 type & 13 & 24/32 [62.50, 87.50] & 75.00 & 32/32 [100.00, 100.00] & 100.00 & 0 \\
A2 type & 18 & 24/32 [62.50, 87.50] & 75.00 & 32/32 [100.00, 100.00] & 100.00 & 0 \\
A2 type & 22 & 24/32 [62.50, 87.50] & 75.00 & 32/32 [100.00, 100.00] & 100.00 & 0 \\
A2 type & 27 & 24/32 [62.50, 87.50] & 75.00 & 32/32 [100.00, 100.00] & 100.00 & 0 \\
A2 type & 31 & 24/32 [62.50, 87.50] & 75.00 & 32/32 [100.00, 100.00] & 100.00 & 0 \\
Entity & Majority & 4/32 [0.00, 31.25] & 25.00 & \scNA & \scNA & 2 \\
Entity & Surface & 30/32 [81.25, 100.00] & 87.50 & \scNA & \scNA & 2 \\
Entity & 0 & 12/32 [12.50, 62.50] & 37.50 & 26/32 [62.50, 100.00] & 87.50 & 2 \\
Entity & 4 & 22/32 [43.75, 87.50] & 43.75 & 30/32 [81.25, 100.00] & 87.50 & 2 \\
Entity & 9 & 26/32 [62.50, 100.00] & 56.25 & 30/32 [81.25, 100.00] & 87.50 & 2 \\
Entity & 13 & 26/32 [62.50, 100.00] & 56.25 & 30/32 [81.25, 100.00] & 87.50 & 2 \\
Entity & 18 & 24/32 [50.00, 93.75] & 62.50 & 29/32 [75.00, 100.00] & 75.00 & 2 \\
Entity & 22 & 26/32 [62.50, 100.00] & 56.25 & 30/32 [81.25, 100.00] & 78.12 & 2 \\
Entity & 27 & 30/32 [81.25, 100.00] & 87.50 & 30/32 [81.25, 100.00] & 87.50 & 2 \\
Entity & 31 & 30/32 [81.25, 100.00] & 87.50 & 30/32 [81.25, 100.00] & 87.50 & 2 \\
Protected & Majority & 20/32 [37.50, 87.50] & 81.25 & \scNA & \scNA & 0 \\
Protected & Surface & 19/32 [34.38, 81.25] & 75.00 & \scNA & \scNA & 0 \\
Protected & 0 & 20/32 [37.50, 87.50] & 68.75 & 14/32 [18.75, 68.75] & 75.00 & 0 \\
Protected & 4 & 20/32 [37.50, 87.50] & 62.50 & 16/32 [25.00, 75.00] & 75.00 & 0 \\
Protected & 9 & 18/32 [31.25, 81.25] & 50.00 & 18/32 [31.25, 81.25] & 81.25 & 0 \\
Protected & 13 & 18/32 [31.25, 81.25] & 56.25 & 18/32 [31.25, 81.25] & 78.12 & 0 \\
Protected & 18 & 18/32 [31.25, 81.25] & 62.50 & 19/32 [34.38, 81.25] & 71.88 & 0 \\
Protected & 22 & 20/32 [37.50, 87.50] & 56.25 & 16/32 [28.12, 71.95] & 68.75 & 0 \\
Protected & 27 & 20/32 [37.50, 87.50] & 56.25 & 16/32 [28.12, 71.88] & 68.75 & 0 \\
Protected & 31 & 20/32 [37.50, 87.50] & 37.50 & 17/32 [31.25, 75.00] & 81.25 & 0 \\
\end{longtable}
\endgroup

\subsection{Qwen3.5-4B: B2/FINAL}
\label{app:sc_r1_full_4b_b2_final}

Table~\ref{tab:sc_r1_full_4b_b2_final} reports the symbolic final-query panel. The evaluation has only three program-family clusters, and entity/protected labels are not supplied. Readability of action fields that already appear in the prompt is distinct from successful counterfactual state interchange.

% Source: supplied evidence report, lines 4295--5661.
\FloatBarrier
\begingroup
\scriptsize
\setlength{\tabcolsep}{3pt}
\renewcommand{\arraystretch}{1.08}
\begin{longtable}{L{0.105\linewidth}L{0.09\linewidth}L{0.25\linewidth}rL{0.25\linewidth}rr}
\caption{Qwen3.5-4B B2/FINAL complete fixed-layer ledger. Residual rows are indexed by actual block; baseline rows have no block. Correct/N and confidence intervals are retained for each available anchor; selection values are descriptive. Evaluation: 18 sequences/3 program families.}\label{tab:sc_r1_full_4b_b2_final}\\
\toprule
Variable & Reader/block & Checkpoint: count [CI] & Sel. & Query: count [CI] & Sel. & $U$ \\
\midrule
\endfirsthead
\multicolumn{7}{l}{\footnotesize Table~\thetable\ continued.}\\
\toprule
Variable & Reader/block & Checkpoint: count [CI] & Sel. & Query: count [CI] & Sel. & $U$ \\
\midrule
\endhead
\midrule\multicolumn{7}{r}{\footnotesize Continued on the next page.}\\
\endfoot
\bottomrule
\endlastfoot
$Z$ & \multicolumn{6}{l}{Constant training label} \\
Entity & \multicolumn{6}{l}{Label not supplied} \\
Protected & \multicolumn{6}{l}{Label not supplied} \\
$s_0$ & Majority & 0/18 [0.00, 0.00] & 0.00 & \scNA & \scNA & 2 \\
$s_0$ & Surface & 11/18 [16.67, 83.33] & 16.67 & \scNA & \scNA & 2 \\
$s_0$ & 0 & 4/18 [0.00, 50.00] & 16.67 & 2/18 [0.00, 16.67] & 16.67 & 2 \\
$s_0$ & 4 & 14/18 [66.67, 83.33] & 16.67 & 16/18 [83.33, 100.00] & 16.67 & 2 \\
$s_0$ & 9 & 11/18 [50.00, 66.67] & 16.67 & 16/18 [83.33, 100.00] & 16.67 & 2 \\
$s_0$ & 13 & 13/18 [66.67, 83.33] & 16.67 & 16/18 [83.33, 100.00] & 16.67 & 2 \\
$s_0$ & 18 & 13/18 [66.67, 83.33] & 16.67 & 11/18 [50.00, 83.33] & 16.67 & 2 \\
$s_0$ & 22 & 11/18 [50.00, 66.67] & 16.67 & 12/18 [50.00, 100.00] & 16.67 & 2 \\
$s_0$ & 27 & 11/18 [50.00, 66.67] & 16.67 & 11/18 [16.67, 83.33] & 16.67 & 2 \\
$s_0$ & 31 & 12/18 [50.00, 83.33] & 16.67 & 10/18 [16.67, 83.33] & 16.67 & 2 \\
$s_1$ & Majority & 3/18 [16.67, 16.67] & 33.33 & \scNA & \scNA & 0 \\
$s_1$ & Surface & 1/18 [0.00, 16.67] & 16.67 & \scNA & \scNA & 0 \\
$s_1$ & 0 & 3/18 [16.67, 16.67] & 22.22 & 3/18 [16.67, 16.67] & 22.22 & 0 \\
$s_1$ & 4 & 3/18 [16.67, 16.67] & 16.67 & 3/18 [16.67, 16.67] & 16.67 & 0 \\
$s_1$ & 9 & 2/18 [0.00, 16.67] & 16.67 & 3/18 [16.67, 16.67] & 16.67 & 0 \\
$s_1$ & 13 & 2/18 [0.00, 16.67] & 16.67 & 2/18 [0.00, 16.67] & 16.67 & 0 \\
$s_1$ & 18 & 2/18 [0.00, 16.67] & 16.67 & 2/18 [0.00, 16.67] & 16.67 & 0 \\
$s_1$ & 22 & 3/18 [16.67, 16.67] & 16.67 & 3/18 [16.67, 16.67] & 16.67 & 0 \\
$s_1$ & 27 & 2/18 [0.00, 16.67] & 16.67 & 3/18 [16.67, 16.67] & 16.67 & 0 \\
$s_1$ & 31 & 2/18 [0.00, 16.67] & 16.67 & 3/18 [16.67, 16.67] & 22.22 & 0 \\
$s_2$ & Majority & 3/18 [16.67, 16.67] & 22.22 & \scNA & \scNA & 5 \\
$s_2$ & Surface & 3/18 [16.67, 16.67] & 16.67 & \scNA & \scNA & 5 \\
$s_2$ & 0 & 4/18 [16.67, 33.33] & 16.67 & 5/18 [16.67, 33.33] & 16.67 & 5 \\
$s_2$ & 4 & 6/18 [33.33, 33.33] & 22.22 & 5/18 [16.67, 33.33] & 16.67 & 5 \\
$s_2$ & 9 & 4/18 [16.67, 33.33] & 22.22 & 5/18 [16.67, 33.33] & 16.67 & 5 \\
$s_2$ & 13 & 4/18 [0.00, 33.33] & 16.67 & 4/18 [16.67, 33.33] & 16.67 & 5 \\
$s_2$ & 18 & 5/18 [16.67, 33.33] & 22.22 & 4/18 [16.67, 33.33] & 16.67 & 5 \\
$s_2$ & 22 & 6/18 [33.33, 33.33] & 22.22 & 6/18 [33.33, 33.33] & 16.67 & 5 \\
$s_2$ & 27 & 6/18 [33.33, 33.33] & 16.67 & 4/18 [16.67, 33.33] & 22.22 & 5 \\
$s_2$ & 31 & 6/18 [33.33, 33.33] & 27.78 & 6/18 [33.33, 33.33] & 16.67 & 5 \\
A1 amt. & Majority & 9/18 [50.00, 50.00] & 50.00 & \scNA & \scNA & 0 \\
A1 amt. & Surface & 16/18 [66.67, 100.00] & 100.00 & \scNA & \scNA & 0 \\
A1 amt. & 0 & 18/18 [100.00, 100.00] & 100.00 & 18/18 [100.00, 100.00] & 100.00 & 0 \\
A1 amt. & 4 & 18/18 [100.00, 100.00] & 100.00 & 18/18 [100.00, 100.00] & 100.00 & 0 \\
A1 amt. & 9 & 18/18 [100.00, 100.00] & 100.00 & 18/18 [100.00, 100.00] & 100.00 & 0 \\
A1 amt. & 13 & 18/18 [100.00, 100.00] & 100.00 & 18/18 [100.00, 100.00] & 100.00 & 0 \\
A1 amt. & 18 & 18/18 [100.00, 100.00] & 100.00 & 18/18 [100.00, 100.00] & 100.00 & 0 \\
A1 amt. & 22 & 18/18 [100.00, 100.00] & 100.00 & 18/18 [100.00, 100.00] & 94.44 & 0 \\
A1 amt. & 27 & 18/18 [100.00, 100.00] & 100.00 & 18/18 [100.00, 100.00] & 100.00 & 0 \\
A1 amt. & 31 & 18/18 [100.00, 100.00] & 100.00 & 18/18 [100.00, 100.00] & 100.00 & 0 \\
A2 amt. & Majority & 6/18 [33.33, 33.33] & 33.33 & \scNA & \scNA & 0 \\
A2 amt. & Surface & 18/18 [100.00, 100.00] & 100.00 & \scNA & \scNA & 0 \\
A2 amt. & 0 & 15/18 [83.33, 83.33] & 83.33 & 18/18 [100.00, 100.00] & 100.00 & 0 \\
A2 amt. & 4 & 15/18 [83.33, 83.33] & 83.33 & 18/18 [100.00, 100.00] & 100.00 & 0 \\
A2 amt. & 9 & 15/18 [83.33, 83.33] & 83.33 & 18/18 [100.00, 100.00] & 100.00 & 0 \\
A2 amt. & 13 & 15/18 [83.33, 83.33] & 83.33 & 18/18 [100.00, 100.00] & 100.00 & 0 \\
A2 amt. & 18 & 15/18 [83.33, 83.33] & 83.33 & 18/18 [100.00, 100.00] & 100.00 & 0 \\
A2 amt. & 22 & 15/18 [83.33, 83.33] & 83.33 & 18/18 [100.00, 100.00] & 100.00 & 0 \\
A2 amt. & 27 & 15/18 [83.33, 83.33] & 83.33 & 18/18 [100.00, 100.00] & 100.00 & 0 \\
A2 amt. & 31 & 15/18 [83.33, 83.33] & 83.33 & 18/18 [100.00, 100.00] & 100.00 & 0 \\
A1 type & Majority & 15/18 [83.33, 83.33] & 83.33 & \scNA & \scNA & 0 \\
A1 type & Surface & 18/18 [100.00, 100.00] & 100.00 & \scNA & \scNA & 0 \\
A1 type & 0 & 18/18 [100.00, 100.00] & 100.00 & 18/18 [100.00, 100.00] & 100.00 & 0 \\
A1 type & 4 & 18/18 [100.00, 100.00] & 100.00 & 18/18 [100.00, 100.00] & 100.00 & 0 \\
A1 type & 9 & 18/18 [100.00, 100.00] & 100.00 & 18/18 [100.00, 100.00] & 100.00 & 0 \\
A1 type & 13 & 18/18 [100.00, 100.00] & 100.00 & 18/18 [100.00, 100.00] & 100.00 & 0 \\
A1 type & 18 & 18/18 [100.00, 100.00] & 100.00 & 18/18 [100.00, 100.00] & 100.00 & 0 \\
A1 type & 22 & 18/18 [100.00, 100.00] & 100.00 & 18/18 [100.00, 100.00] & 100.00 & 0 \\
A1 type & 27 & 18/18 [100.00, 100.00] & 100.00 & 18/18 [100.00, 100.00] & 100.00 & 0 \\
A1 type & 31 & 18/18 [100.00, 100.00] & 100.00 & 18/18 [100.00, 100.00] & 100.00 & 0 \\
A2 type & Majority & 12/18 [66.67, 66.67] & 66.67 & \scNA & \scNA & 0 \\
A2 type & Surface & 18/18 [100.00, 100.00] & 100.00 & \scNA & \scNA & 0 \\
A2 type & 0 & 15/18 [83.33, 83.33] & 83.33 & 18/18 [100.00, 100.00] & 100.00 & 0 \\
A2 type & 4 & 15/18 [83.33, 83.33] & 83.33 & 18/18 [100.00, 100.00] & 100.00 & 0 \\
A2 type & 9 & 15/18 [83.33, 83.33] & 83.33 & 18/18 [100.00, 100.00] & 100.00 & 0 \\
A2 type & 13 & 15/18 [83.33, 83.33] & 83.33 & 18/18 [100.00, 100.00] & 100.00 & 0 \\
A2 type & 18 & 15/18 [83.33, 83.33] & 83.33 & 18/18 [100.00, 100.00] & 100.00 & 0 \\
A2 type & 22 & 15/18 [83.33, 83.33] & 83.33 & 18/18 [100.00, 100.00] & 100.00 & 0 \\
A2 type & 27 & 15/18 [83.33, 83.33] & 83.33 & 18/18 [100.00, 100.00] & 100.00 & 0 \\
A2 type & 31 & 15/18 [83.33, 83.33] & 83.33 & 18/18 [100.00, 100.00] & 100.00 & 0 \\
\end{longtable}
\endgroup

\subsection{Qwen3.5-9B: B1/B01}
\label{app:sc_r1_full_9b_b1_b01}

Table~\ref{tab:sc_r1_full_9b_b1_b01} includes initial and final state variables, both actions, entity identity, and the protected fact for the original full visual context. The complete table prevents the prespecified S1 summary from being read as a claim that every variable is equally unreadable. Class-coverage limitations and the constant target label are retained.

% Source: supplied evidence report, lines 4295--5661.
\FloatBarrier
\begingroup
\scriptsize
\setlength{\tabcolsep}{3pt}
\renewcommand{\arraystretch}{1.08}
\begin{longtable}{L{0.105\linewidth}L{0.09\linewidth}L{0.25\linewidth}rL{0.25\linewidth}rr}
\caption{Qwen3.5-9B B1/B01 complete fixed-layer ledger. Residual rows are indexed by actual block; baseline rows have no block. Correct/N and confidence intervals are retained for each available anchor; selection values are descriptive. Evaluation: 32 sequences/16 worlds.}\label{tab:sc_r1_full_9b_b1_b01}\\
\toprule
Variable & Reader/block & Checkpoint: count [CI] & Sel. & Query: count [CI] & Sel. & $U$ \\
\midrule
\endfirsthead
\multicolumn{7}{l}{\footnotesize Table~\thetable\ continued.}\\
\toprule
Variable & Reader/block & Checkpoint: count [CI] & Sel. & Query: count [CI] & Sel. & $U$ \\
\midrule
\endhead
\midrule\multicolumn{7}{r}{\footnotesize Continued on the next page.}\\
\endfoot
\bottomrule
\endlastfoot
$Z$ & \multicolumn{6}{l}{Constant training label} \\
$s_0$ & Majority & 22/32 [43.75, 87.50] & 62.50 & \scNA & \scNA & 2 \\
$s_0$ & Surface & 21/32 [43.75, 87.50] & 40.62 & \scNA & \scNA & 2 \\
$s_0$ & 0 & 16/32 [25.00, 75.00] & 43.75 & 20/32 [37.50, 87.50] & 37.50 & 2 \\
$s_0$ & 4 & 16/32 [25.00, 75.00] & 37.50 & 19/32 [34.38, 81.25] & 43.75 & 2 \\
$s_0$ & 9 & 18/32 [31.25, 81.25] & 43.75 & 19/32 [37.50, 81.25] & 37.50 & 2 \\
$s_0$ & 13 & 18/32 [31.25, 81.25] & 50.00 & 17/32 [31.17, 75.00] & 43.75 & 2 \\
$s_0$ & 18 & 18/32 [31.25, 81.25] & 43.75 & 18/32 [31.25, 81.25] & 46.88 & 2 \\
$s_0$ & 22 & 20/32 [37.50, 87.50] & 37.50 & 23/32 [50.00, 90.62] & 46.88 & 2 \\
$s_0$ & 27 & 20/32 [37.50, 87.50] & 37.50 & 21/32 [43.75, 87.50] & 34.38 & 2 \\
$s_0$ & 31 & 20/32 [37.50, 81.25] & 37.50 & 20/32 [37.50, 87.50] & 34.38 & 2 \\
$s_1$ & Majority & 10/32 [12.50, 56.25] & 43.75 & \scNA & \scNA & 0 \\
$s_1$ & Surface & 22/32 [46.88, 87.50] & 65.62 & \scNA & \scNA & 0 \\
$s_1$ & 0 & 16/32 [25.00, 75.00] & 50.00 & 14/32 [18.75, 68.75] & 15.62 & 0 \\
$s_1$ & 4 & 18/32 [31.25, 81.25] & 25.00 & 16/32 [25.00, 75.00] & 31.25 & 0 \\
$s_1$ & 9 & 20/32 [37.50, 81.25] & 43.75 & 20/32 [40.62, 81.33] & 40.62 & 0 \\
$s_1$ & 13 & 16/32 [25.00, 75.00] & 50.00 & 19/32 [37.50, 81.25] & 40.62 & 0 \\
$s_1$ & 18 & 18/32 [31.25, 81.25] & 62.50 & 19/32 [37.50, 81.25] & 53.12 & 0 \\
$s_1$ & 22 & 20/32 [37.50, 87.50] & 56.25 & 18/32 [31.25, 81.25] & 46.88 & 0 \\
$s_1$ & 27 & 22/32 [43.75, 87.50] & 56.25 & 18/32 [31.25, 81.25] & 37.50 & 0 \\
$s_1$ & 31 & 20/32 [37.50, 81.25] & 56.25 & 16/32 [25.00, 75.00] & 50.00 & 0 \\
$s_2$ & Majority & 11/32 [21.88, 43.75] & 31.25 & \scNA & \scNA & 0 \\
$s_2$ & Surface & 23/32 [50.00, 90.62] & 56.25 & \scNA & \scNA & 0 \\
$s_2$ & 0 & 8/32 [12.50, 37.50] & 21.88 & 13/32 [25.00, 59.38] & 34.38 & 0 \\
$s_2$ & 4 & 12/32 [28.12, 46.88] & 25.00 & 18/32 [37.50, 75.00] & 37.50 & 0 \\
$s_2$ & 9 & 9/32 [15.62, 40.62] & 21.88 & 17/32 [34.38, 71.88] & 34.38 & 0 \\
$s_2$ & 13 & 9/32 [15.62, 40.62] & 21.88 & 18/32 [34.38, 75.00] & 43.75 & 0 \\
$s_2$ & 18 & 11/32 [21.88, 43.75] & 28.12 & 17/32 [34.38, 71.88] & 43.75 & 0 \\
$s_2$ & 22 & 11/32 [21.88, 43.75] & 31.25 & 20/32 [40.62, 84.38] & 40.62 & 0 \\
$s_2$ & 27 & 12/32 [28.12, 46.88] & 28.12 & 20/32 [40.62, 84.38] & 46.88 & 0 \\
$s_2$ & 31 & 11/32 [21.88, 43.75] & 28.12 & 18/32 [37.50, 75.00] & 37.50 & 0 \\
A1 amt. & Majority & 16/32 [25.00, 75.00] & 50.00 & \scNA & \scNA & 0 \\
A1 amt. & Surface & 32/32 [100.00, 100.00] & 100.00 & \scNA & \scNA & 0 \\
A1 amt. & 0 & 32/32 [100.00, 100.00] & 100.00 & 29/32 [75.00, 100.00] & 50.00 & 0 \\
A1 amt. & 4 & 32/32 [100.00, 100.00] & 100.00 & 32/32 [100.00, 100.00] & 100.00 & 0 \\
A1 amt. & 9 & 32/32 [100.00, 100.00] & 100.00 & 32/32 [100.00, 100.00] & 100.00 & 0 \\
A1 amt. & 13 & 32/32 [100.00, 100.00] & 100.00 & 32/32 [100.00, 100.00] & 100.00 & 0 \\
A1 amt. & 18 & 32/32 [100.00, 100.00] & 100.00 & 32/32 [100.00, 100.00] & 100.00 & 0 \\
A1 amt. & 22 & 32/32 [100.00, 100.00] & 100.00 & 32/32 [100.00, 100.00] & 100.00 & 0 \\
A1 amt. & 27 & 32/32 [100.00, 100.00] & 100.00 & 32/32 [100.00, 100.00] & 100.00 & 0 \\
A1 amt. & 31 & 32/32 [100.00, 100.00] & 100.00 & 32/32 [100.00, 100.00] & 100.00 & 0 \\
A2 amt. & Majority & 16/32 [50.00, 50.00] & 50.00 & \scNA & \scNA & 0 \\
A2 amt. & Surface & 32/32 [100.00, 100.00] & 100.00 & \scNA & \scNA & 0 \\
A2 amt. & 0 & 16/32 [50.00, 50.00] & 50.00 & 30/32 [84.38, 100.00] & 87.50 & 0 \\
A2 amt. & 4 & 16/32 [50.00, 50.00] & 50.00 & 32/32 [100.00, 100.00] & 100.00 & 0 \\
A2 amt. & 9 & 16/32 [50.00, 50.00] & 50.00 & 32/32 [100.00, 100.00] & 100.00 & 0 \\
A2 amt. & 13 & 16/32 [50.00, 50.00] & 50.00 & 32/32 [100.00, 100.00] & 100.00 & 0 \\
A2 amt. & 18 & 16/32 [50.00, 50.00] & 50.00 & 32/32 [100.00, 100.00] & 100.00 & 0 \\
A2 amt. & 22 & 16/32 [50.00, 50.00] & 50.00 & 32/32 [100.00, 100.00] & 100.00 & 0 \\
A2 amt. & 27 & 16/32 [50.00, 50.00] & 50.00 & 32/32 [100.00, 100.00] & 100.00 & 0 \\
A2 amt. & 31 & 16/32 [50.00, 50.00] & 50.00 & 31/32 [90.62, 100.00] & 96.88 & 0 \\
A1 type & Majority & 16/32 [25.00, 75.00] & 50.00 & \scNA & \scNA & 0 \\
A1 type & Surface & 32/32 [100.00, 100.00] & 100.00 & \scNA & \scNA & 0 \\
A1 type & 0 & 32/32 [100.00, 100.00] & 100.00 & 29/32 [75.00, 100.00] & 50.00 & 0 \\
A1 type & 4 & 32/32 [100.00, 100.00] & 100.00 & 32/32 [100.00, 100.00] & 100.00 & 0 \\
A1 type & 9 & 32/32 [100.00, 100.00] & 100.00 & 32/32 [100.00, 100.00] & 100.00 & 0 \\
A1 type & 13 & 32/32 [100.00, 100.00] & 100.00 & 32/32 [100.00, 100.00] & 100.00 & 0 \\
A1 type & 18 & 32/32 [100.00, 100.00] & 100.00 & 32/32 [100.00, 100.00] & 100.00 & 0 \\
A1 type & 22 & 32/32 [100.00, 100.00] & 100.00 & 32/32 [100.00, 100.00] & 100.00 & 0 \\
A1 type & 27 & 32/32 [100.00, 100.00] & 100.00 & 32/32 [100.00, 100.00] & 100.00 & 0 \\
A1 type & 31 & 32/32 [100.00, 100.00] & 100.00 & 32/32 [100.00, 100.00] & 100.00 & 0 \\
A2 type & Majority & 24/32 [62.50, 87.50] & 75.00 & \scNA & \scNA & 0 \\
A2 type & Surface & 32/32 [100.00, 100.00] & 100.00 & \scNA & \scNA & 0 \\
A2 type & 0 & 24/32 [62.50, 87.50] & 75.00 & 32/32 [100.00, 100.00] & 87.50 & 0 \\
A2 type & 4 & 24/32 [62.50, 87.50] & 75.00 & 32/32 [100.00, 100.00] & 100.00 & 0 \\
A2 type & 9 & 24/32 [62.50, 87.50] & 75.00 & 32/32 [100.00, 100.00] & 100.00 & 0 \\
A2 type & 13 & 24/32 [62.50, 87.50] & 75.00 & 32/32 [100.00, 100.00] & 100.00 & 0 \\
A2 type & 18 & 24/32 [62.50, 87.50] & 75.00 & 32/32 [100.00, 100.00] & 100.00 & 0 \\
A2 type & 22 & 24/32 [62.50, 87.50] & 75.00 & 32/32 [100.00, 100.00] & 100.00 & 0 \\
A2 type & 27 & 24/32 [62.50, 87.50] & 75.00 & 32/32 [100.00, 100.00] & 100.00 & 0 \\
A2 type & 31 & 24/32 [62.50, 87.50] & 75.00 & 32/32 [100.00, 100.00] & 100.00 & 0 \\
Entity & Majority & 4/32 [0.00, 31.25] & 25.00 & \scNA & \scNA & 2 \\
Entity & Surface & 30/32 [81.25, 100.00] & 87.50 & \scNA & \scNA & 2 \\
Entity & 0 & 28/32 [68.75, 100.00] & 62.50 & 29/32 [75.00, 100.00] & 87.50 & 2 \\
Entity & 4 & 28/32 [68.75, 100.00] & 68.75 & 30/32 [81.25, 100.00] & 87.50 & 2 \\
Entity & 9 & 30/32 [81.25, 100.00] & 75.00 & 30/32 [81.25, 100.00] & 87.50 & 2 \\
Entity & 13 & 28/32 [68.75, 100.00] & 68.75 & 30/32 [81.25, 100.00] & 87.50 & 2 \\
Entity & 18 & 28/32 [68.75, 100.00] & 68.75 & 30/32 [81.25, 100.00] & 87.50 & 2 \\
Entity & 22 & 26/32 [62.50, 100.00] & 75.00 & 30/32 [81.25, 100.00] & 84.38 & 2 \\
Entity & 27 & 30/32 [81.25, 100.00] & 81.25 & 30/32 [81.25, 100.00] & 87.50 & 2 \\
Entity & 31 & 30/32 [81.25, 100.00] & 87.50 & 30/32 [81.25, 100.00] & 87.50 & 2 \\
Protected & Majority & 20/32 [37.50, 87.50] & 81.25 & \scNA & \scNA & 0 \\
Protected & Surface & 16/32 [25.00, 75.00] & 78.12 & \scNA & \scNA & 0 \\
Protected & 0 & 18/32 [31.25, 81.25] & 81.25 & 16/32 [25.00, 75.00] & 87.50 & 0 \\
Protected & 4 & 18/32 [31.25, 81.25] & 87.50 & 15/32 [25.00, 71.88] & 68.75 & 0 \\
Protected & 9 & 14/32 [18.75, 68.75] & 81.25 & 14/32 [18.75, 68.75] & 68.75 & 0 \\
Protected & 13 & 16/32 [25.00, 75.00] & 68.75 & 13/32 [18.75, 65.62] & 75.00 & 0 \\
Protected & 18 & 16/32 [25.00, 75.00] & 81.25 & 14/32 [18.75, 68.75] & 75.00 & 0 \\
Protected & 22 & 16/32 [25.00, 75.00] & 68.75 & 13/32 [18.75, 65.62] & 71.88 & 0 \\
Protected & 27 & 14/32 [18.75, 68.75] & 62.50 & 15/32 [25.00, 68.75] & 75.00 & 0 \\
Protected & 31 & 16/32 [25.00, 75.00] & 68.75 & 20/32 [37.50, 87.50] & 78.12 & 0 \\
\end{longtable}
\endgroup

\subsection{Qwen3.5-9B: B1/B06}
\label{app:sc_r1_full_9b_b1_b06}

Table~\ref{tab:sc_r1_full_9b_b1_b06} uses the explicit intermediate-checkpoint context. High checkpoint readability can reflect a value that is directly supplied in the input; it is not independent evidence that the unassisted model constructs or causally uses that state. All query-anchor results remain visible as a contrasting readout location.

% Source: supplied evidence report, lines 4295--5661.
\FloatBarrier
\begingroup
\scriptsize
\setlength{\tabcolsep}{3pt}
\renewcommand{\arraystretch}{1.08}
\begin{longtable}{L{0.105\linewidth}L{0.09\linewidth}L{0.25\linewidth}rL{0.25\linewidth}rr}
\caption{Qwen3.5-9B B1/B06 complete fixed-layer ledger. Residual rows are indexed by actual block; baseline rows have no block. Correct/N and confidence intervals are retained for each available anchor; selection values are descriptive. Evaluation: 32 sequences/16 worlds.}\label{tab:sc_r1_full_9b_b1_b06}\\
\toprule
Variable & Reader/block & Checkpoint: count [CI] & Sel. & Query: count [CI] & Sel. & $U$ \\
\midrule
\endfirsthead
\multicolumn{7}{l}{\footnotesize Table~\thetable\ continued.}\\
\toprule
Variable & Reader/block & Checkpoint: count [CI] & Sel. & Query: count [CI] & Sel. & $U$ \\
\midrule
\endhead
\midrule\multicolumn{7}{r}{\footnotesize Continued on the next page.}\\
\endfoot
\bottomrule
\endlastfoot
$Z$ & \multicolumn{6}{l}{Constant training label} \\
$s_0$ & Majority & 22/32 [43.75, 87.50] & 62.50 & \scNA & \scNA & 2 \\
$s_0$ & Surface & 21/32 [43.75, 87.50] & 71.88 & \scNA & \scNA & 2 \\
$s_0$ & 0 & 26/32 [62.50, 100.00] & 87.50 & 20/32 [37.50, 87.50] & 37.50 & 2 \\
$s_0$ & 4 & 30/32 [81.25, 100.00] & 93.75 & 21/32 [43.75, 87.50] & 50.00 & 2 \\
$s_0$ & 9 & 30/32 [81.25, 100.00] & 93.75 & 20/32 [37.50, 87.50] & 43.75 & 2 \\
$s_0$ & 13 & 30/32 [81.25, 100.00] & 93.75 & 20/32 [37.50, 87.50] & 53.12 & 2 \\
$s_0$ & 18 & 30/32 [81.25, 100.00] & 93.75 & 20/32 [37.50, 87.50] & 62.50 & 2 \\
$s_0$ & 22 & 30/32 [81.25, 100.00] & 93.75 & 22/32 [43.75, 87.50] & 46.88 & 2 \\
$s_0$ & 27 & 30/32 [81.25, 100.00] & 93.75 & 24/32 [50.00, 93.75] & 43.75 & 2 \\
$s_0$ & 31 & 30/32 [81.25, 100.00] & 93.75 & 20/32 [37.50, 87.50] & 34.38 & 2 \\
$s_1$ & Majority & 10/32 [12.50, 56.25] & 43.75 & \scNA & \scNA & 0 \\
$s_1$ & Surface & 26/32 [62.50, 96.88] & 96.88 & \scNA & \scNA & 0 \\
$s_1$ & 0 & 32/32 [100.00, 100.00] & 100.00 & 14/32 [18.75, 68.75] & 18.75 & 0 \\
$s_1$ & 4 & 32/32 [100.00, 100.00] & 100.00 & 19/32 [37.50, 81.25] & 37.50 & 0 \\
$s_1$ & 9 & 32/32 [100.00, 100.00] & 100.00 & 22/32 [43.75, 87.50] & 81.25 & 0 \\
$s_1$ & 13 & 32/32 [100.00, 100.00] & 100.00 & 25/32 [56.25, 93.75] & 87.50 & 0 \\
$s_1$ & 18 & 32/32 [100.00, 100.00] & 100.00 & 24/32 [56.25, 93.75] & 68.75 & 0 \\
$s_1$ & 22 & 30/32 [81.25, 100.00] & 93.75 & 23/32 [50.00, 90.62] & 68.75 & 0 \\
$s_1$ & 27 & 32/32 [100.00, 100.00] & 100.00 & 20/32 [37.50, 81.25] & 71.88 & 0 \\
$s_1$ & 31 & 32/32 [100.00, 100.00] & 100.00 & 20/32 [37.50, 81.25] & 43.75 & 0 \\
$s_2$ & Majority & 11/32 [21.88, 43.75] & 31.25 & \scNA & \scNA & 0 \\
$s_2$ & Surface & 24/32 [56.25, 93.75] & 78.12 & \scNA & \scNA & 0 \\
$s_2$ & 0 & 16/32 [50.00, 50.00] & 50.00 & 15/32 [28.12, 65.62] & 40.62 & 0 \\
$s_2$ & 4 & 16/32 [50.00, 50.00] & 50.00 & 22/32 [50.00, 84.38] & 50.00 & 0 \\
$s_2$ & 9 & 15/32 [40.62, 50.00] & 46.88 & 22/32 [50.00, 87.50] & 65.62 & 0 \\
$s_2$ & 13 & 15/32 [40.62, 50.00] & 46.88 & 23/32 [53.12, 90.62] & 65.62 & 0 \\
$s_2$ & 18 & 15/32 [40.62, 50.00] & 46.88 & 21/32 [43.75, 84.38] & 71.88 & 0 \\
$s_2$ & 22 & 15/32 [40.62, 50.00] & 46.88 & 21/32 [43.75, 84.38] & 65.62 & 0 \\
$s_2$ & 27 & 15/32 [40.62, 50.00] & 46.88 & 20/32 [40.62, 81.25] & 50.00 & 0 \\
$s_2$ & 31 & 15/32 [40.62, 50.00] & 46.88 & 17/32 [37.50, 68.75] & 37.50 & 0 \\
A1 amt. & Majority & 16/32 [25.00, 75.00] & 50.00 & \scNA & \scNA & 0 \\
A1 amt. & Surface & 31/32 [90.62, 100.00] & 100.00 & \scNA & \scNA & 0 \\
A1 amt. & 0 & 32/32 [100.00, 100.00] & 93.75 & 30/32 [81.25, 100.00] & 53.12 & 0 \\
A1 amt. & 4 & 32/32 [100.00, 100.00] & 100.00 & 32/32 [100.00, 100.00] & 100.00 & 0 \\
A1 amt. & 9 & 32/32 [100.00, 100.00] & 100.00 & 32/32 [100.00, 100.00] & 100.00 & 0 \\
A1 amt. & 13 & 32/32 [100.00, 100.00] & 100.00 & 32/32 [100.00, 100.00] & 100.00 & 0 \\
A1 amt. & 18 & 32/32 [100.00, 100.00] & 100.00 & 32/32 [100.00, 100.00] & 100.00 & 0 \\
A1 amt. & 22 & 32/32 [100.00, 100.00] & 100.00 & 32/32 [100.00, 100.00] & 100.00 & 0 \\
A1 amt. & 27 & 32/32 [100.00, 100.00] & 100.00 & 32/32 [100.00, 100.00] & 100.00 & 0 \\
A1 amt. & 31 & 32/32 [100.00, 100.00] & 100.00 & 32/32 [100.00, 100.00] & 100.00 & 0 \\
A2 amt. & Majority & 16/32 [50.00, 50.00] & 50.00 & \scNA & \scNA & 0 \\
A2 amt. & Surface & 30/32 [84.38, 100.00] & 96.88 & \scNA & \scNA & 0 \\
A2 amt. & 0 & 16/32 [50.00, 50.00] & 50.00 & 30/32 [84.38, 100.00] & 90.62 & 0 \\
A2 amt. & 4 & 16/32 [50.00, 50.00] & 50.00 & 31/32 [90.62, 100.00] & 96.88 & 0 \\
A2 amt. & 9 & 16/32 [50.00, 50.00] & 50.00 & 32/32 [100.00, 100.00] & 100.00 & 0 \\
A2 amt. & 13 & 16/32 [50.00, 50.00] & 50.00 & 32/32 [100.00, 100.00] & 100.00 & 0 \\
A2 amt. & 18 & 16/32 [50.00, 50.00] & 50.00 & 32/32 [100.00, 100.00] & 100.00 & 0 \\
A2 amt. & 22 & 16/32 [50.00, 50.00] & 50.00 & 32/32 [100.00, 100.00] & 96.88 & 0 \\
A2 amt. & 27 & 16/32 [50.00, 50.00] & 50.00 & 32/32 [100.00, 100.00] & 96.88 & 0 \\
A2 amt. & 31 & 16/32 [50.00, 50.00] & 50.00 & 30/32 [84.38, 100.00] & 96.88 & 0 \\
A1 type & Majority & 16/32 [25.00, 75.00] & 50.00 & \scNA & \scNA & 0 \\
A1 type & Surface & 31/32 [90.62, 100.00] & 100.00 & \scNA & \scNA & 0 \\
A1 type & 0 & 32/32 [100.00, 100.00] & 93.75 & 30/32 [81.25, 100.00] & 53.12 & 0 \\
A1 type & 4 & 32/32 [100.00, 100.00] & 100.00 & 32/32 [100.00, 100.00] & 100.00 & 0 \\
A1 type & 9 & 32/32 [100.00, 100.00] & 100.00 & 32/32 [100.00, 100.00] & 100.00 & 0 \\
A1 type & 13 & 32/32 [100.00, 100.00] & 100.00 & 32/32 [100.00, 100.00] & 100.00 & 0 \\
A1 type & 18 & 32/32 [100.00, 100.00] & 100.00 & 32/32 [100.00, 100.00] & 100.00 & 0 \\
A1 type & 22 & 32/32 [100.00, 100.00] & 100.00 & 32/32 [100.00, 100.00] & 100.00 & 0 \\
A1 type & 27 & 32/32 [100.00, 100.00] & 100.00 & 32/32 [100.00, 100.00] & 100.00 & 0 \\
A1 type & 31 & 32/32 [100.00, 100.00] & 100.00 & 32/32 [100.00, 100.00] & 100.00 & 0 \\
A2 type & Majority & 24/32 [62.50, 87.50] & 75.00 & \scNA & \scNA & 0 \\
A2 type & Surface & 31/32 [90.62, 100.00] & 100.00 & \scNA & \scNA & 0 \\
A2 type & 0 & 24/32 [62.50, 87.50] & 75.00 & 32/32 [100.00, 100.00] & 93.75 & 0 \\
A2 type & 4 & 24/32 [62.50, 87.50] & 75.00 & 32/32 [100.00, 100.00] & 100.00 & 0 \\
A2 type & 9 & 24/32 [62.50, 87.50] & 75.00 & 32/32 [100.00, 100.00] & 100.00 & 0 \\
A2 type & 13 & 24/32 [62.50, 87.50] & 75.00 & 32/32 [100.00, 100.00] & 100.00 & 0 \\
A2 type & 18 & 24/32 [62.50, 87.50] & 75.00 & 32/32 [100.00, 100.00] & 100.00 & 0 \\
A2 type & 22 & 24/32 [62.50, 87.50] & 75.00 & 32/32 [100.00, 100.00] & 100.00 & 0 \\
A2 type & 27 & 24/32 [62.50, 87.50] & 75.00 & 32/32 [100.00, 100.00] & 100.00 & 0 \\
A2 type & 31 & 24/32 [62.50, 87.50] & 75.00 & 32/32 [100.00, 100.00] & 100.00 & 0 \\
Entity & Majority & 4/32 [0.00, 31.25] & 25.00 & \scNA & \scNA & 2 \\
Entity & Surface & 30/32 [81.25, 100.00] & 87.50 & \scNA & \scNA & 2 \\
Entity & 0 & 16/32 [25.00, 75.00] & 56.25 & 30/32 [81.25, 100.00] & 87.50 & 2 \\
Entity & 4 & 20/32 [37.50, 87.50] & 62.50 & 30/32 [81.25, 100.00] & 87.50 & 2 \\
Entity & 9 & 28/32 [68.75, 100.00] & 68.75 & 30/32 [81.25, 100.00] & 87.50 & 2 \\
Entity & 13 & 28/32 [68.75, 100.00] & 50.00 & 30/32 [81.25, 100.00] & 87.50 & 2 \\
Entity & 18 & 24/32 [50.00, 93.75] & 62.50 & 30/32 [81.25, 100.00] & 87.50 & 2 \\
Entity & 22 & 24/32 [50.00, 93.75] & 56.25 & 30/32 [81.25, 100.00] & 84.38 & 2 \\
Entity & 27 & 28/32 [68.75, 100.00] & 81.25 & 30/32 [81.25, 100.00] & 87.50 & 2 \\
Entity & 31 & 30/32 [81.25, 100.00] & 87.50 & 30/32 [81.25, 100.00] & 87.50 & 2 \\
Protected & Majority & 20/32 [37.50, 87.50] & 81.25 & \scNA & \scNA & 0 \\
Protected & Surface & 19/32 [34.38, 81.25] & 75.00 & \scNA & \scNA & 0 \\
Protected & 0 & 22/32 [43.75, 87.50] & 75.00 & 18/32 [31.25, 81.25] & 87.50 & 0 \\
Protected & 4 & 18/32 [31.25, 81.25] & 68.75 & 18/32 [31.25, 81.25] & 68.75 & 0 \\
Protected & 9 & 16/32 [25.00, 75.00] & 50.00 & 14/32 [18.75, 68.75] & 71.88 & 0 \\
Protected & 13 & 18/32 [31.25, 81.25] & 50.00 & 14/32 [18.75, 68.75] & 65.62 & 0 \\
Protected & 18 & 18/32 [31.25, 81.25] & 56.25 & 15/32 [25.00, 71.88] & 65.62 & 0 \\
Protected & 22 & 20/32 [37.50, 87.50] & 56.25 & 12/32 [12.50, 62.50] & 75.00 & 0 \\
Protected & 27 & 16/32 [25.00, 75.00] & 62.50 & 16/32 [28.12, 71.88] & 75.00 & 0 \\
Protected & 31 & 18/32 [31.25, 81.25] & 62.50 & 19/32 [34.38, 81.25] & 81.25 & 0 \\
\end{longtable}
\endgroup

\subsection{Qwen3.5-9B: B2/FINAL}
\label{app:sc_r1_full_9b_b2_final}

Table~\ref{tab:sc_r1_full_9b_b2_final} reports the symbolic final-query panel. The evaluation has only three program-family clusters, and entity/protected labels are not supplied. Readability of action fields that already appear in the prompt is distinct from successful counterfactual state interchange.

% Source: supplied evidence report, lines 4295--5661.
\FloatBarrier
\begingroup
\scriptsize
\setlength{\tabcolsep}{3pt}
\renewcommand{\arraystretch}{1.08}
\begin{longtable}{L{0.105\linewidth}L{0.09\linewidth}L{0.25\linewidth}rL{0.25\linewidth}rr}
\caption{Qwen3.5-9B B2/FINAL complete fixed-layer ledger. Residual rows are indexed by actual block; baseline rows have no block. Correct/N and confidence intervals are retained for each available anchor; selection values are descriptive. Evaluation: 18 sequences/3 program families.}\label{tab:sc_r1_full_9b_b2_final}\\
\toprule
Variable & Reader/block & Checkpoint: count [CI] & Sel. & Query: count [CI] & Sel. & $U$ \\
\midrule
\endfirsthead
\multicolumn{7}{l}{\footnotesize Table~\thetable\ continued.}\\
\toprule
Variable & Reader/block & Checkpoint: count [CI] & Sel. & Query: count [CI] & Sel. & $U$ \\
\midrule
\endhead
\midrule\multicolumn{7}{r}{\footnotesize Continued on the next page.}\\
\endfoot
\bottomrule
\endlastfoot
$Z$ & \multicolumn{6}{l}{Constant training label} \\
Entity & \multicolumn{6}{l}{Label not supplied} \\
Protected & \multicolumn{6}{l}{Label not supplied} \\
$s_0$ & Majority & 0/18 [0.00, 0.00] & 0.00 & \scNA & \scNA & 2 \\
$s_0$ & Surface & 11/18 [16.67, 83.33] & 16.67 & \scNA & \scNA & 2 \\
$s_0$ & 0 & 6/18 [0.00, 66.67] & 16.67 & 4/18 [0.00, 50.00] & 16.67 & 2 \\
$s_0$ & 4 & 12/18 [66.67, 66.67] & 16.67 & 16/18 [83.33, 100.00] & 16.67 & 2 \\
$s_0$ & 9 & 9/18 [16.67, 66.67] & 16.67 & 16/18 [83.33, 100.00] & 16.67 & 2 \\
$s_0$ & 13 & 9/18 [16.67, 66.67] & 16.67 & 15/18 [66.67, 100.00] & 16.67 & 2 \\
$s_0$ & 18 & 9/18 [16.67, 66.67] & 11.11 & 10/18 [50.00, 66.67] & 16.67 & 2 \\
$s_0$ & 22 & 11/18 [50.00, 66.67] & 11.11 & 11/18 [50.00, 83.33] & 16.67 & 2 \\
$s_0$ & 27 & 13/18 [66.67, 83.33] & 16.67 & 9/18 [33.33, 66.67] & 16.67 & 2 \\
$s_0$ & 31 & 13/18 [66.67, 83.33] & 16.67 & 12/18 [50.00, 83.33] & 16.67 & 2 \\
$s_1$ & Majority & 3/18 [16.67, 16.67] & 33.33 & \scNA & \scNA & 0 \\
$s_1$ & Surface & 1/18 [0.00, 16.67] & 16.67 & \scNA & \scNA & 0 \\
$s_1$ & 0 & 3/18 [16.67, 16.67] & 22.22 & 3/18 [16.67, 16.67] & 16.67 & 0 \\
$s_1$ & 4 & 3/18 [16.67, 16.67] & 16.67 & 3/18 [16.67, 16.67] & 22.22 & 0 \\
$s_1$ & 9 & 3/18 [16.67, 16.67] & 16.67 & 3/18 [16.67, 16.67] & 22.22 & 0 \\
$s_1$ & 13 & 3/18 [16.67, 16.67] & 22.22 & 2/18 [0.00, 16.67] & 16.67 & 0 \\
$s_1$ & 18 & 2/18 [0.00, 16.67] & 22.22 & 3/18 [16.67, 16.67] & 16.67 & 0 \\
$s_1$ & 22 & 2/18 [0.00, 16.67] & 27.78 & 2/18 [0.00, 16.67] & 16.67 & 0 \\
$s_1$ & 27 & 3/18 [16.67, 16.67] & 22.22 & 2/18 [0.00, 16.67] & 16.67 & 0 \\
$s_1$ & 31 & 3/18 [16.67, 16.67] & 22.22 & 3/18 [16.67, 16.67] & 16.67 & 0 \\
$s_2$ & Majority & 3/18 [16.67, 16.67] & 22.22 & \scNA & \scNA & 5 \\
$s_2$ & Surface & 3/18 [16.67, 16.67] & 16.67 & \scNA & \scNA & 5 \\
$s_2$ & 0 & 6/18 [33.33, 33.33] & 16.67 & 5/18 [16.67, 33.33] & 16.67 & 5 \\
$s_2$ & 4 & 6/18 [33.33, 33.33] & 27.78 & 6/18 [33.33, 33.33] & 16.67 & 5 \\
$s_2$ & 9 & 5/18 [16.67, 33.33] & 22.22 & 6/18 [33.33, 33.33] & 16.67 & 5 \\
$s_2$ & 13 & 6/18 [33.33, 33.33] & 22.22 & 6/18 [33.33, 33.33] & 16.67 & 5 \\
$s_2$ & 18 & 5/18 [16.67, 33.33] & 16.67 & 5/18 [16.67, 33.33] & 16.67 & 5 \\
$s_2$ & 22 & 4/18 [16.67, 33.33] & 16.67 & 5/18 [16.67, 33.33] & 11.11 & 5 \\
$s_2$ & 27 & 4/18 [0.00, 33.33] & 16.67 & 5/18 [16.67, 33.33] & 11.11 & 5 \\
$s_2$ & 31 & 5/18 [16.67, 33.33] & 16.67 & 6/18 [33.33, 33.33] & 16.67 & 5 \\
A1 amt. & Majority & 9/18 [50.00, 50.00] & 50.00 & \scNA & \scNA & 0 \\
A1 amt. & Surface & 16/18 [66.67, 100.00] & 100.00 & \scNA & \scNA & 0 \\
A1 amt. & 0 & 18/18 [100.00, 100.00] & 100.00 & 18/18 [100.00, 100.00] & 100.00 & 0 \\
A1 amt. & 4 & 18/18 [100.00, 100.00] & 100.00 & 18/18 [100.00, 100.00] & 100.00 & 0 \\
A1 amt. & 9 & 18/18 [100.00, 100.00] & 100.00 & 18/18 [100.00, 100.00] & 100.00 & 0 \\
A1 amt. & 13 & 18/18 [100.00, 100.00] & 100.00 & 18/18 [100.00, 100.00] & 100.00 & 0 \\
A1 amt. & 18 & 18/18 [100.00, 100.00] & 100.00 & 18/18 [100.00, 100.00] & 100.00 & 0 \\
A1 amt. & 22 & 18/18 [100.00, 100.00] & 100.00 & 18/18 [100.00, 100.00] & 100.00 & 0 \\
A1 amt. & 27 & 18/18 [100.00, 100.00] & 100.00 & 18/18 [100.00, 100.00] & 100.00 & 0 \\
A1 amt. & 31 & 18/18 [100.00, 100.00] & 100.00 & 18/18 [100.00, 100.00] & 100.00 & 0 \\
A2 amt. & Majority & 6/18 [33.33, 33.33] & 33.33 & \scNA & \scNA & 0 \\
A2 amt. & Surface & 18/18 [100.00, 100.00] & 100.00 & \scNA & \scNA & 0 \\
A2 amt. & 0 & 15/18 [83.33, 83.33] & 83.33 & 18/18 [100.00, 100.00] & 100.00 & 0 \\
A2 amt. & 4 & 15/18 [83.33, 83.33] & 83.33 & 18/18 [100.00, 100.00] & 100.00 & 0 \\
A2 amt. & 9 & 15/18 [83.33, 83.33] & 83.33 & 18/18 [100.00, 100.00] & 100.00 & 0 \\
A2 amt. & 13 & 15/18 [83.33, 83.33] & 83.33 & 18/18 [100.00, 100.00] & 100.00 & 0 \\
A2 amt. & 18 & 15/18 [83.33, 83.33] & 83.33 & 18/18 [100.00, 100.00] & 100.00 & 0 \\
A2 amt. & 22 & 15/18 [83.33, 83.33] & 83.33 & 18/18 [100.00, 100.00] & 100.00 & 0 \\
A2 amt. & 27 & 15/18 [83.33, 83.33] & 83.33 & 18/18 [100.00, 100.00] & 100.00 & 0 \\
A2 amt. & 31 & 15/18 [83.33, 83.33] & 83.33 & 18/18 [100.00, 100.00] & 100.00 & 0 \\
A1 type & Majority & 15/18 [83.33, 83.33] & 83.33 & \scNA & \scNA & 0 \\
A1 type & Surface & 18/18 [100.00, 100.00] & 100.00 & \scNA & \scNA & 0 \\
A1 type & 0 & 18/18 [100.00, 100.00] & 100.00 & 18/18 [100.00, 100.00] & 100.00 & 0 \\
A1 type & 4 & 18/18 [100.00, 100.00] & 100.00 & 18/18 [100.00, 100.00] & 100.00 & 0 \\
A1 type & 9 & 18/18 [100.00, 100.00] & 100.00 & 18/18 [100.00, 100.00] & 100.00 & 0 \\
A1 type & 13 & 18/18 [100.00, 100.00] & 100.00 & 18/18 [100.00, 100.00] & 100.00 & 0 \\
A1 type & 18 & 18/18 [100.00, 100.00] & 100.00 & 18/18 [100.00, 100.00] & 100.00 & 0 \\
A1 type & 22 & 18/18 [100.00, 100.00] & 100.00 & 18/18 [100.00, 100.00] & 100.00 & 0 \\
A1 type & 27 & 18/18 [100.00, 100.00] & 100.00 & 18/18 [100.00, 100.00] & 100.00 & 0 \\
A1 type & 31 & 18/18 [100.00, 100.00] & 100.00 & 18/18 [100.00, 100.00] & 100.00 & 0 \\
A2 type & Majority & 12/18 [66.67, 66.67] & 66.67 & \scNA & \scNA & 0 \\
A2 type & Surface & 18/18 [100.00, 100.00] & 100.00 & \scNA & \scNA & 0 \\
A2 type & 0 & 15/18 [83.33, 83.33] & 83.33 & 18/18 [100.00, 100.00] & 100.00 & 0 \\
A2 type & 4 & 15/18 [83.33, 83.33] & 83.33 & 18/18 [100.00, 100.00] & 100.00 & 0 \\
A2 type & 9 & 15/18 [83.33, 83.33] & 83.33 & 18/18 [100.00, 100.00] & 100.00 & 0 \\
A2 type & 13 & 15/18 [83.33, 83.33] & 83.33 & 18/18 [100.00, 100.00] & 100.00 & 0 \\
A2 type & 18 & 15/18 [83.33, 83.33] & 83.33 & 18/18 [100.00, 100.00] & 100.00 & 0 \\
A2 type & 22 & 15/18 [83.33, 83.33] & 83.33 & 18/18 [100.00, 100.00] & 100.00 & 0 \\
A2 type & 27 & 15/18 [83.33, 83.33] & 83.33 & 18/18 [100.00, 100.00] & 100.00 & 0 \\
A2 type & 31 & 15/18 [83.33, 83.33] & 83.33 & 18/18 [100.00, 100.00] & 100.00 & 0 \\
\end{longtable}
\endgroup

\subsection{Qwen3.5-27B: B1/B01}
\label{app:sc_r1_full_27b_b1_b01}

Table~\ref{tab:sc_r1_full_27b_b1_b01} includes initial and final state variables, both actions, entity identity, and the protected fact for the original full visual context. The complete table prevents the prespecified S1 summary from being read as a claim that every variable is equally unreadable. Class-coverage limitations and the constant target label are retained.

% Source: supplied evidence report, lines 4295--5661.
\FloatBarrier
\begingroup
\scriptsize
\setlength{\tabcolsep}{3pt}
\renewcommand{\arraystretch}{1.08}
\begin{longtable}{L{0.105\linewidth}L{0.09\linewidth}L{0.25\linewidth}rL{0.25\linewidth}rr}
\caption{Qwen3.5-27B B1/B01 complete fixed-layer ledger. Residual rows are indexed by actual block; baseline rows have no block. Correct/N and confidence intervals are retained for each available anchor; selection values are descriptive. Evaluation: 32 sequences/16 worlds.}\label{tab:sc_r1_full_27b_b1_b01}\\
\toprule
Variable & Reader/block & Checkpoint: count [CI] & Sel. & Query: count [CI] & Sel. & $U$ \\
\midrule
\endfirsthead
\multicolumn{7}{l}{\footnotesize Table~\thetable\ continued.}\\
\toprule
Variable & Reader/block & Checkpoint: count [CI] & Sel. & Query: count [CI] & Sel. & $U$ \\
\midrule
\endhead
\midrule\multicolumn{7}{r}{\footnotesize Continued on the next page.}\\
\endfoot
\bottomrule
\endlastfoot
$Z$ & \multicolumn{6}{l}{Constant training label} \\
$s_0$ & Majority & 22/32 [43.75, 87.50] & 62.50 & \scNA & \scNA & 2 \\
$s_0$ & Surface & 21/32 [43.75, 87.50] & 40.62 & \scNA & \scNA & 2 \\
$s_0$ & 0 & 16/32 [25.00, 75.00] & 25.00 & 20/32 [37.50, 87.50] & 34.38 & 2 \\
$s_0$ & 9 & 18/32 [31.25, 81.25] & 43.75 & 20/32 [37.50, 87.50] & 37.50 & 2 \\
$s_0$ & 18 & 22/32 [43.75, 87.50] & 37.50 & 22/32 [43.75, 87.50] & 31.25 & 2 \\
$s_0$ & 27 & 22/32 [43.75, 87.50] & 37.50 & 22/32 [43.75, 87.50] & 40.62 & 2 \\
$s_0$ & 36 & 20/32 [37.50, 87.50] & 50.00 & 20/32 [37.50, 87.50] & 53.12 & 2 \\
$s_0$ & 45 & 22/32 [43.75, 87.50] & 50.00 & 22/32 [43.75, 87.50] & 50.00 & 2 \\
$s_0$ & 54 & 18/32 [31.25, 81.25] & 50.00 & 22/32 [43.75, 87.50] & 59.38 & 2 \\
$s_0$ & 63 & 22/32 [43.75, 87.50] & 43.75 & 22/32 [43.75, 87.50] & 34.38 & 2 \\
$s_1$ & Majority & 10/32 [12.50, 56.25] & 43.75 & \scNA & \scNA & 0 \\
$s_1$ & Surface & 22/32 [46.88, 87.50] & 65.62 & \scNA & \scNA & 0 \\
$s_1$ & 0 & 14/32 [18.75, 68.75] & 37.50 & 16/32 [28.12, 71.88] & 25.00 & 0 \\
$s_1$ & 9 & 14/32 [18.75, 68.75] & 37.50 & 16/32 [25.00, 75.00] & 34.38 & 0 \\
$s_1$ & 18 & 18/32 [31.25, 81.25] & 37.50 & 18/32 [31.25, 81.25] & 40.62 & 0 \\
$s_1$ & 27 & 18/32 [31.25, 81.25] & 50.00 & 17/32 [31.25, 75.00] & 40.62 & 0 \\
$s_1$ & 36 & 18/32 [31.25, 81.25] & 50.00 & 20/32 [37.50, 87.50] & 56.25 & 0 \\
$s_1$ & 45 & 16/32 [25.00, 75.00] & 56.25 & 23/32 [50.00, 90.62] & 56.25 & 0 \\
$s_1$ & 54 & 18/32 [31.25, 81.25] & 56.25 & 20/32 [37.50, 87.50] & 46.88 & 0 \\
$s_1$ & 63 & 18/32 [31.25, 81.25] & 62.50 & 20/32 [37.50, 81.25] & 62.50 & 0 \\
$s_2$ & Majority & 11/32 [21.88, 43.75] & 31.25 & \scNA & \scNA & 0 \\
$s_2$ & Surface & 23/32 [50.00, 90.62] & 56.25 & \scNA & \scNA & 0 \\
$s_2$ & 0 & 9/32 [15.62, 40.62] & 28.12 & 15/32 [28.12, 65.62] & 28.12 & 0 \\
$s_2$ & 9 & 10/32 [18.75, 43.75] & 28.12 & 18/32 [37.50, 75.00] & 53.12 & 0 \\
$s_2$ & 18 & 11/32 [21.88, 43.75] & 34.38 & 17/32 [31.25, 75.00] & 53.12 & 0 \\
$s_2$ & 27 & 10/32 [18.75, 43.75] & 28.12 & 18/32 [34.38, 78.12] & 43.75 & 0 \\
$s_2$ & 36 & 10/32 [18.75, 40.62] & 28.12 & 17/32 [31.25, 75.00] & 50.00 & 0 \\
$s_2$ & 45 & 10/32 [18.75, 43.75] & 28.12 & 19/32 [40.62, 78.12] & 50.00 & 0 \\
$s_2$ & 54 & 11/32 [21.88, 43.75] & 31.25 & 18/32 [34.38, 75.00] & 50.00 & 0 \\
$s_2$ & 63 & 10/32 [18.75, 43.75] & 25.00 & 20/32 [43.75, 81.25] & 40.62 & 0 \\
A1 amt. & Majority & 16/32 [25.00, 75.00] & 50.00 & \scNA & \scNA & 0 \\
A1 amt. & Surface & 32/32 [100.00, 100.00] & 100.00 & \scNA & \scNA & 0 \\
A1 amt. & 0 & 32/32 [100.00, 100.00] & 100.00 & 28/32 [68.75, 100.00] & 90.62 & 0 \\
A1 amt. & 9 & 32/32 [100.00, 100.00] & 100.00 & 32/32 [100.00, 100.00] & 100.00 & 0 \\
A1 amt. & 18 & 32/32 [100.00, 100.00] & 100.00 & 32/32 [100.00, 100.00] & 100.00 & 0 \\
A1 amt. & 27 & 32/32 [100.00, 100.00] & 100.00 & 32/32 [100.00, 100.00] & 100.00 & 0 \\
A1 amt. & 36 & 32/32 [100.00, 100.00] & 100.00 & 32/32 [100.00, 100.00] & 100.00 & 0 \\
A1 amt. & 45 & 32/32 [100.00, 100.00] & 100.00 & 32/32 [100.00, 100.00] & 100.00 & 0 \\
A1 amt. & 54 & 32/32 [100.00, 100.00] & 100.00 & 32/32 [100.00, 100.00] & 100.00 & 0 \\
A1 amt. & 63 & 32/32 [100.00, 100.00] & 100.00 & 32/32 [100.00, 100.00] & 100.00 & 0 \\
A2 amt. & Majority & 16/32 [50.00, 50.00] & 50.00 & \scNA & \scNA & 0 \\
A2 amt. & Surface & 32/32 [100.00, 100.00] & 100.00 & \scNA & \scNA & 0 \\
A2 amt. & 0 & 16/32 [50.00, 50.00] & 50.00 & 31/32 [90.62, 100.00] & 96.88 & 0 \\
A2 amt. & 9 & 16/32 [50.00, 50.00] & 50.00 & 32/32 [100.00, 100.00] & 100.00 & 0 \\
A2 amt. & 18 & 16/32 [50.00, 50.00] & 50.00 & 32/32 [100.00, 100.00] & 100.00 & 0 \\
A2 amt. & 27 & 16/32 [50.00, 50.00] & 50.00 & 32/32 [100.00, 100.00] & 100.00 & 0 \\
A2 amt. & 36 & 16/32 [50.00, 50.00] & 50.00 & 32/32 [100.00, 100.00] & 100.00 & 0 \\
A2 amt. & 45 & 16/32 [50.00, 50.00] & 50.00 & 32/32 [100.00, 100.00] & 100.00 & 0 \\
A2 amt. & 54 & 16/32 [50.00, 50.00] & 50.00 & 32/32 [100.00, 100.00] & 100.00 & 0 \\
A2 amt. & 63 & 16/32 [50.00, 50.00] & 50.00 & 32/32 [100.00, 100.00] & 100.00 & 0 \\
A1 type & Majority & 16/32 [25.00, 75.00] & 50.00 & \scNA & \scNA & 0 \\
A1 type & Surface & 32/32 [100.00, 100.00] & 100.00 & \scNA & \scNA & 0 \\
A1 type & 0 & 32/32 [100.00, 100.00] & 100.00 & 28/32 [68.75, 100.00] & 90.62 & 0 \\
A1 type & 9 & 32/32 [100.00, 100.00] & 100.00 & 32/32 [100.00, 100.00] & 100.00 & 0 \\
A1 type & 18 & 32/32 [100.00, 100.00] & 100.00 & 32/32 [100.00, 100.00] & 100.00 & 0 \\
A1 type & 27 & 32/32 [100.00, 100.00] & 100.00 & 32/32 [100.00, 100.00] & 100.00 & 0 \\
A1 type & 36 & 32/32 [100.00, 100.00] & 100.00 & 32/32 [100.00, 100.00] & 100.00 & 0 \\
A1 type & 45 & 32/32 [100.00, 100.00] & 100.00 & 32/32 [100.00, 100.00] & 100.00 & 0 \\
A1 type & 54 & 32/32 [100.00, 100.00] & 100.00 & 32/32 [100.00, 100.00] & 100.00 & 0 \\
A1 type & 63 & 32/32 [100.00, 100.00] & 100.00 & 32/32 [100.00, 100.00] & 100.00 & 0 \\
A2 type & Majority & 24/32 [62.50, 87.50] & 75.00 & \scNA & \scNA & 0 \\
A2 type & Surface & 32/32 [100.00, 100.00] & 100.00 & \scNA & \scNA & 0 \\
A2 type & 0 & 24/32 [62.50, 87.50] & 75.00 & 32/32 [100.00, 100.00] & 100.00 & 0 \\
A2 type & 9 & 24/32 [62.50, 87.50] & 75.00 & 32/32 [100.00, 100.00] & 100.00 & 0 \\
A2 type & 18 & 24/32 [62.50, 87.50] & 75.00 & 32/32 [100.00, 100.00] & 100.00 & 0 \\
A2 type & 27 & 24/32 [62.50, 87.50] & 75.00 & 32/32 [100.00, 100.00] & 100.00 & 0 \\
A2 type & 36 & 24/32 [62.50, 87.50] & 75.00 & 32/32 [100.00, 100.00] & 100.00 & 0 \\
A2 type & 45 & 24/32 [62.50, 87.50] & 75.00 & 32/32 [100.00, 100.00] & 100.00 & 0 \\
A2 type & 54 & 24/32 [62.50, 87.50] & 75.00 & 32/32 [100.00, 100.00] & 100.00 & 0 \\
A2 type & 63 & 24/32 [62.50, 87.50] & 75.00 & 32/32 [100.00, 100.00] & 100.00 & 0 \\
Entity & Majority & 4/32 [0.00, 31.25] & 25.00 & \scNA & \scNA & 2 \\
Entity & Surface & 30/32 [81.25, 100.00] & 87.50 & \scNA & \scNA & 2 \\
Entity & 0 & 28/32 [68.75, 100.00] & 81.25 & 30/32 [81.25, 100.00] & 87.50 & 2 \\
Entity & 9 & 30/32 [81.25, 100.00] & 87.50 & 30/32 [81.25, 100.00] & 87.50 & 2 \\
Entity & 18 & 30/32 [81.25, 100.00] & 75.00 & 30/32 [81.25, 100.00] & 87.50 & 2 \\
Entity & 27 & 30/32 [81.25, 100.00] & 75.00 & 30/32 [81.25, 100.00] & 87.50 & 2 \\
Entity & 36 & 30/32 [81.25, 100.00] & 81.25 & 30/32 [81.25, 100.00] & 81.25 & 2 \\
Entity & 45 & 30/32 [81.25, 100.00] & 81.25 & 30/32 [81.25, 100.00] & 81.25 & 2 \\
Entity & 54 & 26/32 [62.50, 100.00] & 75.00 & 30/32 [81.25, 100.00] & 81.25 & 2 \\
Entity & 63 & 28/32 [68.75, 100.00] & 68.75 & 30/32 [81.25, 100.00] & 87.50 & 2 \\
Protected & Majority & 20/32 [37.50, 87.50] & 81.25 & \scNA & \scNA & 0 \\
Protected & Surface & 16/32 [25.00, 75.00] & 78.12 & \scNA & \scNA & 0 \\
Protected & 0 & 14/32 [18.75, 68.75] & 62.50 & 10/32 [12.50, 56.25] & 62.50 & 0 \\
Protected & 9 & 12/32 [12.50, 62.50] & 75.00 & 12/32 [12.50, 62.50] & 84.38 & 0 \\
Protected & 18 & 14/32 [18.75, 68.75] & 75.00 & 15/32 [25.00, 71.88] & 75.00 & 0 \\
Protected & 27 & 14/32 [18.75, 68.75] & 75.00 & 16/32 [25.00, 75.00] & 75.00 & 0 \\
Protected & 36 & 12/32 [12.50, 62.50] & 68.75 & 14/32 [18.75, 68.75] & 62.50 & 0 \\
Protected & 45 & 8/32 [6.25, 43.75] & 75.00 & 17/32 [28.12, 75.00] & 59.38 & 0 \\
Protected & 54 & 14/32 [18.75, 68.75] & 81.25 & 20/32 [40.62, 84.38] & 59.38 & 0 \\
Protected & 63 & 18/32 [31.25, 81.25] & 75.00 & 14/32 [18.75, 68.75] & 78.12 & 0 \\
\end{longtable}
\endgroup

\subsection{Qwen3.5-27B: B1/B06}
\label{app:sc_r1_full_27b_b1_b06}

Table~\ref{tab:sc_r1_full_27b_b1_b06} uses the explicit intermediate-checkpoint context. High checkpoint readability can reflect a value that is directly supplied in the input; it is not independent evidence that the unassisted model constructs or causally uses that state. All query-anchor results remain visible as a contrasting readout location.

% Source: supplied evidence report, lines 4295--5661.
\FloatBarrier
\begingroup
\scriptsize
\setlength{\tabcolsep}{3pt}
\renewcommand{\arraystretch}{1.08}
\begin{longtable}{L{0.105\linewidth}L{0.09\linewidth}L{0.25\linewidth}rL{0.25\linewidth}rr}
\caption{Qwen3.5-27B B1/B06 complete fixed-layer ledger. Residual rows are indexed by actual block; baseline rows have no block. Correct/N and confidence intervals are retained for each available anchor; selection values are descriptive. Evaluation: 32 sequences/16 worlds.}\label{tab:sc_r1_full_27b_b1_b06}\\
\toprule
Variable & Reader/block & Checkpoint: count [CI] & Sel. & Query: count [CI] & Sel. & $U$ \\
\midrule
\endfirsthead
\multicolumn{7}{l}{\footnotesize Table~\thetable\ continued.}\\
\toprule
Variable & Reader/block & Checkpoint: count [CI] & Sel. & Query: count [CI] & Sel. & $U$ \\
\midrule
\endhead
\midrule\multicolumn{7}{r}{\footnotesize Continued on the next page.}\\
\endfoot
\bottomrule
\endlastfoot
$Z$ & \multicolumn{6}{l}{Constant training label} \\
$s_0$ & Majority & 22/32 [43.75, 87.50] & 62.50 & \scNA & \scNA & 2 \\
$s_0$ & Surface & 21/32 [43.75, 87.50] & 71.88 & \scNA & \scNA & 2 \\
$s_0$ & 0 & 26/32 [62.50, 100.00] & 68.75 & 18/32 [31.25, 81.25] & 34.38 & 2 \\
$s_0$ & 9 & 30/32 [81.25, 100.00] & 93.75 & 20/32 [37.50, 87.50] & 46.88 & 2 \\
$s_0$ & 18 & 30/32 [81.25, 100.00] & 93.75 & 22/32 [43.75, 87.50] & 65.62 & 2 \\
$s_0$ & 27 & 30/32 [81.25, 100.00] & 93.75 & 23/32 [50.00, 90.62] & 71.88 & 2 \\
$s_0$ & 36 & 30/32 [81.25, 100.00] & 93.75 & 25/32 [56.25, 96.88] & 81.25 & 2 \\
$s_0$ & 45 & 30/32 [81.25, 100.00] & 93.75 & 21/32 [43.75, 87.50] & 81.25 & 2 \\
$s_0$ & 54 & 30/32 [81.25, 100.00] & 93.75 & 24/32 [56.25, 93.75] & 81.25 & 2 \\
$s_0$ & 63 & 30/32 [81.25, 100.00] & 93.75 & 24/32 [50.00, 93.75] & 65.62 & 2 \\
$s_1$ & Majority & 10/32 [12.50, 56.25] & 43.75 & \scNA & \scNA & 0 \\
$s_1$ & Surface & 26/32 [62.50, 96.88] & 96.88 & \scNA & \scNA & 0 \\
$s_1$ & 0 & 32/32 [100.00, 100.00] & 100.00 & 18/32 [31.25, 81.25] & 56.25 & 0 \\
$s_1$ & 9 & 32/32 [100.00, 100.00] & 100.00 & 16/32 [28.12, 71.88] & 59.38 & 0 \\
$s_1$ & 18 & 32/32 [100.00, 100.00] & 100.00 & 24/32 [56.25, 93.75] & 71.88 & 0 \\
$s_1$ & 27 & 32/32 [100.00, 100.00] & 100.00 & 26/32 [62.50, 96.88] & 78.12 & 0 \\
$s_1$ & 36 & 30/32 [81.25, 100.00] & 100.00 & 25/32 [56.25, 93.75] & 87.50 & 0 \\
$s_1$ & 45 & 30/32 [81.25, 100.00] & 100.00 & 24/32 [50.00, 93.75] & 81.25 & 0 \\
$s_1$ & 54 & 32/32 [100.00, 100.00] & 100.00 & 25/32 [56.25, 93.75] & 75.00 & 0 \\
$s_1$ & 63 & 32/32 [100.00, 100.00] & 100.00 & 20/32 [37.50, 87.50] & 81.25 & 0 \\
$s_2$ & Majority & 11/32 [21.88, 43.75] & 31.25 & \scNA & \scNA & 0 \\
$s_2$ & Surface & 24/32 [56.25, 93.75] & 78.12 & \scNA & \scNA & 0 \\
$s_2$ & 0 & 16/32 [50.00, 50.00] & 50.00 & 16/32 [34.38, 65.62] & 65.62 & 0 \\
$s_2$ & 9 & 16/32 [50.00, 50.00] & 50.00 & 19/32 [37.50, 81.25] & 46.88 & 0 \\
$s_2$ & 18 & 15/32 [40.62, 50.00] & 50.00 & 19/32 [40.62, 78.12] & 53.12 & 0 \\
$s_2$ & 27 & 15/32 [40.62, 50.00] & 50.00 & 24/32 [56.25, 90.62] & 65.62 & 0 \\
$s_2$ & 36 & 15/32 [40.62, 50.00] & 46.88 & 25/32 [56.25, 93.75] & 71.88 & 0 \\
$s_2$ & 45 & 15/32 [40.62, 50.00] & 46.88 & 25/32 [59.38, 93.75] & 68.75 & 0 \\
$s_2$ & 54 & 15/32 [40.62, 50.00] & 46.88 & 26/32 [62.50, 96.88] & 68.75 & 0 \\
$s_2$ & 63 & 16/32 [50.00, 50.00] & 50.00 & 24/32 [56.25, 90.62] & 65.62 & 0 \\
A1 amt. & Majority & 16/32 [25.00, 75.00] & 50.00 & \scNA & \scNA & 0 \\
A1 amt. & Surface & 31/32 [90.62, 100.00] & 100.00 & \scNA & \scNA & 0 \\
A1 amt. & 0 & 32/32 [100.00, 100.00] & 100.00 & 28/32 [68.75, 100.00] & 84.38 & 0 \\
A1 amt. & 9 & 32/32 [100.00, 100.00] & 100.00 & 32/32 [100.00, 100.00] & 100.00 & 0 \\
A1 amt. & 18 & 32/32 [100.00, 100.00] & 100.00 & 32/32 [100.00, 100.00] & 100.00 & 0 \\
A1 amt. & 27 & 32/32 [100.00, 100.00] & 100.00 & 32/32 [100.00, 100.00] & 100.00 & 0 \\
A1 amt. & 36 & 32/32 [100.00, 100.00] & 100.00 & 32/32 [100.00, 100.00] & 100.00 & 0 \\
A1 amt. & 45 & 32/32 [100.00, 100.00] & 100.00 & 32/32 [100.00, 100.00] & 100.00 & 0 \\
A1 amt. & 54 & 32/32 [100.00, 100.00] & 100.00 & 32/32 [100.00, 100.00] & 100.00 & 0 \\
A1 amt. & 63 & 32/32 [100.00, 100.00] & 100.00 & 32/32 [100.00, 100.00] & 100.00 & 0 \\
A2 amt. & Majority & 16/32 [50.00, 50.00] & 50.00 & \scNA & \scNA & 0 \\
A2 amt. & Surface & 30/32 [84.38, 100.00] & 96.88 & \scNA & \scNA & 0 \\
A2 amt. & 0 & 16/32 [50.00, 50.00] & 50.00 & 31/32 [90.62, 100.00] & 96.88 & 0 \\
A2 amt. & 9 & 16/32 [50.00, 50.00] & 50.00 & 32/32 [100.00, 100.00] & 100.00 & 0 \\
A2 amt. & 18 & 16/32 [50.00, 50.00] & 50.00 & 32/32 [100.00, 100.00] & 100.00 & 0 \\
A2 amt. & 27 & 16/32 [50.00, 50.00] & 50.00 & 32/32 [100.00, 100.00] & 100.00 & 0 \\
A2 amt. & 36 & 16/32 [50.00, 50.00] & 50.00 & 32/32 [100.00, 100.00] & 100.00 & 0 \\
A2 amt. & 45 & 16/32 [50.00, 50.00] & 50.00 & 32/32 [100.00, 100.00] & 100.00 & 0 \\
A2 amt. & 54 & 16/32 [50.00, 50.00] & 50.00 & 32/32 [100.00, 100.00] & 100.00 & 0 \\
A2 amt. & 63 & 16/32 [50.00, 50.00] & 50.00 & 32/32 [100.00, 100.00] & 96.88 & 0 \\
A1 type & Majority & 16/32 [25.00, 75.00] & 50.00 & \scNA & \scNA & 0 \\
A1 type & Surface & 31/32 [90.62, 100.00] & 100.00 & \scNA & \scNA & 0 \\
A1 type & 0 & 32/32 [100.00, 100.00] & 100.00 & 28/32 [68.75, 100.00] & 84.38 & 0 \\
A1 type & 9 & 32/32 [100.00, 100.00] & 100.00 & 32/32 [100.00, 100.00] & 100.00 & 0 \\
A1 type & 18 & 32/32 [100.00, 100.00] & 100.00 & 32/32 [100.00, 100.00] & 100.00 & 0 \\
A1 type & 27 & 32/32 [100.00, 100.00] & 100.00 & 32/32 [100.00, 100.00] & 100.00 & 0 \\
A1 type & 36 & 32/32 [100.00, 100.00] & 100.00 & 32/32 [100.00, 100.00] & 100.00 & 0 \\
A1 type & 45 & 32/32 [100.00, 100.00] & 100.00 & 32/32 [100.00, 100.00] & 100.00 & 0 \\
A1 type & 54 & 32/32 [100.00, 100.00] & 100.00 & 32/32 [100.00, 100.00] & 100.00 & 0 \\
A1 type & 63 & 32/32 [100.00, 100.00] & 100.00 & 32/32 [100.00, 100.00] & 100.00 & 0 \\
A2 type & Majority & 24/32 [62.50, 87.50] & 75.00 & \scNA & \scNA & 0 \\
A2 type & Surface & 31/32 [90.62, 100.00] & 100.00 & \scNA & \scNA & 0 \\
A2 type & 0 & 24/32 [62.50, 87.50] & 75.00 & 32/32 [100.00, 100.00] & 100.00 & 0 \\
A2 type & 9 & 24/32 [62.50, 87.50] & 75.00 & 32/32 [100.00, 100.00] & 100.00 & 0 \\
A2 type & 18 & 24/32 [62.50, 87.50] & 75.00 & 32/32 [100.00, 100.00] & 100.00 & 0 \\
A2 type & 27 & 24/32 [62.50, 87.50] & 75.00 & 32/32 [100.00, 100.00] & 100.00 & 0 \\
A2 type & 36 & 24/32 [62.50, 87.50] & 75.00 & 32/32 [100.00, 100.00] & 100.00 & 0 \\
A2 type & 45 & 24/32 [62.50, 87.50] & 75.00 & 32/32 [100.00, 100.00] & 100.00 & 0 \\
A2 type & 54 & 24/32 [62.50, 87.50] & 75.00 & 32/32 [100.00, 100.00] & 100.00 & 0 \\
A2 type & 63 & 24/32 [62.50, 87.50] & 75.00 & 32/32 [100.00, 100.00] & 100.00 & 0 \\
Entity & Majority & 4/32 [0.00, 31.25] & 25.00 & \scNA & \scNA & 2 \\
Entity & Surface & 30/32 [81.25, 100.00] & 87.50 & \scNA & \scNA & 2 \\
Entity & 0 & 26/32 [62.50, 100.00] & 81.25 & 30/32 [81.25, 100.00] & 87.50 & 2 \\
Entity & 9 & 30/32 [81.25, 100.00] & 81.25 & 30/32 [81.25, 100.00] & 87.50 & 2 \\
Entity & 18 & 30/32 [81.25, 100.00] & 75.00 & 30/32 [81.25, 100.00] & 87.50 & 2 \\
Entity & 27 & 30/32 [81.25, 100.00] & 75.00 & 30/32 [81.25, 100.00] & 87.50 & 2 \\
Entity & 36 & 22/32 [43.75, 87.50] & 50.00 & 30/32 [81.25, 100.00] & 81.25 & 2 \\
Entity & 45 & 22/32 [43.75, 87.50] & 37.50 & 30/32 [81.25, 100.00] & 78.12 & 2 \\
Entity & 54 & 20/32 [37.50, 87.50] & 56.25 & 30/32 [81.25, 100.00] & 81.25 & 2 \\
Entity & 63 & 30/32 [81.25, 100.00] & 87.50 & 30/32 [81.25, 100.00] & 87.50 & 2 \\
Protected & Majority & 20/32 [37.50, 87.50] & 81.25 & \scNA & \scNA & 0 \\
Protected & Surface & 19/32 [34.38, 81.25] & 75.00 & \scNA & \scNA & 0 \\
Protected & 0 & 14/32 [18.75, 68.75] & 56.25 & 10/32 [12.50, 56.25] & 62.50 & 0 \\
Protected & 9 & 18/32 [31.25, 81.25] & 68.75 & 13/32 [18.75, 62.50] & 84.38 & 0 \\
Protected & 18 & 16/32 [25.00, 75.00] & 62.50 & 16/32 [25.00, 75.00] & 75.00 & 0 \\
Protected & 27 & 14/32 [18.75, 68.75] & 62.50 & 16/32 [25.00, 75.00] & 65.62 & 0 \\
Protected & 36 & 20/32 [37.50, 87.50] & 68.75 & 14/32 [18.75, 68.75] & 62.50 & 0 \\
Protected & 45 & 18/32 [31.25, 81.25] & 75.00 & 17/32 [28.12, 78.12] & 56.25 & 0 \\
Protected & 54 & 18/32 [31.25, 81.25] & 62.50 & 15/32 [25.00, 71.88] & 65.62 & 0 \\
Protected & 63 & 20/32 [37.50, 87.50] & 75.00 & 16/32 [25.00, 75.00] & 78.12 & 0 \\
\end{longtable}
\endgroup

\subsection{Qwen3.5-27B: B2/FINAL}
\label{app:sc_r1_full_27b_b2_final}

Table~\ref{tab:sc_r1_full_27b_b2_final} reports the symbolic final-query panel. The evaluation has only three program-family clusters, and entity/protected labels are not supplied. Readability of action fields that already appear in the prompt is distinct from successful counterfactual state interchange.

% Source: supplied evidence report, lines 4295--5661.
\FloatBarrier
\begingroup
\scriptsize
\setlength{\tabcolsep}{3pt}
\renewcommand{\arraystretch}{1.08}
\begin{longtable}{L{0.105\linewidth}L{0.09\linewidth}L{0.25\linewidth}rL{0.25\linewidth}rr}
\caption{Qwen3.5-27B B2/FINAL complete fixed-layer ledger. Residual rows are indexed by actual block; baseline rows have no block. Correct/N and confidence intervals are retained for each available anchor; selection values are descriptive. Evaluation: 18 sequences/3 program families.}\label{tab:sc_r1_full_27b_b2_final}\\
\toprule
Variable & Reader/block & Checkpoint: count [CI] & Sel. & Query: count [CI] & Sel. & $U$ \\
\midrule
\endfirsthead
\multicolumn{7}{l}{\footnotesize Table~\thetable\ continued.}\\
\toprule
Variable & Reader/block & Checkpoint: count [CI] & Sel. & Query: count [CI] & Sel. & $U$ \\
\midrule
\endhead
\midrule\multicolumn{7}{r}{\footnotesize Continued on the next page.}\\
\endfoot
\bottomrule
\endlastfoot
$Z$ & \multicolumn{6}{l}{Constant training label} \\
Entity & \multicolumn{6}{l}{Label not supplied} \\
Protected & \multicolumn{6}{l}{Label not supplied} \\
$s_0$ & Majority & 0/18 [0.00, 0.00] & 0.00 & \scNA & \scNA & 2 \\
$s_0$ & Surface & 11/18 [16.67, 83.33] & 16.67 & \scNA & \scNA & 2 \\
$s_0$ & 0 & 14/18 [66.67, 83.33] & 16.67 & 6/18 [0.00, 83.33] & 16.67 & 2 \\
$s_0$ & 9 & 12/18 [66.67, 66.67] & 16.67 & 15/18 [83.33, 83.33] & 16.67 & 2 \\
$s_0$ & 18 & 9/18 [16.67, 66.67] & 16.67 & 15/18 [83.33, 83.33] & 16.67 & 2 \\
$s_0$ & 27 & 9/18 [16.67, 66.67] & 16.67 & 15/18 [83.33, 83.33] & 16.67 & 2 \\
$s_0$ & 36 & 9/18 [16.67, 66.67] & 16.67 & 11/18 [50.00, 66.67] & 16.67 & 2 \\
$s_0$ & 45 & 1/18 [0.00, 16.67] & 5.56 & 8/18 [0.00, 66.67] & 16.67 & 2 \\
$s_0$ & 54 & 2/18 [0.00, 33.33] & 5.56 & 1/18 [0.00, 16.67] & 16.67 & 2 \\
$s_0$ & 63 & 6/18 [0.00, 83.33] & 11.11 & 3/18 [0.00, 33.33] & 11.11 & 2 \\
$s_1$ & Majority & 3/18 [16.67, 16.67] & 33.33 & \scNA & \scNA & 0 \\
$s_1$ & Surface & 1/18 [0.00, 16.67] & 16.67 & \scNA & \scNA & 0 \\
$s_1$ & 0 & 3/18 [16.67, 16.67] & 22.22 & 3/18 [16.67, 16.67] & 22.22 & 0 \\
$s_1$ & 9 & 3/18 [16.67, 16.67] & 22.22 & 3/18 [16.67, 16.67] & 16.67 & 0 \\
$s_1$ & 18 & 2/18 [0.00, 16.67] & 22.22 & 3/18 [16.67, 16.67] & 16.67 & 0 \\
$s_1$ & 27 & 2/18 [0.00, 16.67] & 16.67 & 3/18 [16.67, 16.67] & 16.67 & 0 \\
$s_1$ & 36 & 2/18 [0.00, 16.67] & 16.67 & 3/18 [16.67, 16.67] & 22.22 & 0 \\
$s_1$ & 45 & 3/18 [16.67, 16.67] & 16.67 & 3/18 [16.67, 16.67] & 16.67 & 0 \\
$s_1$ & 54 & 3/18 [16.67, 16.67] & 16.67 & 3/18 [16.67, 16.67] & 16.67 & 0 \\
$s_1$ & 63 & 3/18 [16.67, 16.67] & 16.67 & 3/18 [16.67, 16.67] & 16.67 & 0 \\
$s_2$ & Majority & 3/18 [16.67, 16.67] & 22.22 & \scNA & \scNA & 5 \\
$s_2$ & Surface & 3/18 [16.67, 16.67] & 16.67 & \scNA & \scNA & 5 \\
$s_2$ & 0 & 6/18 [33.33, 33.33] & 16.67 & 4/18 [16.67, 33.33] & 16.67 & 5 \\
$s_2$ & 9 & 5/18 [16.67, 33.33] & 16.67 & 6/18 [33.33, 33.33] & 16.67 & 5 \\
$s_2$ & 18 & 5/18 [16.67, 33.33] & 16.67 & 5/18 [16.67, 33.33] & 16.67 & 5 \\
$s_2$ & 27 & 5/18 [16.67, 33.33] & 16.67 & 6/18 [33.33, 33.33] & 16.67 & 5 \\
$s_2$ & 36 & 4/18 [0.00, 33.33] & 16.67 & 4/18 [0.00, 33.33] & 16.67 & 5 \\
$s_2$ & 45 & 5/18 [16.67, 33.33] & 16.67 & 5/18 [16.67, 33.33] & 16.67 & 5 \\
$s_2$ & 54 & 5/18 [16.67, 33.33] & 16.67 & 3/18 [16.67, 16.67] & 16.67 & 5 \\
$s_2$ & 63 & 4/18 [16.67, 33.33] & 16.67 & 7/18 [33.33, 50.00] & 27.78 & 5 \\
A1 amt. & Majority & 9/18 [50.00, 50.00] & 50.00 & \scNA & \scNA & 0 \\
A1 amt. & Surface & 16/18 [66.67, 100.00] & 100.00 & \scNA & \scNA & 0 \\
A1 amt. & 0 & 18/18 [100.00, 100.00] & 100.00 & 18/18 [100.00, 100.00] & 100.00 & 0 \\
A1 amt. & 9 & 18/18 [100.00, 100.00] & 100.00 & 18/18 [100.00, 100.00] & 100.00 & 0 \\
A1 amt. & 18 & 18/18 [100.00, 100.00] & 100.00 & 18/18 [100.00, 100.00] & 100.00 & 0 \\
A1 amt. & 27 & 18/18 [100.00, 100.00] & 100.00 & 18/18 [100.00, 100.00] & 100.00 & 0 \\
A1 amt. & 36 & 18/18 [100.00, 100.00] & 100.00 & 18/18 [100.00, 100.00] & 100.00 & 0 \\
A1 amt. & 45 & 18/18 [100.00, 100.00] & 100.00 & 18/18 [100.00, 100.00] & 100.00 & 0 \\
A1 amt. & 54 & 18/18 [100.00, 100.00] & 100.00 & 18/18 [100.00, 100.00] & 100.00 & 0 \\
A1 amt. & 63 & 18/18 [100.00, 100.00] & 100.00 & 18/18 [100.00, 100.00] & 100.00 & 0 \\
A2 amt. & Majority & 6/18 [33.33, 33.33] & 33.33 & \scNA & \scNA & 0 \\
A2 amt. & Surface & 18/18 [100.00, 100.00] & 100.00 & \scNA & \scNA & 0 \\
A2 amt. & 0 & 15/18 [83.33, 83.33] & 83.33 & 18/18 [100.00, 100.00] & 100.00 & 0 \\
A2 amt. & 9 & 15/18 [83.33, 83.33] & 83.33 & 18/18 [100.00, 100.00] & 100.00 & 0 \\
A2 amt. & 18 & 15/18 [83.33, 83.33] & 83.33 & 18/18 [100.00, 100.00] & 100.00 & 0 \\
A2 amt. & 27 & 15/18 [83.33, 83.33] & 83.33 & 18/18 [100.00, 100.00] & 100.00 & 0 \\
A2 amt. & 36 & 15/18 [83.33, 83.33] & 83.33 & 18/18 [100.00, 100.00] & 100.00 & 0 \\
A2 amt. & 45 & 15/18 [83.33, 83.33] & 83.33 & 18/18 [100.00, 100.00] & 100.00 & 0 \\
A2 amt. & 54 & 15/18 [83.33, 83.33] & 83.33 & 18/18 [100.00, 100.00] & 100.00 & 0 \\
A2 amt. & 63 & 15/18 [83.33, 83.33] & 83.33 & 18/18 [100.00, 100.00] & 100.00 & 0 \\
A1 type & Majority & 15/18 [83.33, 83.33] & 83.33 & \scNA & \scNA & 0 \\
A1 type & Surface & 18/18 [100.00, 100.00] & 100.00 & \scNA & \scNA & 0 \\
A1 type & 0 & 18/18 [100.00, 100.00] & 100.00 & 18/18 [100.00, 100.00] & 100.00 & 0 \\
A1 type & 9 & 18/18 [100.00, 100.00] & 100.00 & 18/18 [100.00, 100.00] & 100.00 & 0 \\
A1 type & 18 & 18/18 [100.00, 100.00] & 100.00 & 18/18 [100.00, 100.00] & 100.00 & 0 \\
A1 type & 27 & 18/18 [100.00, 100.00] & 100.00 & 18/18 [100.00, 100.00] & 100.00 & 0 \\
A1 type & 36 & 18/18 [100.00, 100.00] & 100.00 & 18/18 [100.00, 100.00] & 100.00 & 0 \\
A1 type & 45 & 18/18 [100.00, 100.00] & 100.00 & 18/18 [100.00, 100.00] & 100.00 & 0 \\
A1 type & 54 & 18/18 [100.00, 100.00] & 100.00 & 18/18 [100.00, 100.00] & 100.00 & 0 \\
A1 type & 63 & 18/18 [100.00, 100.00] & 100.00 & 18/18 [100.00, 100.00] & 100.00 & 0 \\
A2 type & Majority & 12/18 [66.67, 66.67] & 66.67 & \scNA & \scNA & 0 \\
A2 type & Surface & 18/18 [100.00, 100.00] & 100.00 & \scNA & \scNA & 0 \\
A2 type & 0 & 15/18 [83.33, 83.33] & 83.33 & 18/18 [100.00, 100.00] & 100.00 & 0 \\
A2 type & 9 & 15/18 [83.33, 83.33] & 83.33 & 18/18 [100.00, 100.00] & 100.00 & 0 \\
A2 type & 18 & 15/18 [83.33, 83.33] & 83.33 & 18/18 [100.00, 100.00] & 100.00 & 0 \\
A2 type & 27 & 15/18 [83.33, 83.33] & 83.33 & 18/18 [100.00, 100.00] & 100.00 & 0 \\
A2 type & 36 & 15/18 [83.33, 83.33] & 83.33 & 18/18 [100.00, 100.00] & 100.00 & 0 \\
A2 type & 45 & 15/18 [83.33, 83.33] & 83.33 & 18/18 [100.00, 100.00] & 100.00 & 0 \\
A2 type & 54 & 15/18 [83.33, 83.33] & 83.33 & 18/18 [100.00, 100.00] & 100.00 & 0 \\
A2 type & 63 & 15/18 [83.33, 83.33] & 83.33 & 18/18 [100.00, 100.00] & 100.00 & 0 \\
\end{longtable}
\endgroup

\else
The full fixed-layer ledger is disabled in this shortened reading copy.
The complete source retains it in this section; set
\texttt{\string\SCFullProbeTablestrue} in the preamble to typeset all 1,365
records. The prespecified comparison remains in
Appendix~\ref{app:sc_probes}.
\fi

\end{document}